\documentclass[runningheads]{llncs}

\usepackage{eccv}

\usepackage{eccvabbrv}

\usepackage{graphicx}
\usepackage{booktabs}

\usepackage[accsupp]{axessibility}  %

\usepackage[utf8]{inputenc} %
\usepackage[T1]{fontenc}    %
\usepackage{url}            %
\usepackage{booktabs}       %
\usepackage{amsfonts}       %
\usepackage{nicefrac}       %
\usepackage{microtype}      %

\usepackage{caption}
\usepackage{multirow}
\usepackage{xspace}
\usepackage{mathtools}
\usepackage{wasysym}
\usepackage{marvosym}
\usepackage{xcolor}
\usepackage{color}
\usepackage{tabularx}
\usepackage{float}
\usepackage{diagbox,tabu,stackengine}
\usepackage{bbm}
\usepackage{bm}
\usepackage{mwe}
\usepackage{graphbox}
\usepackage{sidecap}

\usepackage{algorithm}
\usepackage[noend]{algpseudocode}
\usepackage{comment}
\usepackage{cancel}
\usepackage{ulem}
\usepackage{array}
\newcommand{\PreserveBackslash}[1]{\let\temp=\\#1\let\\=\temp}
\newcolumntype{C}[1]{>{\PreserveBackslash\centering}p{#1}}
\newcolumntype{R}[1]{>{\PreserveBackslash\raggedleft}p{#1}}
\newcolumntype{L}[1]{>{\PreserveBackslash\raggedright}p{#1}}

\usepackage{wrapfig}

\usepackage{xspace}
\usepackage{csquotes}
\makeatletter
\DeclareRobustCommand\onedot{\futurelet\@let@token\@onedot}
\def\@onedot{\ifx\@let@token.\else.\null\fi\xspace}
\def\eg{\textit{e.g}\onedot} 
\def\ie{\textit{i.e}\onedot}

\def\wrt{w.r.t\onedot} 
\def\etal{\textit{et al}\onedot}
\makeatother

\newcommand{\mysubsubsection}[1]{\vspace{-4mm}\subsubsection{#1 }}

\usepackage{amsmath,amsfonts,bm}

\def\eqref#1{equation~\ref{#1}}

\def\1{\bm{1}}

\def\rvepsilon{{\bm{\epsilon}}}

\def\rvx{{\mathbf{x}}}

\def\rvz{{\mathbf{z}}}

\def\vzero{{\bm{0}}}

\def\vtheta{{\bm{\theta}}}

\def\vc{{\bm{c}}}

\def\vh{{\bm{h}}}

\def\vx{{\bm{x}}}

\def\vz{{\bm{z}}}

\def\mA{{\bm{A}}}
\def\mB{{\bm{B}}}

\def\mI{{\bm{I}}}

\def\mW{{\bm{W}}}

\DeclareMathAlphabet{\mathsfit}{\encodingdefault}{\sfdefault}{m}{sl}
\SetMathAlphabet{\mathsfit}{bold}{\encodingdefault}{\sfdefault}{bx}{n}

\def\gD{{\mathcal{D}}}
\def\gE{{\mathcal{E}}}

\def\gN{{\mathcal{N}}}

\def\sZ{{\mathbb{Z}}}

\newcommand{\px}{p(\mathbf{x})}

\newcommand{\E}{\mathbb{E}}
\newcommand{\Ls}{\mathcal{L}}
\newcommand{\R}{\mathbb{R}}

\newcommand{\dmparam}{\vtheta}
\newcommand{\denoiser}{\rvepsilon_{\dmparam}}
\newcommand{\gparam}{\bm{\phi}}
\newcommand{\gnet}{g_{\gparam}}

\newcommand{\ourmodel}{DiPIR\xspace}

\usepackage[pagebackref,breaklinks,colorlinks,citecolor=eccvblue]{hyperref}

\newcommand{\orcidlink}[1]{}

\begin{document}

\title{Photorealistic Object Insertion with Diffusion-Guided Inverse Rendering}

\author{Ruofan Liang\inst{1,2,3}\orcidlink{0009-0005-7667-1809} \and
Zan Gojcic\inst{1}\orcidlink{0000-0001-6392-2158} \and
Merlin Nimier-David\inst{1}\orcidlink{0000-0002-6234-3143} \and
David Acuna\inst{1}\orcidlink{0000-0003-1308-1500} \and\\
Nandita Vijaykumar\inst{2,3}\orcidlink{0000-0003-3315-9336} \and
Sanja Fidler\inst{1,2,3}\orcidlink{0000-0003-1040-3260} \and
Zian Wang \inst{1,2,3}\orcidlink{0000-0003-4166-3807}
}

\authorrunning{R.~Liang et al.}

\institute{
$^{1}$NVIDIA \quad $^{2}$University of Toronto \quad $^{3}$Vector Institute}

\maketitle

\begin{abstract}
The correct insertion of virtual objects in images of real-world scenes requires a deep understanding of the scene's lighting, geometry and materials, as well as the image formation process.
While recent large-scale diffusion models have shown strong generative and inpainting capabilities, we find that current models do not sufficiently ``understand'' the scene shown in a single picture to generate consistent lighting effects (shadows, bright reflections, etc.) while preserving the identity and details of the composited object.
We propose using a personalized large diffusion model as \textit{guidance} to a physically based inverse rendering process.
Our method recovers scene lighting and tone-mapping parameters, allowing the photorealistic composition of arbitrary virtual objects in single frames or videos of indoor or outdoor scenes.
Our physically based pipeline further enables automatic materials and tone-mapping refinement.%
\keywords{Inverse rendering \and Diffusion models \and Personalization \and Virtual object insertion \and Physically based rendering}
\end{abstract}

\section{Introduction}
\label{sec:intro}

Virtual object insertion enables a range of applications from virtual production, to interactive gaming and synthetic data generation. To produce photorealistic insertions, the interactions between the virtual objects and the environment need to be modeled faithfully, such as accurate specular highlights and shadows.

A standard virtual object insertion pipeline typically includes three key steps: i) lighting estimation from the input image, ii) 3D proxy geometry creation, and iii) composited image rendering in a rendering engine. %
However, the first and arguably most important step is still an open research question. 
Lighting estimation is particularly challenging when dealing with limited inputs such as a \textit{single} image, captured using a consumer device with a low dynamic range. 
Indeed, inverse rendering is a fundamentally ill-posed problem. 

To constrain its solution space, prior works either aimed to define hand-crafted priors~\cite{land1971lightness1,grosse2009ground,bousseau2009user,zhao2012closed} or to learn them from the data~\cite{li2018learning,legendre2019deeplight,garon2019fast,hold2017deep,hold2019deep,zhu2021cvpr,wang2021learning,wang2022neural,SOLD-Net,gardner2019deep,gardner2017learning}. 
However, the former often fall short when applied to real-world scenes, while the latter suffer from scarcity of the ground truth data. 
As a result, these algorithms are often heavily tailored to a specific domain, e.g. indoor~\cite{neuralSengupta19,gardner2019deep,gardner2017learning,garon2019fast,wang2021learning} or outdoor scenes~\cite{hold2017deep,hold2019deep,wang2022neural,zhu2021cvpr,SOLD-Net}.

To address these challenges, we propose to reuse the strong image generation priors learned by large diffusion models (DMs)~\cite{rombach2021highresolution} as a guidance for inverse rendering.
Unlike hand-crafted or supervised data-driven priors that are often specific to a domain, DMs are trained on massive datasets and show a remarkable ``understanding'' of the world and the underlying physical concepts.
While DMs still often fail to produce accurate lighting effects such as shadows and reflections~\cite{sarkar2023geometry} in their generations, we observe that they can provide valuable guidance when combined with a physically-based renderer and are adapted to the scene.

Specifically, we present \textit{Diffusion Prior for Inverse Rendering} (\ourmodel), building on three main contributions. \textbf{First}, we use a physically based renderer to accurately simulate the interaction between the light and the 3D asset to generate the final composited image.
We also account for the unknown tone-mapping curve to mimic the camera sensor response.
\textbf{Second}, we propose a lightweight personalization scheme of the pre-trained DM, based on the input image and the type of inserted asset. 
\textbf{Third}, we design a variant of the SDS loss~\cite{poole2022dreamfusion} which makes use of this personalization and improves training stability. 

\begin{figure*}[t]
    \centering
    \includegraphics[width=\textwidth]{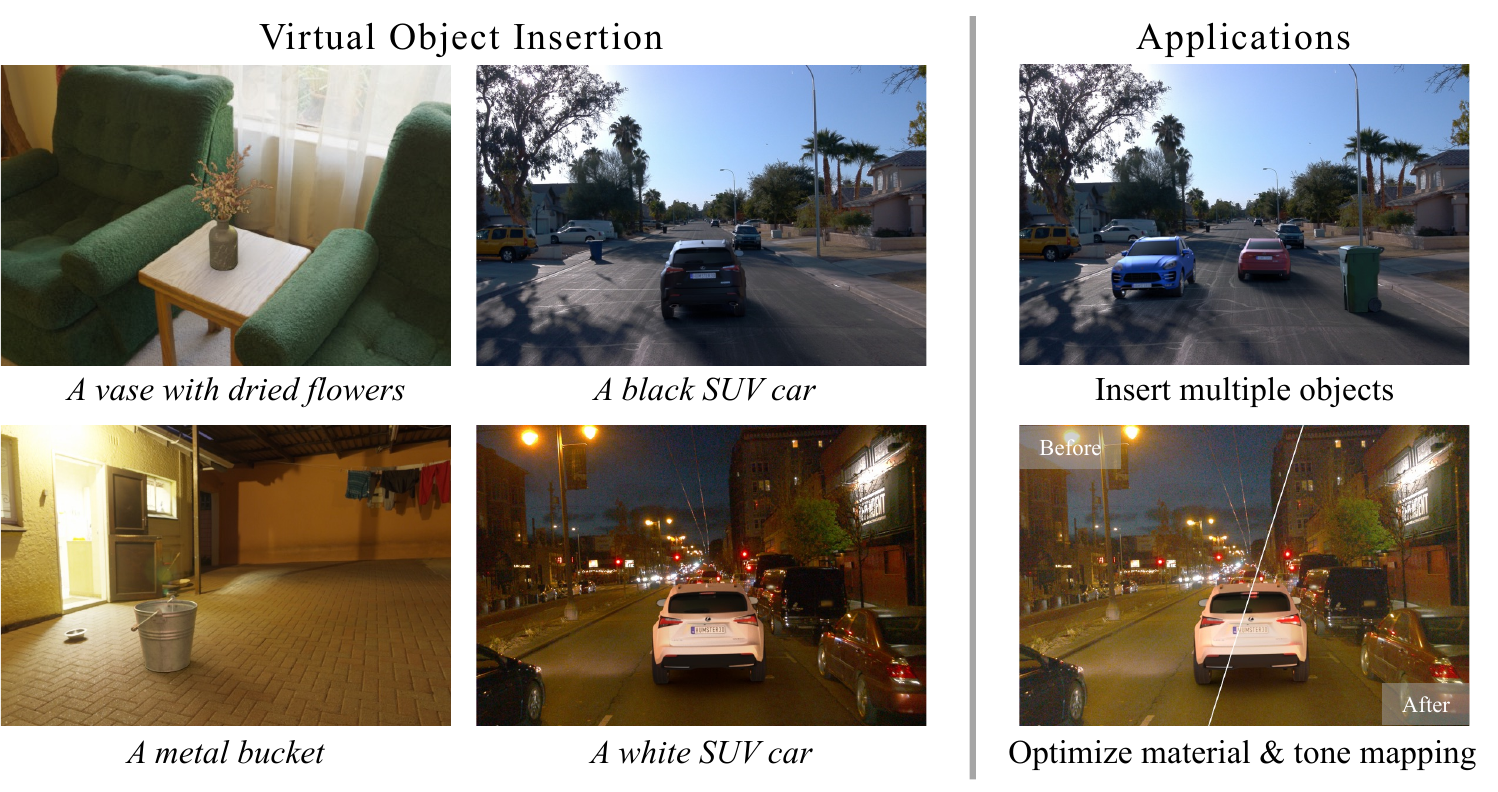}
    \captionsetup{font=footnotesize}
    \vspace{-6mm}
    \caption{
        We propose \ourmodel{}, a physically based method to recover lighting from a single image, enabling arbitrary virtual object compositing into indoor and outdoor scenes, as well as material and tone-mapping optimization. Project page: \url{https://research.nvidia.com/labs/toronto-ai/DiPIR/}
    }
    \label{fig:teaser}
    \vspace{-5mm}
\end{figure*}

In \ourmodel, the DM acts similarly to a human evaluator. 
It takes the edited image as input and propagates the feedback signal to physically-based scene attributes via differentiable rendering, thus enabling end-to-end optimization. 
We experimentally show that \ourmodel outperforms existing state-of-the-art lighting estimation methods for object insertion across indoor and outdoor datasets.

\section{Related Work}
\vspace{3mm}
\mysubsubsection{Inverse rendering} is the task of recovering intrinsic properties of a scene, including materials, shape, and lighting, from a single or multiple images~\cite{barrow1978recovering}. The main challenge of inverse rendering lies in the ill-posed nature of the task. Prior knowledge about materials and lighting effects is crucial to solving under-constrained inverse rendering problems, but how to best define and incorporate such priors remains an open research question.

Early methods formulated inverse rendering as an optimization problem and used hand-crafted regularization terms such as local smoothness and sparsity of materials~\cite{land1971lightness1,barrow1978recovering,grosse2009ground,bousseau2009user,zhao2012closed,barron2014shape} to constrain the solution space. However, real-world lighting effects are often complex and difficult to describe using hand-crafted priors.
The advent of deep learning has facilitated the learning of data-driven priors from ground truth supervision~\cite{bell2014intrinsic,kovacs2017shading,li2018cgintrinsics,neuralSengupta19,yu19inverserendernet,li2020inverse,li2020openrooms,Boss2020-TwoShotShapeAndBrdf,wang2021learning,wang2022neural,wimbauer2022rendering}. 
Yet, acquiring real-world data with accurately annotated intrinsic decomposition necessitates specialized capture devices~\cite{grosse2009ground} or expert annotators~\cite{bell2014intrinsic,kovacs2017shading}. Collecting real-world datasets at a large scale is therefore extremely challenging. As a result, existing methods often resort to synthetic datasets~\cite{neuralSengupta19,garon2019fast,li2020inverse,wang2021learning,zhu2021cvpr,SOLD-Net}, or are carefully designed to be training data-efficient~\cite{hold2017deep,hold2019deep,wang2022neural}. 
The resulting algorithms are therefore often tailored to specific domains, such as indoor~\cite{neuralSengupta19,gardner2019deep,gardner2017learning,garon2019fast,wang2021learning} or day-time outdoor scenes~\cite{hold2017deep,hold2019deep,wang2022neural,zhu2021cvpr,SOLD-Net}.

\mysubsubsection{Lighting estimation} is a subtask of inverse rendering that specifically focuses on inferring lighting and is a core component of photorealistic virtual object insertion pipelines. 
Existing lighting estimation methods are often designed as feed-forward neural networks that take a single image as input and directly regress lighting information in form of spherical lobes~\cite{garon2019fast,li2020inverse,zhao2020pointar,zhan2021emlight}, environment maps~\cite{gardner2017learning,neuralSengupta19,song2019neural,zhu2021cvpr,SOLD-Net}, parametric light sources~\cite{gardner2019deep}, or volumetric lighting~\cite{srinivasan2020lighthouse,wang2021learning,wang2022neural,li2023spatiotemporally}.
The environment lighting can also be recovered through the proxy geometry shown in the single image \cite{legendre2019deeplight,yu2023accidental}.
Recently, lighting estimation was also formulated as a generative task. StyleLight~\cite{wang2022stylelight} is based on a dual-branch StyleGAN model, which first performs GAN inversion through the LDR branch and then generates the panorama with the HDR branch. 
EverLight~\cite{Dastjerdi_2023_ICCV} instead regresses an initial HDR environment map from the input image and later refines it with a GAN model~\cite{karimi2022ImmerseGAN}. 
Concurrent to our work, DiffusionLight~\cite{Phongthawee2023DiffusionLight} proposed to inpaint a chrome ball in the center of the given image using an inpainting diffusion model and unwarp it to an environment map. To obtain the HDR environment map, DiffusionLight trains a LoRA~\cite{hu2022lora} for exposure bracketing and merges multiple generated LDR chrome balls with different exposures. 

Note that lighting is fundamentally a High Dynamic Range (HDR) quantity, and reproducing this full range still remains challenging. HDR is crucial for basic effects such as sharp shadows, particularly in outdoor day-time settings where the sun is several orders of magnitudes brighter than the rest of the environment.

\mysubsubsection{Physically-based differentiable rendering.}
Rendering algorithms such as path tracing~\cite{kajiya1986rendering} aim to produce realistic images of virtual scenes by accurately simulating the physical processes involved in light transport. Physically-based rendering (PBR) typically uses a geometric representation of the scene, scattering functions modeling the behavior of the scene's surfaces, as well as camera and lighting models.
Recent years have seen the development of many \textit{differentiable} rendering methods and frameworks~\cite{li2018differentiable,cheng2020pathspace,cheng2021pathspace,NimierDavidVicini2019Mitsuba2,nimierdavid2020radiative,vicini2021path,jakob2022mitsuba3}. 

While particular attention has been devoted to the difficult problem of computing derivatives \textit{at discontinuities}~\cite{li2018differentiable,loubet2019reparametrizing,bangaru2020unbiased,yan2022efficient,zhang2023projective}, such as due to visibility changes, we focus here on the comparatively simpler lighting, material and tone-mapping derivatives, which generally do not introduce discontinuities. 

\mysubsubsection{Diffusion model priors and personalization.} 
Priors learned by the text-to-image DMs~\cite{rombach2021highresolution,balaji2022eDiff-I, saharia2022photorealistic,dai2023emu}, on large-scale datasets can be adapted to various applications such as 
monocular depth estimation~\cite{ke2023repurposing}, authentic image completion~\cite{tang2023realfill}, and image restoration~\cite{chari2023personalized}. Among various adaptation techniques, DreamBooth~\cite{ruiz2022dreambooth} proposes to finetune the DM on subject images, while Textual Inversion~\cite{gal2022textual} optimizes new word embeddings for target generation. LoRA~\cite{hu2022lora,shah2023ziplora} simplifies adaptation by freezing the pretrained model weights and injecting trainable rank decomposition matrices, thus greatly reducing the number of trainable parameters. Most similar to ours are concurrent works on DM adaptation for lighting estimation~\cite{Phongthawee2023DiffusionLight} and intrinsic image decomposition~\cite{kocsis2023iid}. The former adapts the DM to exposure bracketing for LDR to HDR lifting, while the latter fine-tunes a pretrained SD model to directly output the albedo and BRDF properties of an image.

\section{Preliminaries}

\subsubsection{Diffusion Models.}

Diffusion models are a family of generative models built around two key processes. A forward process, which gradually adds noise to data samples  $\rvx \sim \px$ removing their structure over time $t$. This is achieved using a noise schedule determined by $\alpha_t$ and $\sigma_t$ as $\rvx_t=\alpha_t \rvx + \sigma_t \rvepsilon, \rvepsilon \sim \gN(\vzero,\mI)$. In contrast, the reverse process gradually removes this noise, restoring the structure. The reverse process is  parameterized  by a conditional neural network, $\denoiser$, trained to predict the noise $\rvepsilon$ at a given timestep $t$ according to the following simplified objective~\cite{ho2020denoising}:
\begin{equation}
    \E_{\rvx \sim \px,\rvepsilon \sim \gN(\vzero,\mI), t \sim T} \left[ \, w(t)|| \denoiser(\rvx_t,t,\vc) -\rvepsilon ||^2_2 \, \right],
    \label{eq:vanilla_difussion}
\end{equation}
where $\vc$ represents a condition (\eg~text, image, etc.) that allows controlling the generation process, $w(t)$ represents a time-conditional weighting, and $T$ is a set containing a selection of timesteps.

In this work, we use a Latent Diffusion Model (LDM)~\cite{rombach2021highresolution} in which the diffusion process is conducted in a lower-dimensional latent space. Specifically, the encoder $\gE$ maps samples from the data distribution $\rvx \sim \px$ into a latent space $\sZ$. The decoder $\gD$ performs the inverse operation, such that $\gD\left(\gE(\rvx)\right) \approx \rvx$. As follows, for LDMs $\rvx$ in Eq.~\ref{eq:vanilla_difussion} is replaced by its latent  $\rvz=\gE(\rvx)$.

\subsubsection{Personalization and Fine-tuning.}

Fine-tuning all parameters of a pretrained DM requires significant computational resources and time. To alleviate this, LoRA \cite{hu2022lora} injects trainable low-rank decomposition matrices and aims to learn only the variations from the pretrained weights.
Specifically, consider a linear layer represented as $\vh = \mW_0 \vx$. Here $\mW_0 \in \R^{n \times n} $ and $\vx \in \R^n$  are the pretrained weights and input, respectively. Applying LoRA, this layer is modified to $\vh = (\mW_0 + \bm{\Delta W}) \vx=\mW_0 \vx + \mA \mB \vx$. Notably, $ \bm{\Delta W} = \mA \mB$ is a combination of low-rank matrices $ \mA \in \R^{n \times r}$ and $ \mB \in \R^{r \times n}$ with $ r \ll n $.
During LoRA fine-tuning, only the added term $ \bm{\Delta W}$ is updated while the parameters $ \mW_0$ remain unchanged. The rank $r$ is a hyperparameter that trades efficiency with model capacity.

\section{Method} 
\label{sec:method}

\begin{figure*}[t!]
    \centering
     \includegraphics[width=0.99\textwidth]{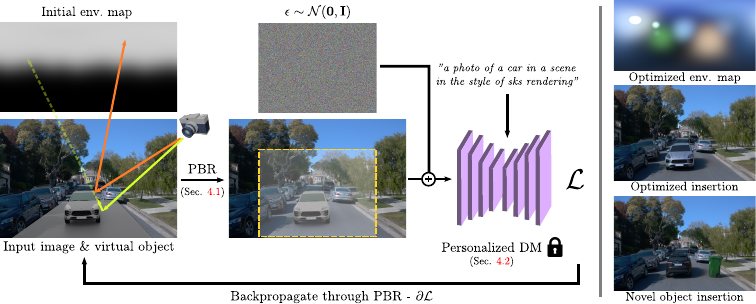}
    \vspace{-2mm}
    \captionsetup{font=footnotesize}
    \caption{
        \textbf{Method overview.} 
        Given an input image, we first construct a virtual 3D scene with a virtual object and proxy plane. 
        Our physically-based renderer then differentiably simulates the interactions of the optimizable environment map with the inserted virtual object and its effect on the background scene (shadowing)  (\textit{left}). 
        At each iteration, the rendered image is diffused and passed through a personalized diffusion model~(\textit{middle}). 
        The gradient of the adapted Score Distillation formulation is propagated back to the environment map and the tone-mapping curve through the differentiable renderer. 
        Upon convergence, we recover lighting and tone-mapping parameters, which allow photorealistic compositing of virtual objects from a single image~(\textit{right}). 
    }
    \label{fig:model} 
    \vspace{-3mm}
\end{figure*}

\newcommand{\diffusionPersonalizationFigure}{%
    \setlength{\columnsep}{1em}
    \setlength{\intextsep}{0em}
    \begin{wrapfigure}[12]{r}{0.5\textwidth}
        \centering
        \vspace{2mm}
        \includegraphics[width=0.5\textwidth]{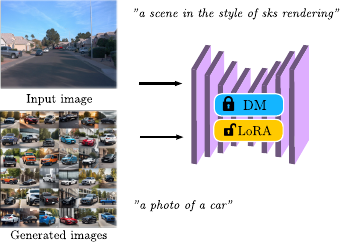}
        \vspace{-5mm}
        \captionsetup{font=scriptsize}
        \caption{
            Personalization with concept preservation.
        }
        \label{fig:personalization}
    \end{wrapfigure}
}

Given a single image as input, \ourmodel recovers scene lighting and tone-mapping parameters, with the goal of photorealistic insertion of virtual objects. An overview of our method is shown in Fig.~\ref{fig:model}. Sec.~\ref{sec:diffrender} describes our representation and the differentiable rendering process, while Sec.~\ref{sec:diffusion_guidance} and Sec.~\ref{sec:optimization_formulation} provide details on diffusion model guidance and optimization formulation, respectively.

\vspace{-3mm}
\subsection{Physically-based Virtual Object Insertion} 
\label{sec:diffrender}
\vspace{3mm}

\mysubsubsection{Virtual scene.} 
Inserting a virtual object $\mathcal{X}$ into an image $\mathbf{I}_\text{bg} \in \mathbb{R}^{h \times w \times 3}$ requires creating a 3D proxy virtual scene, viewed from the correct camera pose. Here, we assume that the user provides a specific placement (pose) for $\mathcal{X}$, but in some cases, an appropriate pose can also be determined automatically, e.g. by detecting the floor plane and scene scale.

To model the effects of the inserted object on the original image, such as shadows cast by $\mathcal{X}$, we also assume a known proxy geometry $\mathcal{P}$. 
We found a simple ground plane acting as a \textit{shadow catcher} underneath the virtual object to be sufficient in all of our experiments.
This proxy plane can be easily placed manually or automatically generated based on e.g. depth or LiDAR data.

\mysubsubsection{Light representation.}

We represent the scene's lighting $\mathbf{L}$ with a set of $N$ optimizable Spherical Gaussian (SG) parameters $\left\{\mathbf{c}_k, \bm{\mu}_k, \sigma_k\right\}_{k=1}^N \in \mathbb{R}^{N \times 7}$, where the radiance for one SG lobe at the direction $\bm{v} \in \mathbb{R}^3$ is defined as
\begin{equation}
    \mathbf{G}_k(\bm{v}; \mathbf{c}_k, \bm{\mu}_k, \sigma_k) = \mathbf{c}e^{-(1 - \bm{v} \cdot \bm{\mu}) / \sigma^2},
    \quad \text{where~}\mathbf{c}_k \in \mathbb{R}^3, \bm{\mu}_k \in \mathbb{R}^3, \sigma_k \in \mathbb{R}_+.
\end{equation}
The overall environment map $\mathbf{L}\in \mathbb{R}^{H \times W \times 3}$ is computed as: 
\begin{equation}
	\mathbf{L}_{i,j} = \sum_{k=1}^N \mathbf{G}_k \left(\bm{v}_{i,j}; \mathbf{c}_k, \bm{\mu}_k, \sigma_k \right),
\end{equation} 
where $\bm{v}_{i,j}$ is the direction corresponding to the pixel $(i, j)$ using the standard spherical environment parameterization.
Note that we chose a SG-based lighting representation for its good convergence properties and simplicity, but many alternatives exist and might be combined with our method.

\mysubsubsection{Differentiable rendering.} 
Inserting a virtual object into a scene involves simulating the interactions of the optimizable environment map with the inserted virtual object (foreground), and the inserted object's effect on the background scene (shadow). We describe below the details respectively.

\paragraph{Foreground image.}
Given the optimized light representation $\mathbf{L}$ from above and the virtual object $\mathcal{X}$ (including its geometry and materials), a foreground image $\mathbf{I}_\text{fg}$ of the inserted object can be rendered directly using standard path tracing:
\begin{equation}
    \mathbf{I}_\text{fg} = \text{PathTrace}(\mathcal{X}, \mathbf{L}, D),
    \label{eq:render_fg}
\end{equation} 
$D$ is the maximum number of interactions along the light path.
Since we do not assume that the provided proxy geometry is accurate nor that it has materials, we omit the effect of light reflected by the object into the scene, for example, due to a highly specular inserted object.

\paragraph{Shadow ratio.} 
As the background image $\mathbf{I}_\text{bg}$ already faithfully represents the scene, our task is limited to simulating the appearance of the inserted object and its effect on the nearby parts of the scene, \ie the shadows cast by the object onto the proxy geometry.

The shadow ratio $\bm{\beta}_\text{shadow}\in \mathbb{R}^{h \times w \times 3}$, which accounts for the effect of the object on the surrounding scene, is computed as the ratio between the radiance received by the proxy geometry with and without the inserted object present. A low value of $\bm{\beta}_\text{shadow}$ indicates a strongly shadowed region:
\begin{equation}
    \bm{\beta}_\text{shadow} = \frac{\text{PathTrace}(\mathcal{X} \cup \mathcal{P}, \mathbf{L}, 1)}{\text{PathTrace}(\mathcal{P}, \mathbf{L}, 1)},
 \label{eq:render_shadow}
\end{equation}
This is similar to the method of Wang~\etal~\cite{wang2022neural}, however we use Multiple Importance Sampling (MIS)~\cite{veach1995optimally} between lighting and BSDF for better sampling efficiency.
The maximum path length has been limited to 1 interaction in order to reduce memory and computational cost.
Unless specific material information is provided with the proxy geometry, we use a Lambertian BSDF, in which case the albedo cancels out in the ratio.
In practice, the computation of the shadow ratio and foreground image are combined in order to reuse shared terms.

\mysubsubsection{Tone-mapping.} 
To compensate for the unknown tone-mapping of the input image, we introduces an optimizable tone correction function $f(\cdot)$ that is applied on the inserted object $\hat{\mathbf{I}}_\text{fg} = f(\mathbf{I}_\text{fg}; \bm{\theta}_\text{fg})$ and shadows $\hat{\bm{\beta}}_\text{shadow} = f(\bm{\beta}_\text{shadow}; \bm{\theta}_\text{shadow})$. 
Our tone curve parameterization follows from \cite{durkan2019neural} and employs  a monotonic rational-quadratic spline as its basic building block. 
This spline is composed of $K_\text{s}$ bins, each defined by the quotient of two quadratic polynomials.
Notably, these functions are differentiable, and allow direct parameterization of the derivatives and heights at each knot. 
We apply a single spline for the foreground image, and different correction splines for each RGB channel for the shadow ratio to provide the flexibility to adjust the color of the shadow. We set $K_\text{s}=5$ for these curves.

\mysubsubsection{Differentiability.}
The final output image is an alpha-composite of the foreground object, shadows, and background image:
\begin{align}
    \mathbf{I}_\text{comp} 
        &= \left( 1 - \mathbf{V}(\mathcal{X}) \right) \cdot \hat{\bm{\beta}}_\text{shadow} \cdot \mathbf{I}_\text{bg} 
           + \mathbf{V}(\mathcal{X}) \cdot \hat{\mathbf{I}}_\text{fg},
    \label{eq:render_composite}
\end{align}
where $\mathbf{V}(\mathcal{X})$ equals $1$ when $\mathcal{X}$ is directly visible from the camera, and $0$ otherwise.

As neither foreground rendering, shadow rendering, or the compositing operation require derivatives \wrt the object placement or other discontinuous quantities, we can rely on automatic differentiation to obtain gradients of any pixelwise loss \wrt the lighting or material properties.
To this end, we use the Path Replay Backpropagation~\cite{vicini2021path} integrator of Mitsuba~3~\cite{jakob2022mitsuba3}.
The optimizable parameters are the Spherical Gaussian coefficients used in the lighting representation, and the tone curves' parameters $\bm{\theta}_\text{fg}, \bm{\theta}_\text{shadow}$.

While each of the operations above include simplifications (e.g. we do not account for secondary lighting from the proxy geometry), we have found them to be sufficient in practice: the remaining imperfections in the simulation can be sufficiently offset by the lighting optimization.

\subsection{Diffusion Guidance}
\label{sec:diffusion_guidance}

The composited image produced by our differentiable rendering pipeline serves as the input to a DM that is used to compute a guidance signal, employing an optimization objective similar to Score Distillation Sampling (SDS)~\cite{poole2022dreamfusion}.
However, while DMs inherently have robust priors for lighting, we found that they do not provide out-of-the-box the necessary guidance for our specific needs. 
Consequently, we propose an adaptive score distillation loss specifically designed for object insertion tasks that exploits a personalization strategy which we detail in the following section.

\mysubsubsection{Personalization with concept preservation.}
Off-the-shelf DMs often do not provide robust guidance for virtual object insertion, especially in out-of-distribution scenes such as outdoor driving environments. 
A potential solution, inspired by \cite{tang2023realfill}, is to adapt the DM using an image from the target scene. 
In our experience, however, this approach often resulted in too much overfitting to the target scene's content, reducing the model’s ability to adapt to the scene with a newly inserted object.
This led to artifacts and an unstable optimization process.

\diffusionPersonalizationFigure
To mitigate this issue, we propose fine-tuning the DM  with a focus on preserving the identity of the objects to be inserted (Fig.~\ref{fig:personalization}).
We specifically achieved this by generating additional synthetic images for the insertable class concept (e.g. car).
We sample those images from the off-the-self DM, starting with a base prompt such as ``a photo of a car'' and appending attributes such as color, background, lighting, and size to ensure diversity in the generated data. 

We employ LoRA with rank 4 to fine-tune the DM, combining the in-domain 
target example with this supplemental data. 
Our training follows the objective described in Eq.~\ref{eq:vanilla_difussion}, where  $\vc$ corresponds to two predefined prompts: \textit{``a scene in the style of sks rendering''} for the target image, and \textit{``a photo of a \{concept class\}''}.
We sample approximately 30-40 supplementary images for indoor scenes and 200 for outdoor scenes.  The total time spent for fine-tuning is typically less than 15 minutes on one high-end GPU with mixed precision training.

\mysubsubsection{Score Distillation with adapted guidance.}
Score Distillation Sampling \cite{poole2022dreamfusion} leverages a pretrained DM to guide the optimization of a differentiable, parametric image-rendering function $\gnet :=\vx$.  Specifically, the parameters $\gparam$ in our scenario corresponds to parameters of the Spherical Gaussian lighting and tone-mapping curves, are updated using the gradient:
\begin{equation}
    \nabla_{\gparam} \Ls_{\text{SDS}}(\gparam,\dmparam) 
    := \E_{\rvepsilon \sim \gN(\vzero,\mI), t \sim T}\left[\,
        w(t) \, \left( \hat{\rvepsilon}_{\dmparam}(\vz_t,t,\vc)-\rvepsilon \right) \, \frac{\partial \vz_t}{\partial \gparam} 
       \,\right],
    \label{eq:sds_gradient}
\end{equation}
where $\vz = \gE(\gnet)$ and $ \hat{\rvepsilon}_{\dmparam}(\vz_t,t,\vc)
    := (1 + s) \, \denoiser(\vz_t,t,\vc) - s \, \denoiser(\vz_t,t,\varnothing)
$. $\hat{\rvepsilon}$ denotes the classifier free guidance (CFG) version \cite{ho2022classifier} of $\denoiser$ used in text-conditioned DMs to enable higher quality generation via a guidance scale parameter $s$.  

Unfortunately, we encountered training instabilities when applying the original formulation of the SDS loss to our problem. Instead, we adopt the following alternative that integrates the LoRA personalization, and we call it LDS loss:
\begin{equation}
    \nabla_{\gparam} \Ls_{\text{LDS}}(\gparam,\dmparam)
    := \E_{\rvepsilon \sim \gN(\vzero,\mI), t \sim T}\left[\,
        w(t) \, \left(
            {\rvepsilon}_{(\dmparam + \bm{\Delta W})}(\vz_t,t,\vc)-{\rvepsilon}_{\dmparam}(\vz_t,t,\varnothing)
        \right) \, \frac{\partial \vz_t}{\partial \gparam}
       \,\right],
\label{eq:sds_ours}
\end{equation}
where ${\rvepsilon}_{(\dmparam + \bm{\Delta W})}$ represents the predicted noise of the LoRA personalized model described in the previous section.  Notably, this loss function bears resemblance to the 
Classifier Score Distillation (CSD)~\cite{yu2023text}. However, it's important to note that in our case, the delta is calculated between the adapted and non-adapted versions of the diffusion model. Intuitively,  this delta guides the optimization process in a direction determined by the personalized model which preserves the concept of the inserted object while also capturing the appearance and semantics of the specific scene.

\subsection{Optimization Formulation}
\label{sec:optimization_formulation}

After completing the diffusion model's personalization, we optimize the lighting and tone-mapping parameters using the following loss function:
\begin{equation}
    \Ls = \Ls_{\text{LDS}} + \lambda_\text{consistency} \Ls_\text{consistency} + \lambda_\text{reg} \Ls_\text{reg}.
\end{equation}

The loss' main component is the diffusion guidance $\Ls_{\text{LDS}}$, which provides a perceptual realism objective on the edited result with object insertion. 
We use the text prompt 
\textit{``a photo of a \{concept class\} in a scene in the style of sks rendering''}, which is a natural combination of the two text prompts used in personalization (Fig.~\ref{fig:personalization}).
Here \textit{\{concept class\}}, \eg \textit{car}, provides the context of the object to insert, while the personalized token \textit{sks} provides the lighting style of the input image. 
The diffusion guidance $\Ls_{\text{LDS}}$ is applied on the edited image $\mathbf{I}_\text{comp}$ (Eq.~\ref{eq:render_composite}), and backpropagated to the optimizable parameters through the differentiable rendering process (Eq.~\ref{eq:sds_ours}).

The two regularizers $\Ls_\text{consistency}$ and $\Ls_\text{reg}$ are related to environment map fusion, which we detail below.

\mysubsubsection{Environment map initialization and fusion.}
\label{sec:two_envmaps_fusion}
The personalized DM provides guidance in two ways: (i) encouraging lighting consistency of the foreground object and the scene, such as the reflections and scale of the inserted object; and (ii) encouraging accurate shadows cast by the inserted object onto the background scene, such as the scale, direction, and color of the shadow. 
However, we empirically observe that these two signals can conflict in the early phase of the optimization. 

To address this, we initialize two separate environment maps, $\mathbf{L}^\text{fg}, \mathbf{L}^\text{shadow}\in \mathbb{R}^{H\times W \times 3}$, to light the foreground inserted object (Eq.~\ref{eq:render_fg}) and cast shadows (Eq.~\ref{eq:render_shadow}) respectively. 
As the optimization progresses, the two environment maps are progressively fused into a single environment map $\mathbf{L}^\text{fused}$
through scaling the the $\mathbf{L}^\text{fg}$ by the relative luminance of between $\mathbf{L}^\text{fg}$ and $\mathbf{L}^\text{shadow}$. We also use the following two regularization terms for this fusion process.
First, we encourage the consistency between the normalized luminance $\tilde{\mathbf{L}}^\text{fg}, \tilde{\mathbf{L}}^\text{shadow} \in \mathbb{R}^{H\times W}$ of the environment maps by minimizing: 
\begin{equation}
    \Ls_\text{consistency} = -\sum_{i, j} \tilde{\mathbf{L}}^{\text{shadow}}_{i, j} \log\left(\tilde{\mathbf{L}}^\text{fg}_{i, j}\right) \Delta\Omega_{i, j}
\end{equation}
where $\Delta\Omega_{i, j}$ is the corresponding solid angle at pixel $(i, j)$ and the gradient for $\tilde{\mathbf{L}}^{\text{shadow}}$ is detached. 
Second, as the shadow environment map $\mathbf{L}^\text{shadow}$ is supervised mainly via the shadow ratio $\bm{\beta}_\text{shadow} \in \mathbb{R}^{H\times W\times 3}$ of Eq.~\ref{eq:render_shadow}, to encourage concentrated high peaks for the sharp shadows and suppress the ambient light in $\mathbf{L}^\text{shadow}$, we also add a L2 regularizer in log-space with the Cauchy loss~\cite{black1996robust}:
\begin{equation}
    \Ls_\text{reg} = \sum_{i, j, c} \log\left(1+2 \left({\mathbf{L}}^\text{shadow}_{i, j, c}\right)^2\right) \Delta\Omega_{i, j}.
\end{equation}
Please refer to the Supplement for more details. 

\mysubsubsection{Training details.}
The personalized DM receives a crop of $\mathbf{I}_\text{comp}$ (Eq.~\ref{eq:render_composite}) with added noise. 
Since the inserted object likely does not cover the full background image, we use the visibility mask of the virtual object $V(\mathcal{X})$ to locate the 2D bounding box of the inserted object, and randomly crop around the inserted object. The size of the cropped image is also randomized, with smaller crops covering local details of the inserted object and larger crops providing more visual cues of the lighting effects within the scene. 
Each scene is optimized for 600 iterations, and the maximum strength of the diffusion guidance is linearly decreased from $0.6$ to $0.3$. We use $\lambda_\text{consistency}=0.03$ and $\lambda_\text{reg}=0.01$. 

\vspace{-2mm}
\section{Experiments}
\label{sec:results}
\vspace{-3mm}

\begin{table}[t!]
    \captionsetup{font=footnotesize}
    \centering
    \caption{
        \textbf{Quantitative user study on outdoor street scenes}.
        For each scene, users are shown two results---one produced by our method, and another produced by one of the baselines---and select which is more realistic.
        We report the results averaged across 3 user studies with 9 users each.
        Our method outperforms all baselines ($> 50\%$) and is preferred in almost all illumination conditions.
    } 
    \vspace{-3mm}
    \label{table:userstudy_waymo} 
    \resizebox{0.73\linewidth}{!}{
        \addtolength{\tabcolsep}{6pt}
        \begin{tabular}{lccccc}
            \toprule
            \multirow{2}{*}{} &  \multicolumn{2}{c}{Daytime} & \multirow{2}{*}{Twilight} & \multirow{2}{*}{Night} & \multirow{2}{*}{All scenes} \\
             &  Sunny &  Cloudy &  &  &  \\
            \midrule 
            Hold-Geoffroy \etal~\cite{hold2019deep} 	         & $60.8$\%  & $66.7$\%  & $74.1$\%  & $85.7$\% & $68.8$\%   \\ 
            NLFE~\cite{wang2022neural}      	                 & $80.4$\%  & $73.3$\%  & $44.4$\%  & $52.4$\% & $67.4$\%   \\ 
            StyleLight~\cite{wang2022stylelight}                 & $76.5$\%  & $91.1$\%  & $66.7$\%  & $66.7$\% & $77.8$\%   \\ 
            DiffusionLight~\cite{Phongthawee2023DiffusionLight}  & $80.4$\%  & $68.9$\%  & $55.6$\%  & $71.4$\% & $70.8$\%   \\ 
            \bottomrule
        \end{tabular}
    }
    \vspace{-2mm}
\end{table} 
\begin{table}[t]
    \captionsetup{font=footnotesize}
    \centering
    \caption{
        \textbf{Quantitative evaluation on PolyHaven scenes}.  
        We report user study preference scores similar to Table~\ref{table:userstudy_waymo}.
        Metrics are computed \wrt to a ``reference'' image where the virtual object is lit by the ground-truth environment map.  
    } 
    \vspace{-3mm}
    \label{table:userstudy_polyhaven} 
    \resizebox{0.73\linewidth}{!}{
        \setlength{\tabcolsep}{4pt}
        \begin{tabular}{lccccc}
            \toprule
            Methods & Ours preferred $\downarrow$ & RMSE $\downarrow$ & SSIM $\uparrow$ & LPIPS \cite{zhang2018unreasonable} $\downarrow$ & si-RMSE \cite{grosse2009ground} $\downarrow$\\
            \midrule 
            Wang \etal~\cite{wang2021learning}                    & $84.8$\%  & $0.063$  & $0.985$ & $0.0175$ & $0.037$  \\ 
            StyleLight~\cite{wang2022stylelight}                 & $75.8$\%  & $0.056$  & $0.986$ & $0.0202$ & $0.039$  \\ 
            DiffusionLight~\cite{Phongthawee2023DiffusionLight}  & $66.7$\%  & $0.062$  & $0.985$ & $0.0162$ & $0.034$  \\ 
            Ours                                                 & /         & $\bm{0.048}$  & $\bm{0.989}$ & $\bm{0.0147}$ & $\bm{0.027}$   \\ 
            \bottomrule
        \end{tabular}
    }
    \vspace{-5mm}
\end{table}

In this section, we evaluate {\ourmodel} on a collection of outdoor and indoor scenes, covering diverse lighting conditions and different application domains. 
We first outline the experiment settings in Sec.~\ref{sec:experiment_settings} and then provide evaluation results on benchmark datasets in Sec.~\ref{sec:benchmark}. Finally, we conduct an ablation study in Sec.~\ref{sec:ablation}.

\begin{table}[t]
    \centering
    \scriptsize
    \captionsetup{font=small}
    \setlength\tabcolsep{0pt}
    \begin{tabularx}{\linewidth}%
        {*{5}{>{\centering\arraybackslash}X}}
        Reference   & Ours & DiffusionLight\cite{Phongthawee2023DiffusionLight} & StyleLight\cite{wang2022stylelight} & Wang \etal\cite{wang2021learning} \\
        \includegraphics[width=0.98\linewidth, trim={0 0 0 100pt},clip]{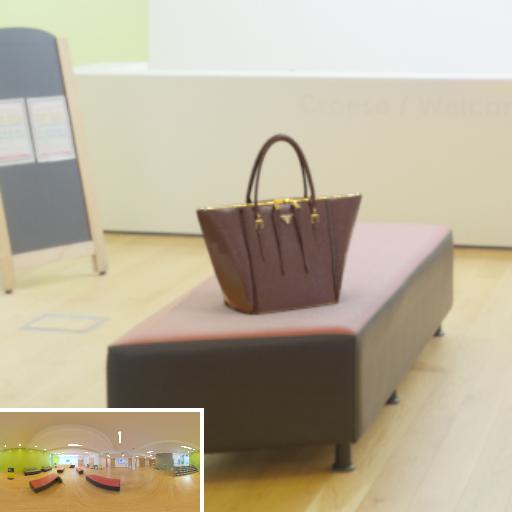} &
        \includegraphics[width=0.98\linewidth, trim={0 0 0 100pt},clip]{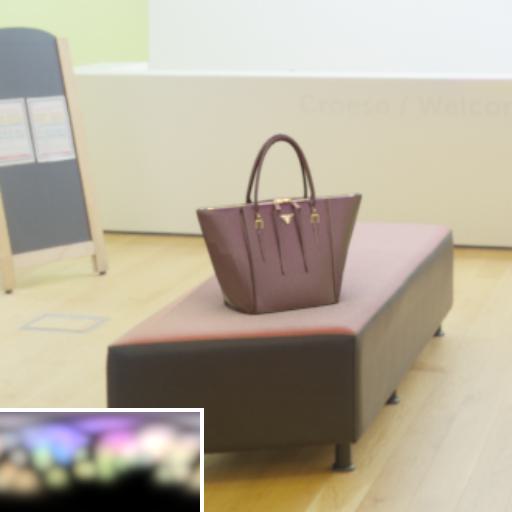} & 
        \includegraphics[width=0.98\linewidth, trim={0 0 0 100pt},clip]{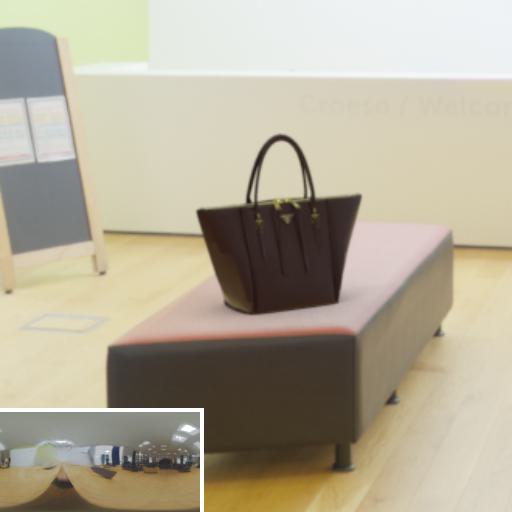} & 
        \includegraphics[width=0.98\linewidth, trim={0 0 0 100pt},clip]{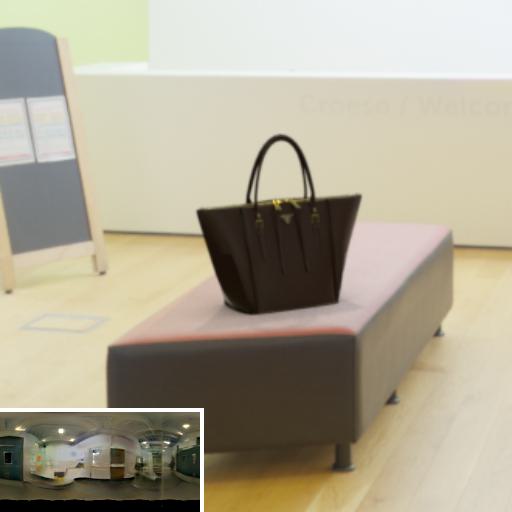} &
        \includegraphics[width=0.98\linewidth, trim={0 0 0 100pt},clip]{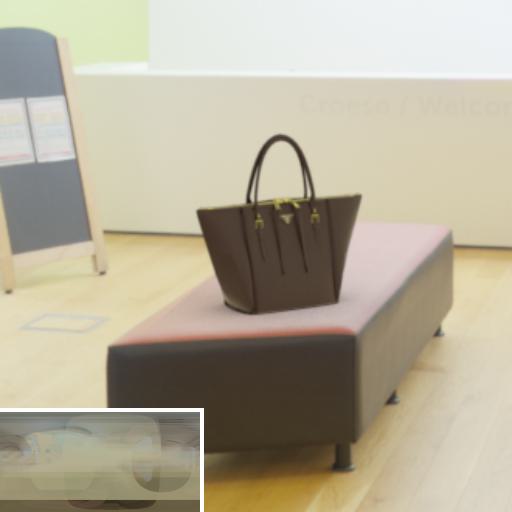}
        \\
        \includegraphics[width=0.98\linewidth, trim={0 0 0 100pt},clip]{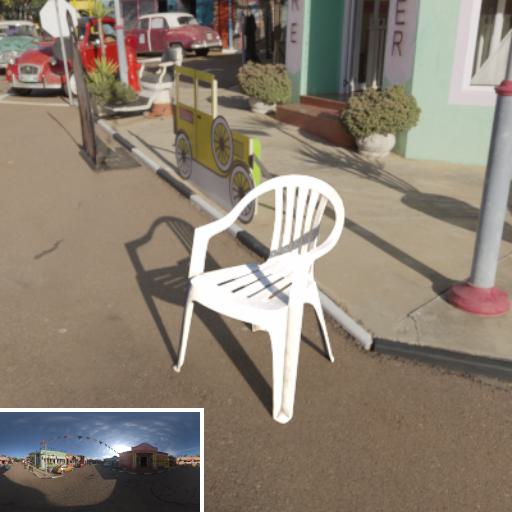} &
        \includegraphics[width=0.98\linewidth, trim={0 0 0 100pt},clip]{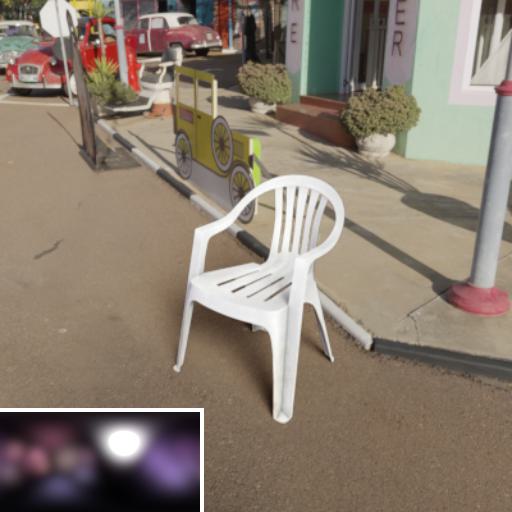} &
        \includegraphics[width=0.98\linewidth, trim={0 0 0 100pt},clip]{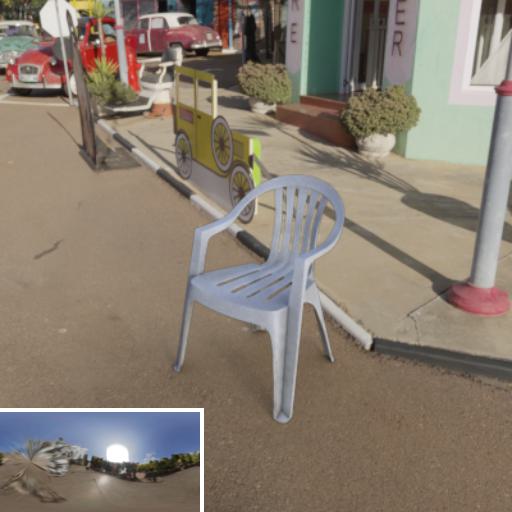} &
        \includegraphics[width=0.98\linewidth, trim={0 0 0 100pt},clip]{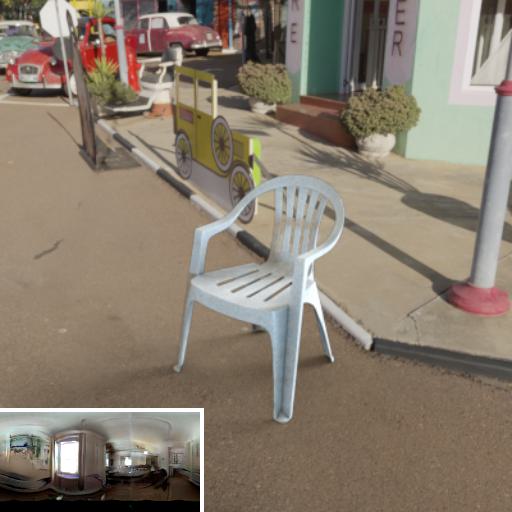} &
        \includegraphics[width=0.98\linewidth, trim={0 0 0 100pt},clip]{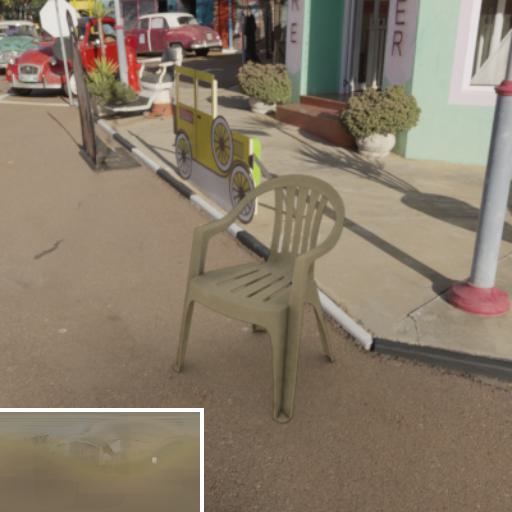}
    \end{tabularx}%
    \makeatletter\def\@captype{figure}\makeatother
    \caption{
        Comparisons on inserting objects into cropped HDRIs from PolyHaven.
    }
    \vspace{-7mm}
    \label{fig:qual_polyhaven}
\end{table}

\begin{table}[t]
    \centering
    \scriptsize
    \captionsetup{font=small}
    \setlength\tabcolsep{0pt}
        \begin{tabularx}{\linewidth}%
        {*{5}{>{\centering\arraybackslash}X}}
        Ours &
        DiffusionLight\cite{Phongthawee2023DiffusionLight} & StyleLight\cite{wang2022stylelight} & NLFE\cite{wang2022neural} & H-G \etal\cite{hold2019deep} \\
        \includegraphics[width=0.98\linewidth, trim={0 0 0 0},clip]{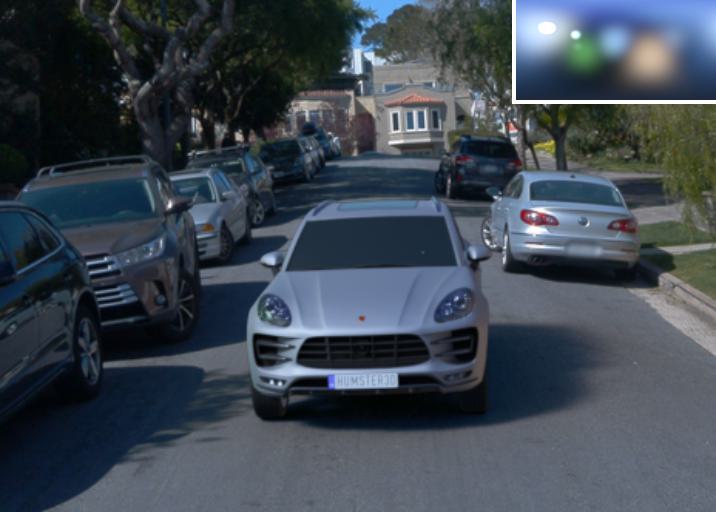} &
        \includegraphics[width=0.98\linewidth, trim={0 0 0 0},clip]{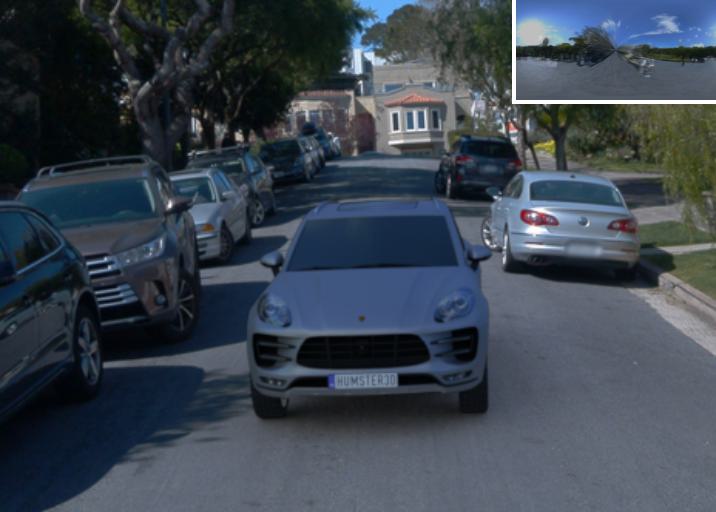} & 
        \includegraphics[width=0.98\linewidth, trim={0 0 0 0},clip]{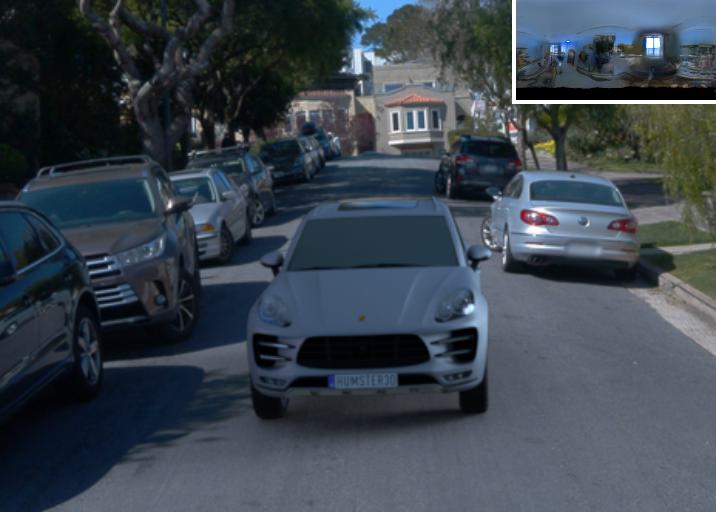} & 
        \includegraphics[width=0.98\linewidth, trim={0 0 0 0},clip]{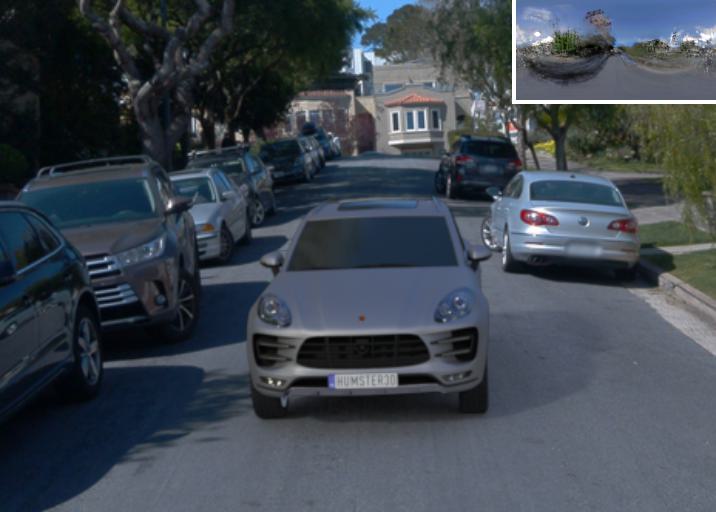} &
        \includegraphics[width=0.98\linewidth, trim={0 0 0 0},clip]{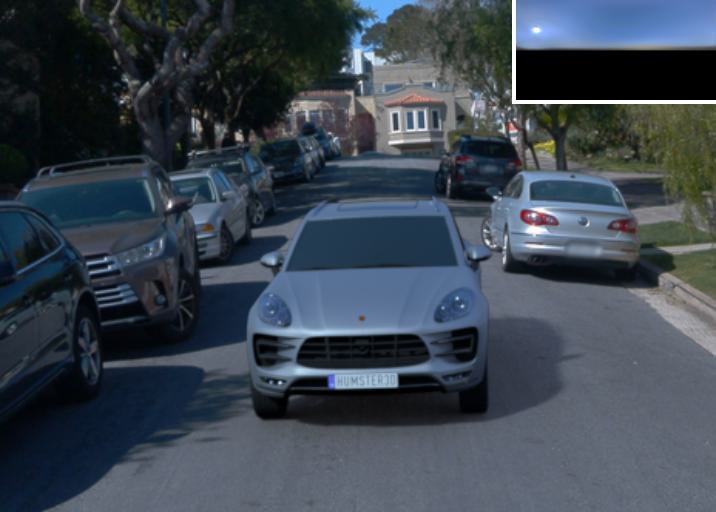}
        \\
        \includegraphics[width=0.98\linewidth, trim={0 0 0 0},clip]{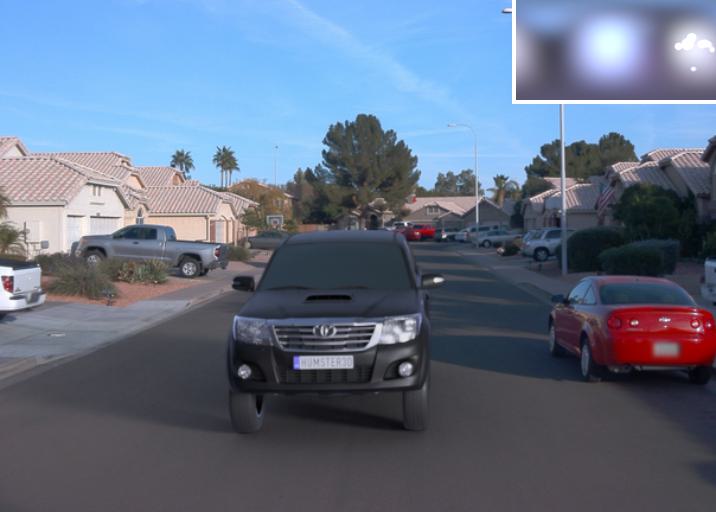} &
        \includegraphics[width=0.98\linewidth, trim={0 0 0 0},clip]{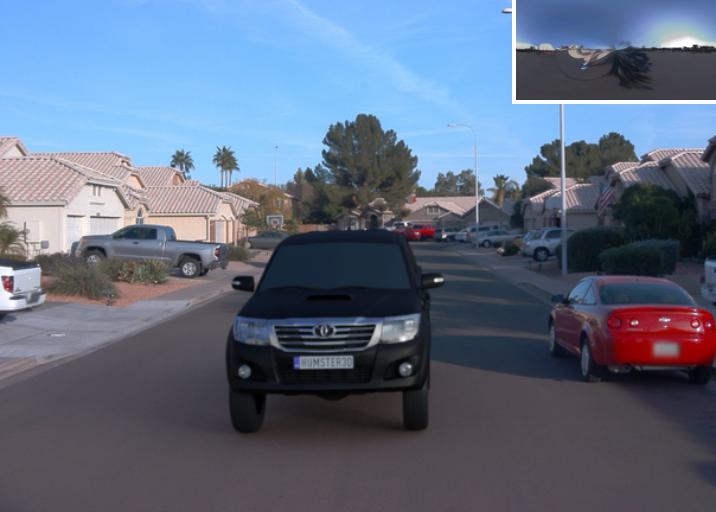} & 
        \includegraphics[width=0.98\linewidth, trim={0 0 0 0},clip]{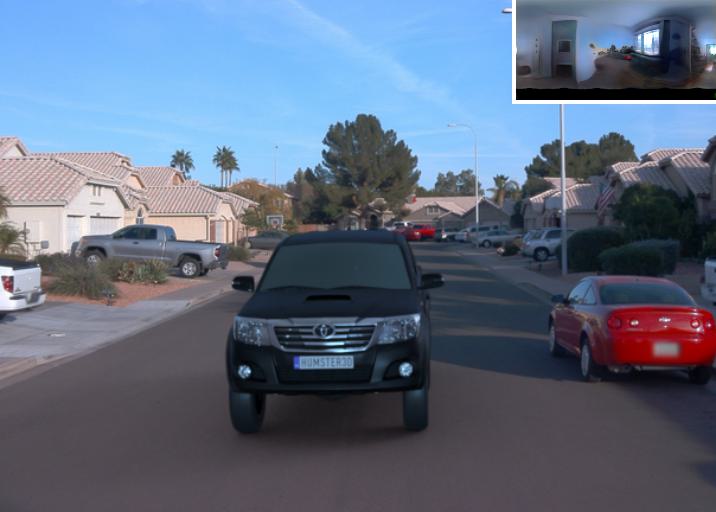} & 
        \includegraphics[width=0.98\linewidth, trim={0 0 0 0},clip]{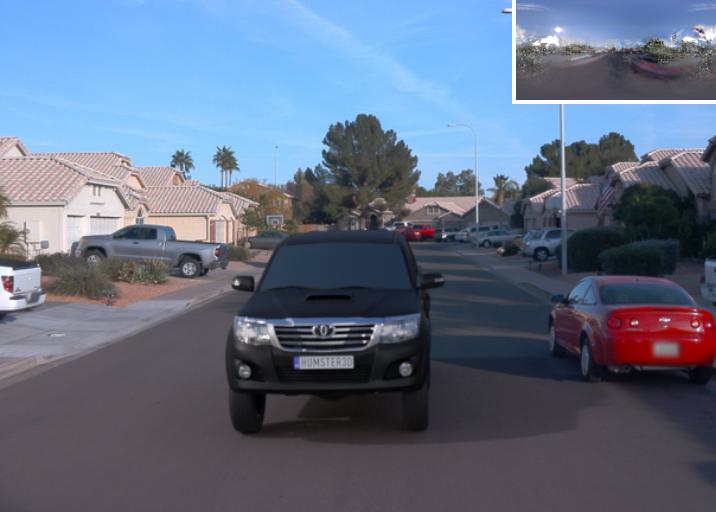} &
        \includegraphics[width=0.98\linewidth, trim={0 0 0 0},clip]{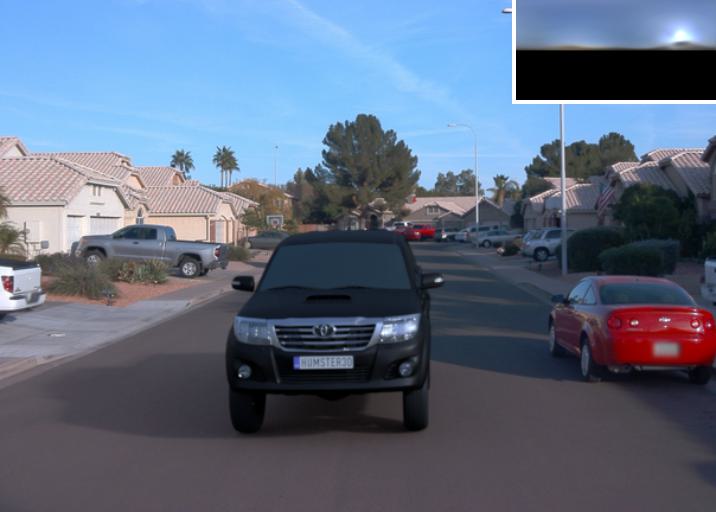}
        \\
        \includegraphics[width=0.98\linewidth, trim={0 0 0 0},clip]{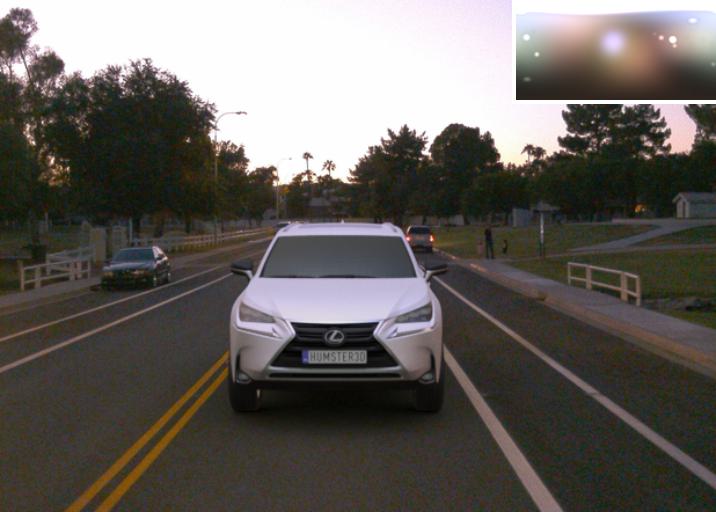} &
        \includegraphics[width=0.98\linewidth, trim={0 0 0 0},clip]{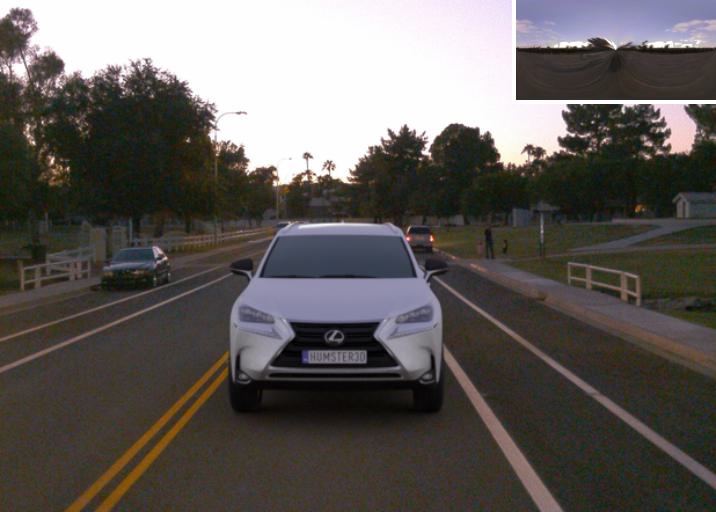} & 
        \includegraphics[width=0.98\linewidth, trim={0 0 0 0},clip]{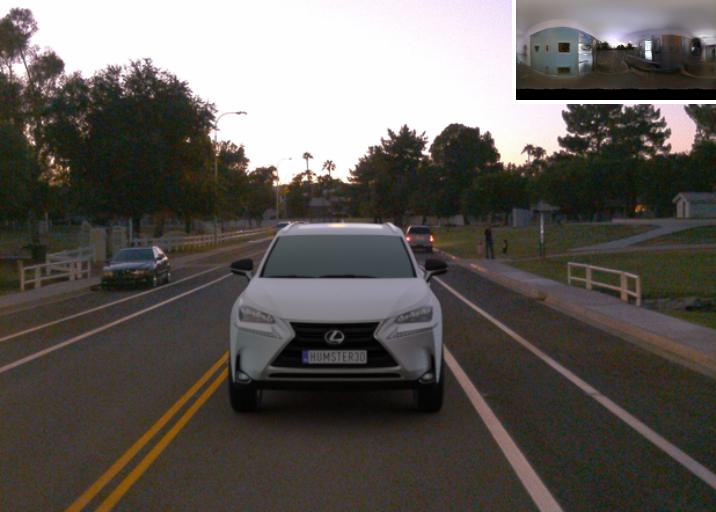} & 
        \includegraphics[width=0.98\linewidth, trim={0 0 0 0},clip]{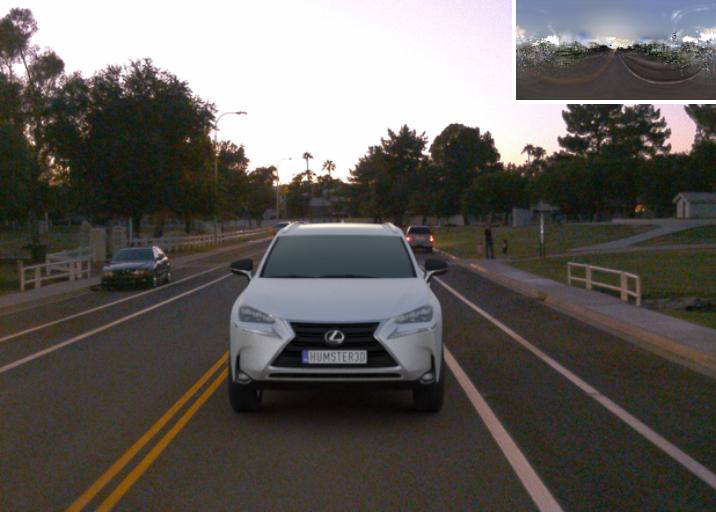} &
        \includegraphics[width=0.98\linewidth, trim={0 0 0 0},clip]{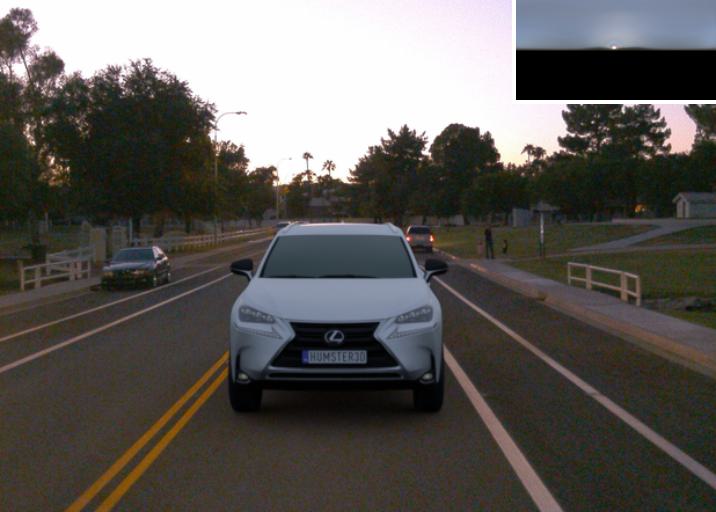}
        \\
        \includegraphics[width=0.98\linewidth, trim={0 0 0 0},clip]{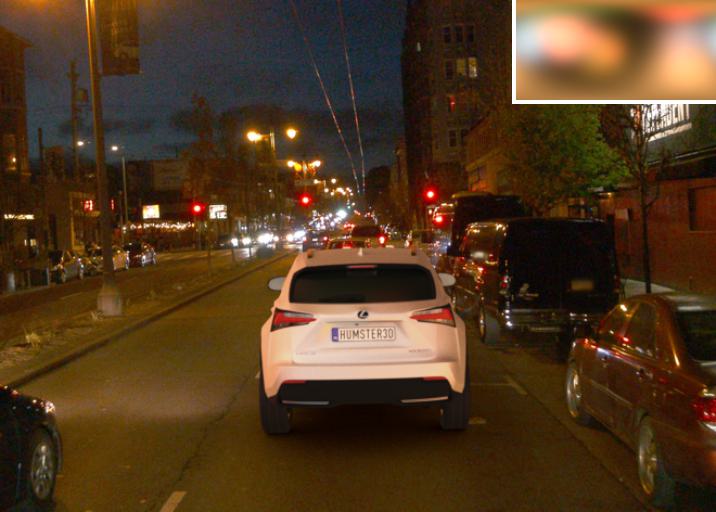} &
        \includegraphics[width=0.98\linewidth, trim={0 0 0 0},clip]{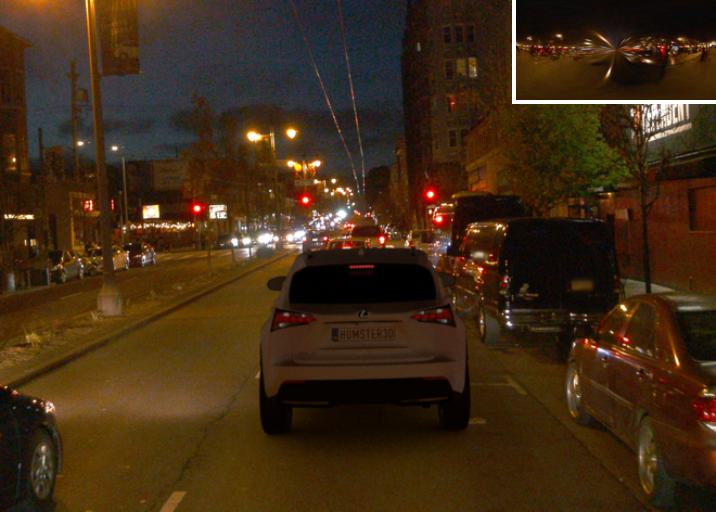} & 
        \includegraphics[width=0.98\linewidth, trim={0 0 0 0},clip]{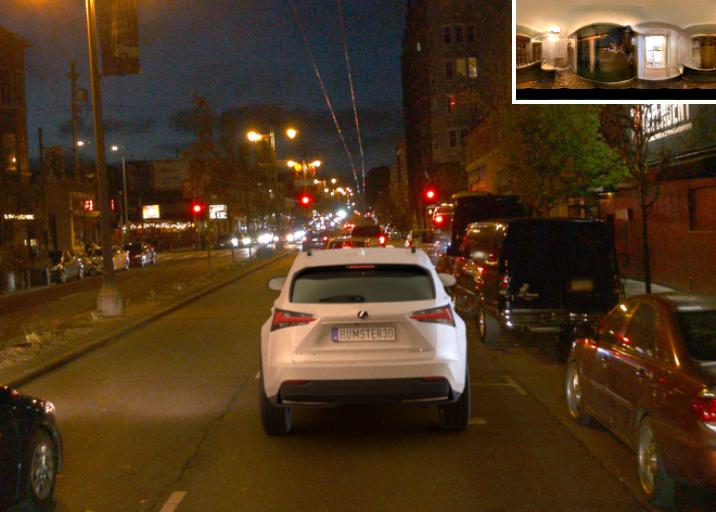} & 
        \includegraphics[width=0.98\linewidth, trim={0 0 0 0},clip]{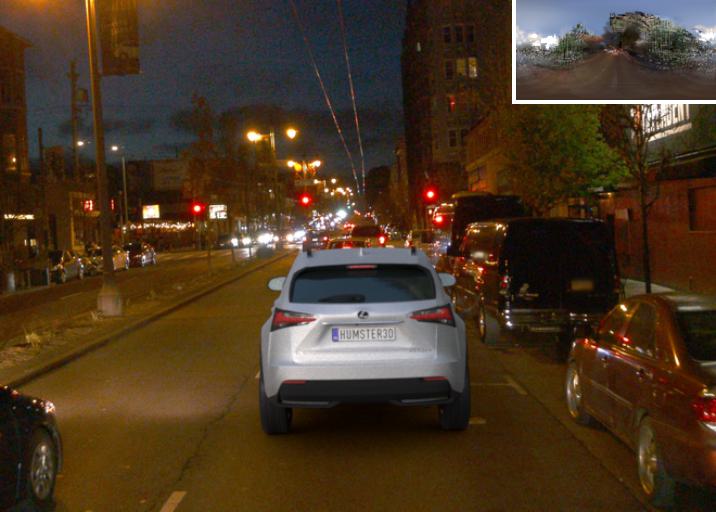} &
        \includegraphics[width=0.98\linewidth, trim={0 0 0 0},clip]{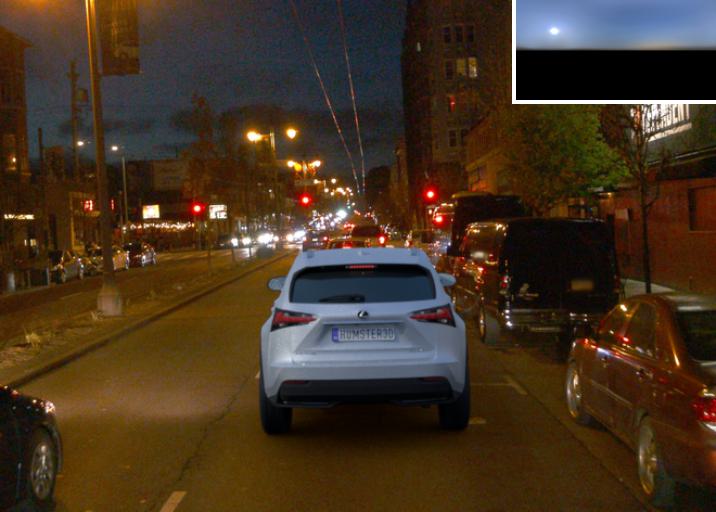}
    \end{tabularx}%
    \makeatletter\def\@captype{figure}\makeatother
    \caption{
        Inserting car assets into Waymo driving scenes.
        Note the direction and sharpness of shadows, as well as overall brightness, color and specular highlights on inserted cars.
    }
    \label{fig:qual_waymo}
    \vspace{-10mm}
\end{table}

\vspace{-3mm}
\subsection{Experiment Settings} 
\label{sec:experiment_settings}
\vspace{3mm} 

\mysubsubsection{Waymo dataset.} 
Following the data split from~\cite{yang2023emernerf}, we use 48 scenes with diverse illumination conditions from the Waymo Open Dataset~\cite{sun2020scalability} including 32 daytime (17 sunny and 15 cloudy), 9 twilight, and 7 night scenes. 
Each scene consists of an input image, a 3D car asset randomly selected from an asset bank, the ground plane, and a location to insert the car. 
Inspired by applications in synthetic data generation, this process is fully automatic: we use the LiDAR point cloud and semantic segmentation~\cite{tao2020hierarchical} to fit a ground plane and detect the empty space to insert a car.

\mysubsubsection{PolyHaven dataset.} 
We use 11 HDR environment maps from PolyHaven~\cite{polyhaven} and manually place a known ground plane and virtual object in each scene.
Each scenario therefore consists of an LDR background image $\mathbf{I}_\text{bg}$ rendered directly from the environment map, the posed virtual object, and the proxy plane. A pseudo-ground-truth rendering is created by lighting the inserted object with the environment map itself.

\mysubsubsection{User study.}
To evaluate the perceptual realism of the virtual object insertion, we conducted a user study where participants received a pair of two object insertion results in a random order -- one generated using our proposed method, the other using a baseline approach.
The participants were then instructed to compare the differences between the two results, inspect the lighting effects of the inserted object, and choose the image they perceived as \textit{more realistic}. 
We invited 9 users to perform a binary selection for each image pair and used majority voting to determine the preferred image for each comparison. In the following, we report the percentage of times that our method was selected over the baseline. This process was repeated three times %
and a preferred percentage > 50\% indicates Ours outperforming baselines. 
We include more details about the user study and statistical evaluation of its results in the Supplement. 

\mysubsubsection{Baselines.} 
On the Waymo dataset of urban street scenes, we compare our method to the representative outdoor lighting estimation methods of Hold-Geoffroy \etal~\cite{hold2019deep} and NLFE~\cite{wang2022neural}; as well as the generative lighting estimation methods StyleLight~\cite{wang2022stylelight} and DiffusionLight~\cite{Phongthawee2023DiffusionLight}. 
On the PolyHaven dataset, we compare with Wang \etal~\cite{wang2021learning}, StyleLight~\cite{wang2022stylelight}, DiffusionLight~\cite{Phongthawee2023DiffusionLight}, and pseudo-ground-truth rendering.

\subsection{Evaluation on Benchmark Datasets}
\label{sec:benchmark}

\vspace{3mm}
\mysubsubsection{Urban street scenes.}
In Table~\ref{table:userstudy_waymo}, we provide quantitative user study results on the Waymo dataset~\cite{sun2020scalability}, separated into 4 subsets based on illumination conditions, as well as on all scenes.
Compared to all baselines, our method outperforms prior state-of-the-art methods (chosen > 50\%), and is preferred in almost every subset.

StyleLight~\cite{wang2022stylelight} is primarily trained on indoor HDR panoramas and suffers from domain gap to outdoor scenes. 
Similarly, the sky model predicted by Hold-Geoffroy~\etal~\cite{hold2019deep} is tailored to outdoor daytime scenes and suffers severely from out-of-domain twilight and night scenes. 
NLFE~\cite{wang2022neural} unprojects the surrounding scene geometry and predicts volumetric lighting, which makes it generalize better to night scenes, but it can fail to estimate an accurate scale and color of the peak for daytime scenes. 
DiffusionLight~\cite{Phongthawee2023DiffusionLight}, a concurrent work, shows impressive results by inpainting a chrome sphere and predicting environment maps with realistic high-frequency details.
However, it often struggles to predict a high-intensity peak in sunny daytime scenes or a proper scale for night scenes. 

A qualitative comparison is shown in Fig.~\ref{fig:qual_waymo}. 
Our method achieves high-quality object insertion, making it a promising approach for applications such as simulating synthetic data, augmented reality navigation and urban planning.

\mysubsubsection{PolyHaven dataset.}
We show the quantitative comparison in Table~\ref{table:userstudy_polyhaven} and qualitative results in Fig.~\ref{fig:qual_polyhaven}. 
Our method is preferred over all baseline methods in the user study, and outperforms baselines on quantitative metrics. 
Our method produces virtual object insertion that naturally and consistently blend in the scenes, making it promising for virtual production tasks. 

\subsection{Ablation Study}
\label{sec:ablation}

We extensively ablate our proposed design choices on diffusion guidance and scene representations, and provide the quantitative ablation in Table~\ref{table:ablation_waymo} with user study and qualitative results in Fig.~\ref{fig:qual_ablation}.  
For the diffusion guidance, we compare with simplified or alternative versions: 
(i) Ours (SDS) removes the adaptive guidance (Eq.~\ref{eq:sds_ours}) and uses the original SDS~\cite{poole2022dreamfusion} formulation (Eq.~\ref{eq:sds_gradient}); (ii) Ours (SDS w/o LoRA) uses the original SDS loss with non-adapted DM; (iii) Ours (w/o concept preservation) uses our LDS loss with adapted DM but not using any class images for concept preservation.
We additionally compare with a baseline of ``dataset update'' which applies a photometric L1 loss on DM repeatedly edited results via SDEdit~\cite{meng2022sdedit}, denoted as Ours (dataset update). 
Although this approach is adopted in text-guided scene editing~\cite{instructnerf2023}, we found it challenging to produce stable gradients from such a ``discrete'' guidance. 

When designing the light and tone representations, we observe that the trainable tone-mapping curve provides additional flexibility to compensate for unknown tone-mapping in the input image, and better match the color and scale of the shadows. The two environment map fusion scheme (Sec.~\ref{sec:two_envmaps_fusion}) allows to robustly recover the high intensity peaks in the early phase of training, and converge to better quality. 
Their respective effect is ablated as ``Ours (w/o tone curve)'' and ``Ours (w/o env. map fusion)''.

\begin{SCtable}[][t]
\captionsetup{font=footnotesize}
    \centering
    \resizebox{0.48\linewidth}{!}{
        \setlength{\tabcolsep}{6pt}
        \begin{tabular}{lc}
            \toprule
            Methods & Ours preferred $\downarrow$ \\
            \midrule 
            Ours (dataset update)           & $85.2$\%    \\ 
            Ours (SDS~\cite{poole2022dreamfusion})  & $74.1$\%    \\ 
            Ours (SDS~\cite{poole2022dreamfusion} w/o LoRA)                               & $90.7$\%    \\ 
            Ours (w/o concept preservation)               & $64.8$\%    \\ 
            Ours (w/o tone curve)                         & $68.5$\%    \\ 
            Ours (w/o env.~map fusion)                    & $66.7$\%    \\ 
            \bottomrule
        \end{tabular}
    }
    \vspace{-4mm}
    \caption{
        \label{table:ablation_waymo} 
        \textbf{Ablation study on outdoor driving scenes~\cite{sun2020scalability}}.
        We report the percentage of images that users preferred \ourmodel compared to its ablated versions.
        Our full pipeline produces results that are preferred more often over its ablated versions.
    }
\end{SCtable} 

\begin{table}[t]
    \vspace{-2mm}
    \centering
    \captionsetup{font=footnotesize}
    \scriptsize
    \setlength\tabcolsep{0pt}
    \begin{tabularx}{\linewidth}%
        {*{4}{>{\centering\arraybackslash}X}}
        \includegraphics[width=0.98\linewidth, trim={0 0 0 0},clip]{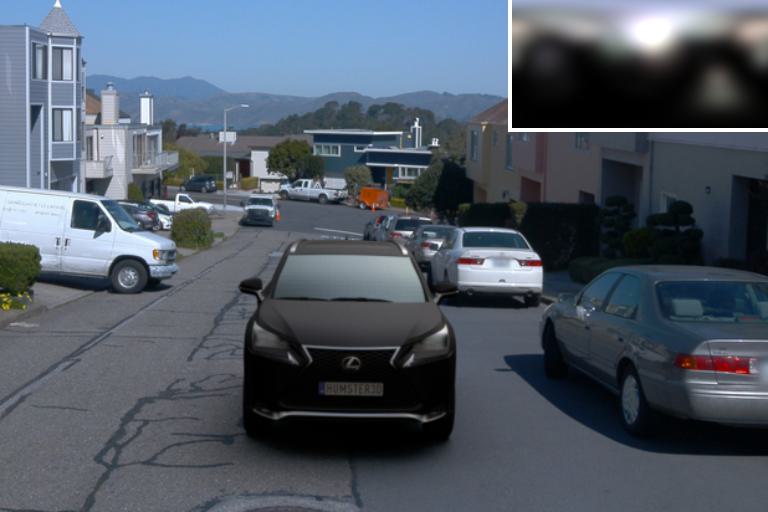} &
        \includegraphics[width=0.98\linewidth, trim={0 0 0 0},clip]{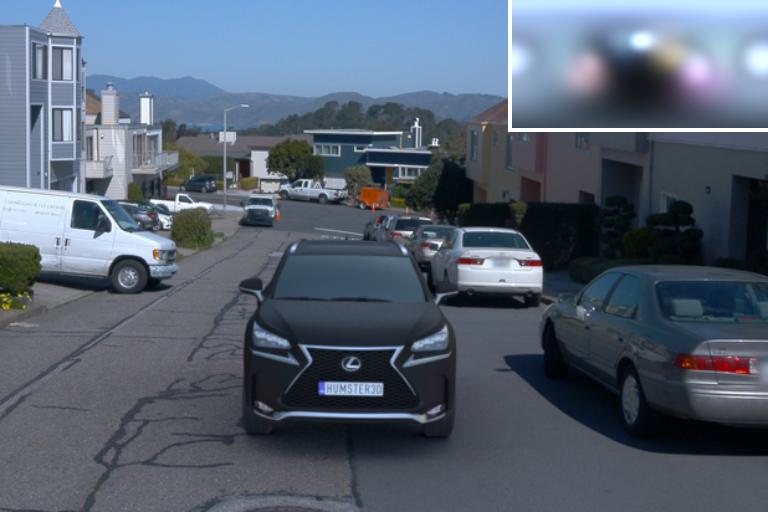} & 
        \includegraphics[width=0.98\linewidth, trim={0 0 0 0},clip]{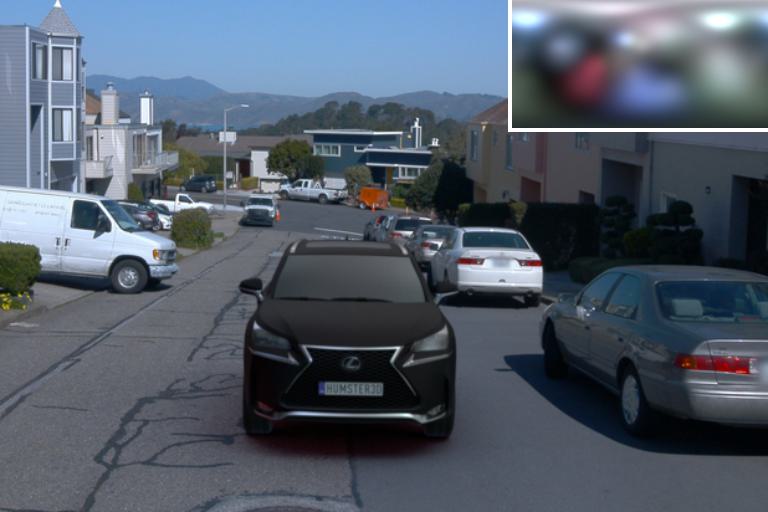} &
        \multirow{3}{*}[0.2\linewidth]{
            \shortstack{
                \includegraphics[width=0.96\linewidth, trim={0 0 0 0},clip]{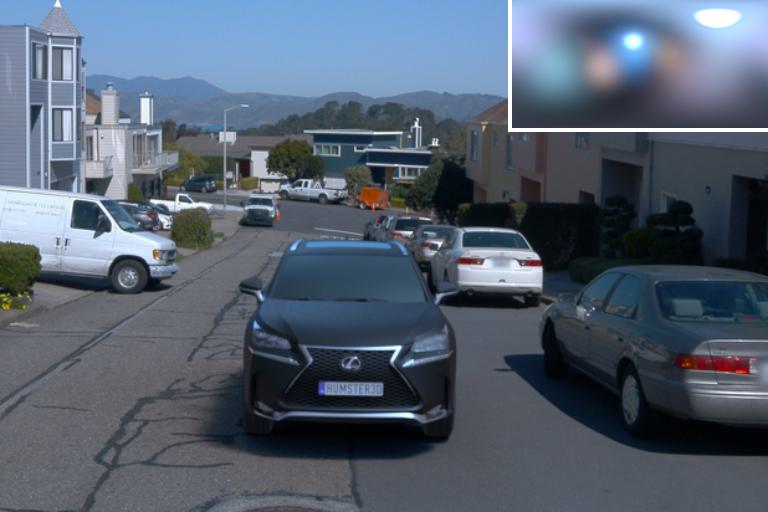} \\
                Ours
            }
        } \\
        Ours (SDS w/o LoRA) & Ours (SDS w/ LoRA) & Ours (dataset update) & \\ 
        \includegraphics[width=0.98\linewidth, trim={0 0 0 0},clip]{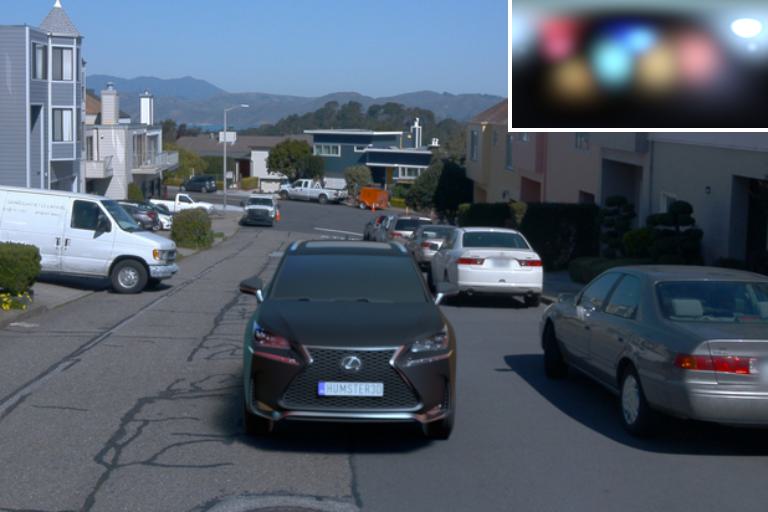} &
        \includegraphics[width=0.98\linewidth, trim={0 0 0 0},clip]{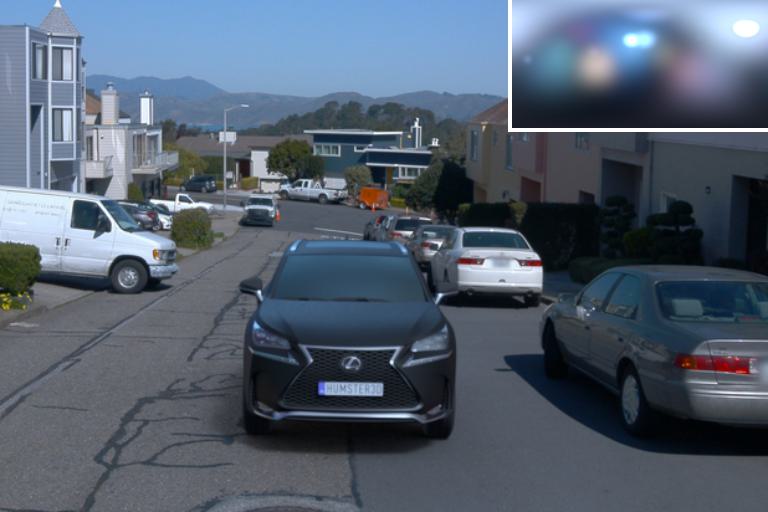} &
        \includegraphics[width=0.98\linewidth, trim={0 0 0 0},clip]{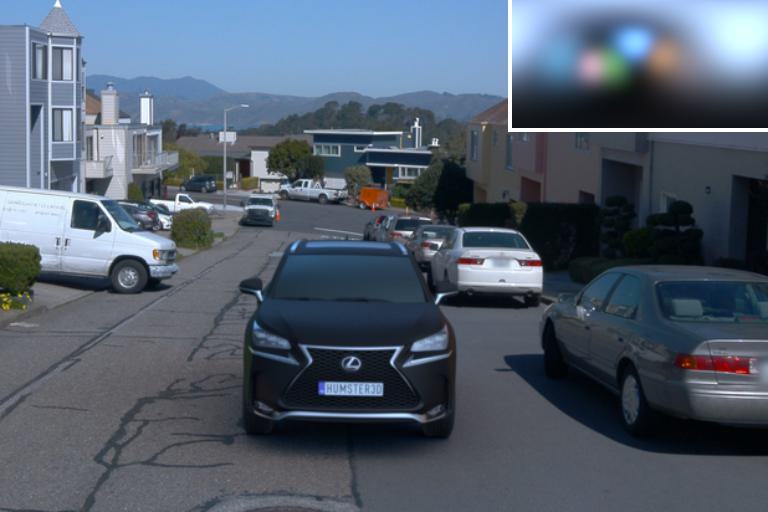} &
        \\
        Ours (w/o concept preservation) & Ours (w/o tone-mapping curve) & Ours (w/o env.~fusion) & 
    \end{tabularx}%
    \makeatletter\def\@captype{figure}\makeatother
    \vspace{5pt}
    \caption{
        Qualitative ablation study of our design choices.
    }
    \label{fig:qual_ablation}
    \vspace{-8mm}
\end{table}

\vspace{-1mm}
\section{Applications}
\label{sec:applications}
\vspace{-2mm}

Since our method recovers physically based lighting information, arbitrary new virtual objects can be inserted after the optimization, as shown in Fig.~\ref{fig:model}.
\ourmodel{} can also optimize other scene attributes such as materials and local lighting.
We perform preliminary experiments in this direction.

\begin{figure*}[t]
\includegraphics[width=\textwidth, trim={0 0 5pt 0},clip]{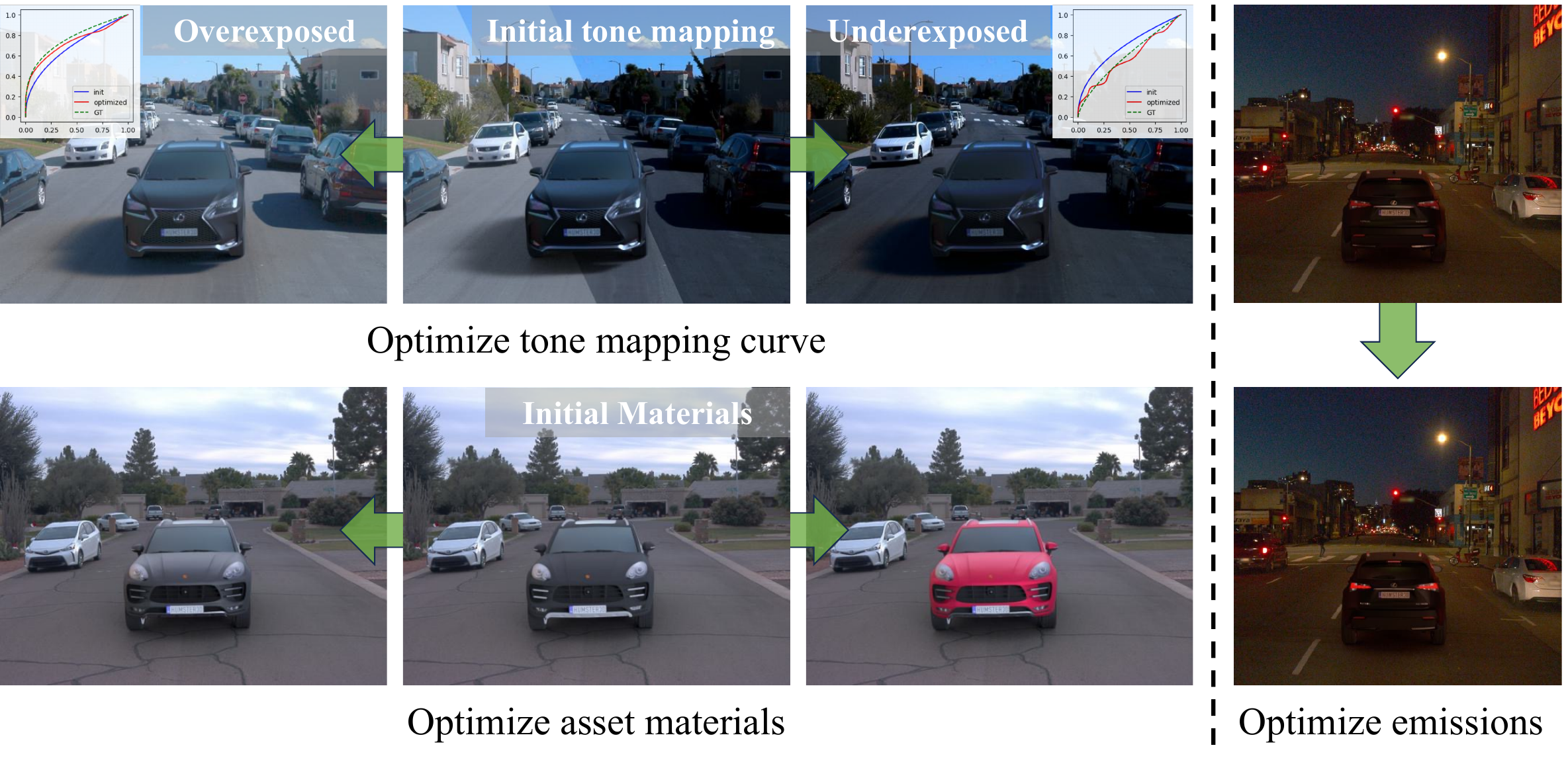}
\vspace{-8mm}
\caption{
    Our physically based inverse rendering pipeline unlocks further applications such as material, local emission and tone-mapping refinement. %
}
\label{fig:qual_application}
\vspace{-3mm}
\end{figure*}

\mysubsubsection{Material optimization.} 
When combined with differentiable rendering, DMs can provide a guidance signal for material attributes, as shown in Fig.~\ref{fig:qual_application}. 
Given a purely diffuse car and enabling Metallic and Roughness properties as optimizable parameters, the diffusion guidance can optimize and make the car look more shiny. 
By additionally changing the text prompt to ``a carmine red car'' and making the base color of the car an optimizable parameter, we show that the DM can propagate the text-condition to the PBR attribute and change the color of the car to red. 
When enabling local emission as an optimizable parameter, the diffusion model can also turn on the headlights of cars in night scenes.

\mysubsubsection{Tone-mapping adjustment.} 
We use a controlled experiment to further evaluate how well DMs understand tone-mapping. 
In Fig.~\ref{fig:qual_application}, we freeze the estimated environment map and apply a manual tone adjustment on the background image. 
The diffusion guidance is used to optimize the tone curve such that the inserted object matches the surrounding background in the final composited result.

\vspace{-3mm}
\section{Discussion}
\label{sec:conclusion}
\vspace{-2mm}

Our method leverages large diffusion models' inherent scene understanding capabilities as guidance to a physically based inverse rendering pipeline. 
We design a diffusion guidance signal with scene-specific personalization and a differentiable inverse rendering pipeline to recover lighting and tone-mapping parameters. 
Our method enables inserting virtual objects into scenes, but also optimizing other scene parameters such as the materials of the inserted object or account for tone-mapping mismatches between cameras.
We believe that this combination of the differentiable rendering process and data-driven priors can be used successfully in many other content creation applications such as relighting and animation.

\mysubsubsection{Limitations and future work.} \label{sec:limitations}
Our Spherical Gaussians-based lighting representation is adequate for general objects~\cite{li2020inverse}, but might not behave realistically for highly specular materials. For more complex lighting representations adding generative priors on the environment map~\cite{lyu2023dpi} is a direction worth exploring. 
The rendering formulation could be extended to account for effects such as reflections from the scene itself onto the inserted object (e.g. color bleeding), but that might introduce more ambiguities and require knowing the materials of the proxy geometry
(refer to Supplement C.4 for failure case examples).
Finally, while DM personalization significantly improves the quality of the results, it adds overhead and complexity to the pipeline. Recent personalization methods that do not require test-time finetuning~\cite{shi2023instantbooth} could be used to mitigate this overhead.

\section*{Acknowledgements} 
The authors are grateful for the feedback received from Nicholas Sharp and Huan Ling during the project. We thank the original artists of the 3D assets used in this work: inciprocal, peyman.khaleghi, Kuutti Siitonen, TurboSquid, and their artists Hum3D and Amaranthus.

\bibliographystyle{splncs04}
\bibliography{main}

\newpage

\appendix
\section*{Supplement}

In the supplement, we provide additional ablation for diffusion guidance (Sec.~\ref{sec:supp_diffusion}), implementation details (Sec.~\ref{sec:implementation_details}), additional results on user study and tone-mapping (Sec.~\ref{sec:additional_results}), and discuss broader impact (Sec.~\ref{sec:supp_discussion}).
Please refer to the \textit{accompanied video} for more qualitative results.

\section{Diffusion Personalization and Score Distillation}
\label{sec:supp_diffusion}

An intuitive approach to understanding the diffusion guidance is to directly visualize the text-to-image generation result of the diffusion model. 
In this section, we provide additional analysis and ablative visualization on our design choices of LoRA personalization (Fig.~\ref{fig:difffusion_bg}) and concept preservation (Fig.~\ref{fig:diffusion_lora}). 

\subsubsection{Personalizing diffusion model.}
\begin{figure*}[h!]
    \vspace{3mm}
    \centering
    \includegraphics[width=\textwidth]{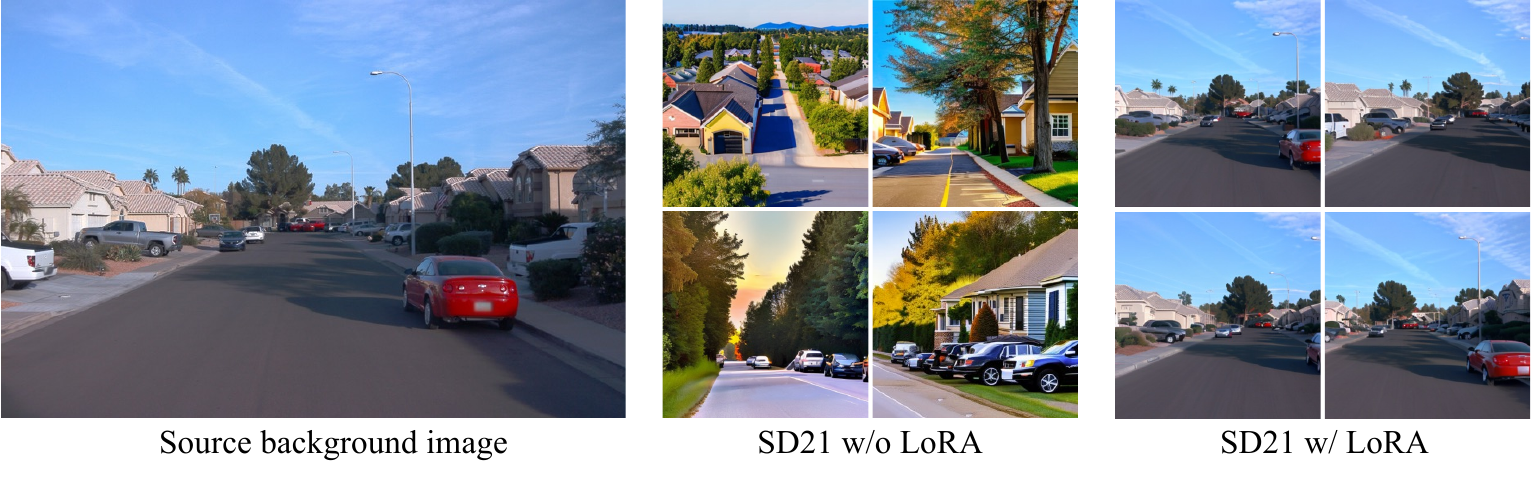}
    \captionsetup{font=footnotesize}
    \vspace{-6mm}
    \caption{
        Illustrations of text-to-image generation results. ``SD21 w/o LoRA'' shows our best-effort prompting results for outdoor street scenes from off-the-shelf Stable Diffusion 2.1. ``SD21 w/ LoRA'' directly uses the training prompt \textit{``a scene in the style of sks rendering''}.
        LoRA personalization enables generating images in a similar domain to the target scene. 
    }
    \label{fig:difffusion_bg}
    \vspace{-4mm}
\end{figure*}

Due to the high stochasticity in the diffusion denoising process, the images generated by a pre-trained diffusion model often cannot be tailored to a specific input image. 
However, in the setting of using the diffusion model for solving the inverse rendering problem of the given scene, it is important to preserve the key context (e.g., shapes, lighting, and shadowing effects) from the unseen input background image.
In Fig.~\ref{fig:difffusion_bg}, we visualize the effect of LoRA personalization. 
After personalization, the diffusion model can generate images in a similar domain to the target scene.

\subsubsection{Concept preservation.}
For the task of virtual object insertion, it is important to ensure the personalized diffusion model does not completely overfit the input background image, and can generalize when inserting new objects into the scene through additional text conditions.
Personalizing diffusion model (DM) along with some generated class images is inspired by the original DreamBooth paper~\cite{ruiz2022dreambooth} which shows that adding in-class images for concept preservation can avoid concept drift and improve the output diversity. 

\begin{figure*}[t!]
    \centering
    \includegraphics[width=\textwidth]{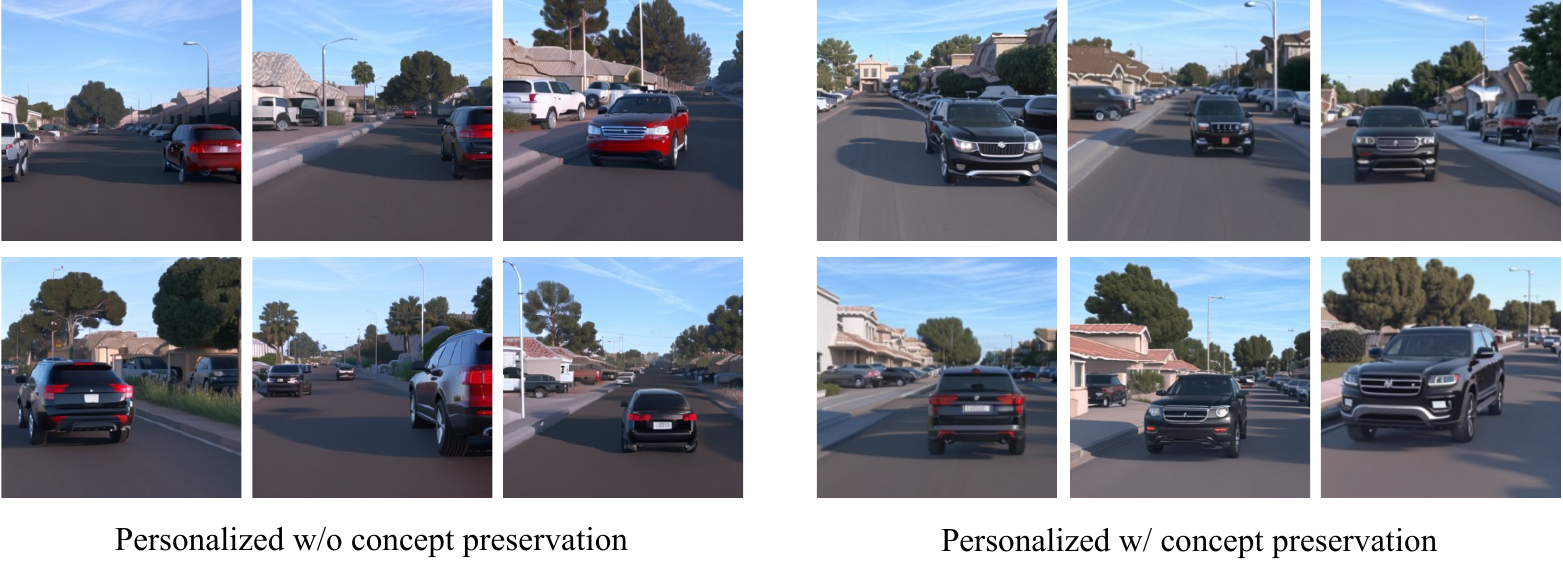}
    \captionsetup{font=footnotesize}
    \vspace{-6mm}
    \caption{
        Text-to-image generation results with prompt \textit{``\textbf{a black SUV car} in the style of sks rendering''}. Concept preservation can facilitate the high-quality generation of both the object and the background scene. 
    }
    \label{fig:diffusion_lora}
    \vspace{-6mm}
\end{figure*}

In Fig.~\ref{fig:diffusion_lora}, we visualize the text-to-image generation results with and without using concept preservation. 
The results show that only personalizing with the input image does not generalize well when adding additional concepts into the text prompt -- Diffusion model does not faithfully follow the additional prompt to synthesize images with ``a black SUV car'' in it. The reddish car color is heavily affected by the car shown in the input background image. 
The results of concept preservation show the benefits of retaining the appearance of newly inserted objects. 

\vspace{-3mm}
\subsubsection{LDS loss design.} 
The original SDS loss tends to generate over-saturated images. Recent work~\cite{yu2023text, ling2023alignyourgaussians} observes that the classifier score $\bm\delta = {\rvepsilon}_{\dmparam}(\vz_t,t,\vc) - {\rvepsilon}_{\dmparam}(\vz_t,t,\varnothing)$ dominates the optimization direction, and directly distilling the classifier score can provide much better quality. 
Our LDS loss  (Eq. \ref{eq:sds_ours}) is inspired by the classifier score distillation and adapts it to the personalized diffusion model.
The delta term in LDS loss is calculated between LoRA fine-tuned conditional denoising term ${\rvepsilon}_{(\dmparam + \bm{\Delta W})}(\vz_t,t,\vc)$ and non-adapted unconditional denoising term ${\rvepsilon}_{\dmparam}(\vz_t,t,\varnothing)$. 
The intuition of not ``using the LoRA fine-tuned model in both terms'' is to encourage gradient towards the personalized model with scene-specific knowledge, while not overly biased by the small amount of training data used in personalization. 
We empirically observe our LDS loss is more stable and leads to better quality. We ablate this design choice in Fig.~\ref{fig:ablation1} and defer rigorous theoretical understanding of this loss to further work. 

\begin{table}[h!]
    \vspace{3mm}
    \centering
    \scriptsize
    \setlength\tabcolsep{0pt}
    \begin{tabularx}{\linewidth}%
        {*{2}{>{\centering\arraybackslash}X}}
        \includegraphics[width=0.98\linewidth, trim={0 0 0 0},clip]{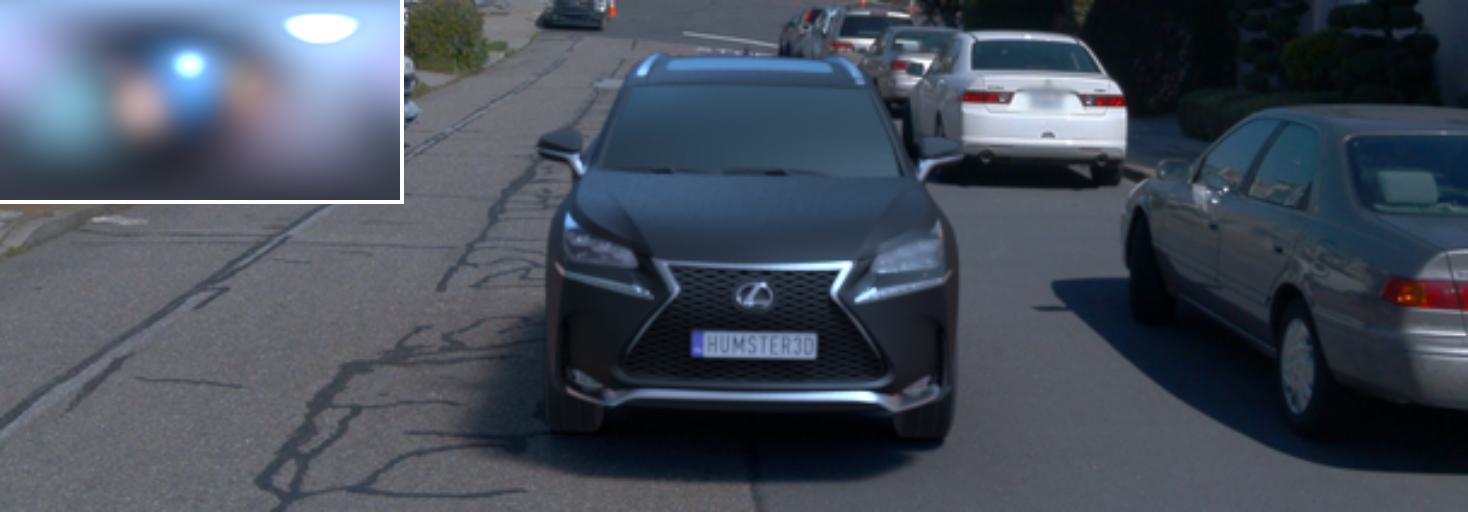} &
        \includegraphics[width=0.98\linewidth, trim={0 0 0 0},clip]{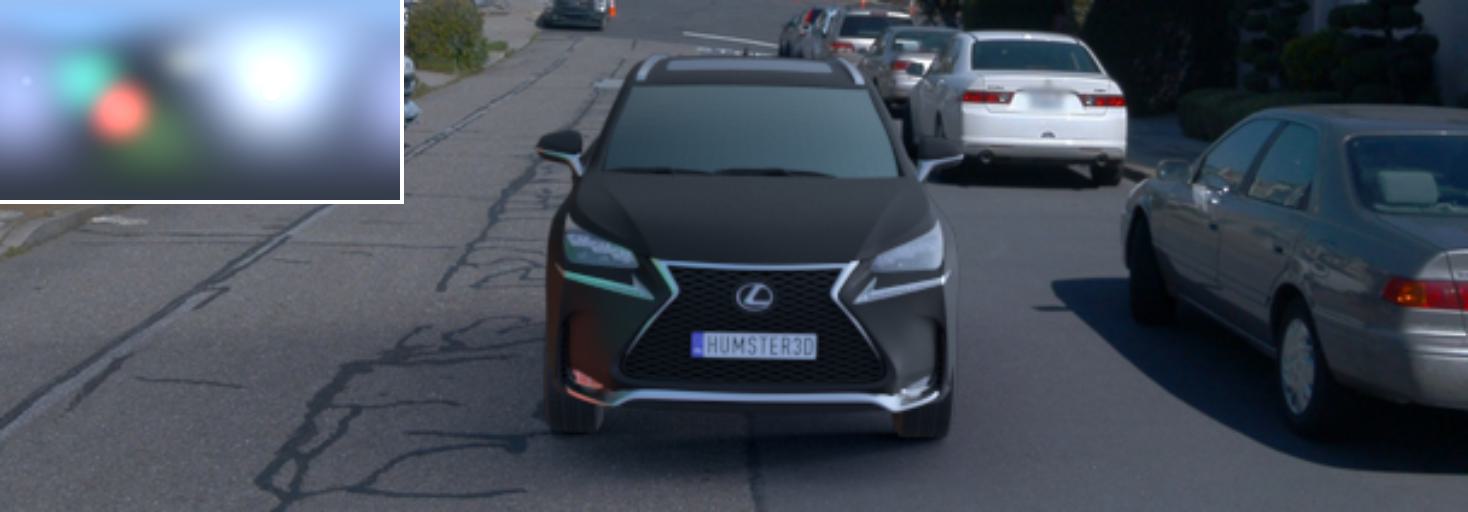} \\
        \textbf{Ours}: ${\rvepsilon}_{(\dmparam + \bm{\Delta W})}(\vc) - {\rvepsilon}_{\dmparam}(\varnothing)$ & 
        ${\rvepsilon}_{(\dmparam + \bm{\Delta W})}(\vc) - {\rvepsilon}_{(\dmparam + \bm{\Delta W})}(\varnothing)$ \\ 
    \end{tabularx}%
    \makeatletter\def\@captype{figure}\makeatother
    \vspace{2mm}
    \caption{
        Ablation on unconditional denoising term in LDS loss.
    }
    \label{fig:ablation1}
\end{table}

\section{Implementation Details} 
\label{sec:implementation_details}

\subsubsection{Diffusion Model}
We use Stable Diffusion 2.1 as our pre-trained diffusion models throughout experiments.  To reduce memory overhead and accelerate the training, we use PyTorch's FP16 mixed floating point training for diffusion models by default. The whole optimization can be run on a GPU with more than 12GB VRAM.

\subsubsection{Rendering and image formation.} 
The differentiable rendering framework is built on Mitsuba 3~\cite{jakob2022mitsuba3}. We use 128 samples per pixel, and spawn 4 rays each for multiple importance sampling (MIS)~\cite{black1996robust} of BSDF and emitters. The output resolution is $256\times384$, which we crop and bilinearly upsample to $512\times512$ to feed into the personalized diffusion model. 

The tone-mapping function for the input image is often unknown, and thus we use the default Reinhard tone-mapping~\cite{reinhard} for the inserted virtual object $\mathbf{I}_\text{fg} = \text{Reinhard}(\mathbf{I}_\text{HDR})$. 
As described in the main paper, the rendered pixels are then passed into the single-channel optimizable tone correction function $\tilde{\mathbf{I}}_\text{fg} = f(\mathbf{I}_\text{fg}; \bm{\theta}_\text{fg})$. The tone correction function $f(\cdot)$ is an optimizable spline curve that differentiably maps real values from the range $[0, 1]$ to $[0, 1]$, which aims to learn the residual of the default Reinhard tone-mapping. 
The shadow ratio is directly multiplied onto the tone-mapped input image, and thus we do not apply additional tone-mapping and directly pass it into the tone correction function $\tilde{\bm{\beta}}_\text{shadow} = f(\bm{\beta}_\text{shadow}; \bm{\theta}_\text{shadow})$. 
All notations in the main paper and supplement operate in linear RGB space following graphics conventions, and we finally convert with gamma correction ($\gamma=2.2$) to produce sRGB output.

\subsubsection{Environment map fusion.} 
Following the description in the main paper, we initialize two sets of optimizable Spherical Gaussian (SG) parameters and compute two separate environment maps, $\mathbf{L}_\text{fg}, \mathbf{L}_\text{shadow}\in \mathbb{R}^{H\times W \times 3}$, to light the foreground inserted object and cast shadows respectively. 

The additional capacity can improve quality and stabilize the training in the early stage of optimization, and we aim to progressively fuse them into a single environment map at the end of optimization. 
Let $\tilde{\mathbf{L}}_\text{*} \in \mathbb{R}^{H\times W}$ denote the luminance of each environment map, 
we compute the fused environment map $\mathbf{L}_\text{fused}\in \mathbb{R}^{H\times W \times 3}$ by adjusting the luminance of the foreground environment map: 
\begin{equation}
  \mathbf{L}_\text{fused} = \mathbf{L}_\text{fg} \cdot \frac{\tilde{\mathbf{L}}_\text{fused}}{\tilde{\mathbf{L}}_\text{fg}}
\end{equation}
where the target luminance of the fused environment map is computed by blending the two environment maps: 
\begin{equation}
\tilde{\mathbf{L}}_\text{fused} = (1-\mathbf{r}) \cdot \tilde{\mathbf{L}}_\text{fg} + \mathbf{r} \cdot \tilde{\mathbf{L}}_\text{shadow}
\end{equation}
\begin{equation}
\mathbf{r} = \frac{\tilde{\mathbf{L}}_\text{fg}}{\max(\tilde{\mathbf{L}}_\text{fg})} \cdot \frac{\tilde{\mathbf{L}}_\text{shadow}}{\tilde{\mathbf{L}}_\text{fg} + \tilde{\mathbf{L}}_\text{shadow}}.
\end{equation}
Here $\mathbf{r} \in \mathbb{R}^{H\times W}$ is the per-pixel blending ratio which encourages the fused luminance to favor $\mathbf{L}_\text{shadow}$ at high luminance pixels and $\mathbf{L}_\text{fg}$ at low luminance pixels.

As the optimization progresses, the fused environment map $\mathbf{L}_\text{fused}$ is linearly scheduled to replace the two environment maps $\mathbf{L}_\text{fg}, \mathbf{L}_\text{shadow}$ for the rendering of $\mathbf{I}_\text{fg}$ and $\bm{\beta}_\text{shadow}$: 
\begin{align}
  \mathbf{I}_\text{fg} &= \text{PathTrace}(\mathcal{X}, \mathbf{L}_\text{fg}^{'}, D) \\
  \bm{\beta}_\text{shadow} &= \frac{\text{PathTrace}(\mathcal{X} \cup \mathcal{P}, \mathbf{L}_\text{shadow}^{'}, 1)}{\text{PathTrace}(\mathcal{P}, \mathbf{L}_\text{shadow}^{'}, 1)} 
\end{align}
where the scalar value $s$ is scheduled to linearly increase from $0$ to $1$: 
\begin{align}
  \mathbf{L}_\text{fg}^{'} &= s \cdot \mathbf{L}_\text{fused} + (1 - s) \cdot \mathbf{L}_\text{fg} \\
  \mathbf{L}_\text{shadow}^{'} &= s \cdot \mathbf{L}_\text{fused} + (1 - s) \cdot \mathbf{L}_\text{shadow}
\end{align}
such that $\mathbf{L}_\text{fg}^{'} = \mathbf{L}_\text{shadow}^{'} = \mathbf{L}_\text{fused}$ at the end of optimization.

\subsubsection{3D assets.} 
We use 6 licensed 3D car models from Turbosquid and 3DModels.org for experiments on Waymo outdoor street scenes, and 11 assets from Sketchfab and PolyHaven for PolyHaven HDRI scenes.

\subsubsection{Running time overhead.}
The total running time is about 26 min ($\sim$13 min for LoRA DM finetuning + $\sim$13 min for distillation sampling) on an RTX A6000 GPU with FP16 mixed precision inference.

\section{Additional Results}
\label{sec:additional_results}

In this section, we provide further experimental details and additional results.

\subsection{User Study} 
\label{sec:supp_user_study} 
User study is a standard approach for assessing perceptual realism of virtual object insertion~\cite{gardner2019deep,garon2019fast,gardner2017learning,wang2022neural,wang2023fegr}. 
Following prior works, we conduct a user study on Amazon Mechanical Turk to compare with prior methods and ablate our design choices.

\subsubsection{User interface.} 
\begin{figure*}[t!]
    \centering
     \includegraphics[width=0.99\textwidth]{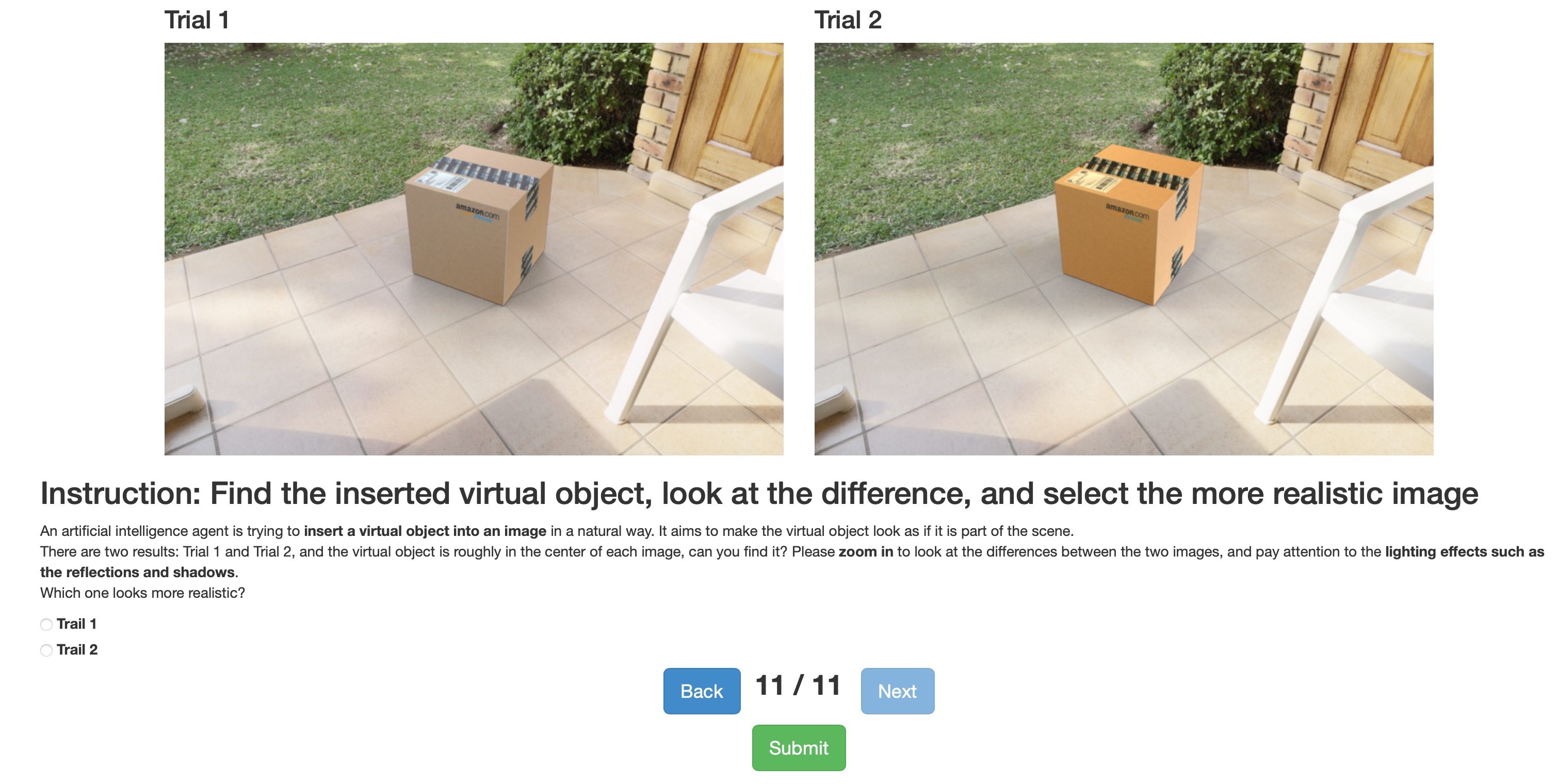}
    \caption{
        Visualization of interface for user study. 
    }
    \label{fig:userstudy_viz} 
\end{figure*}

Participants receive a pair of two object insertion results: one generated using our proposed method, the other one using a baseline approach. 
Participants are instructed to evaluate the differences between the two images, focus on the lighting effects of the inserted objects, and select the image they deemed to be \textit{more realistic}: 
\begin{displayquote}
An artificial intelligence agent is trying to insert a virtual object into an image in a natural way.
It aims to make the virtual object look as if it is part of the scene.
There are two results: Trial 1 and Trial 2, and the virtual object is roughly in the center of each image, can you find it?
Please zoom in to look at the differences between the two images, and pay attention to the lighting effects such as the reflections and shadows.
Which one looks more realistic?
\end{displayquote}
The participants are required to use a 24-inch or larger monitor to view the results, and the images are shuffled in a random order to prevent bias. The user interface is visualized in Fig.~\ref{fig:userstudy_viz}. 

\subsubsection{Statistics.} 
We invited 9 different users for each experiment setting, and repeated each experiment 3 times. 
For benchmarking experiments, there are 48 scenes on the Waymo dataset to compare with 4 baselines, and 11 scenes on the PolyHaven dataset to compare with 3 baselines. 
This results in a number of $48\times4\times9\times3=5184$ and $11\times3\times9\times3=891$ user selections for each dataset. 
For the ablation study, we randomly select a subset of 18 scenes from the Waymo dataset to reduce cost, and compare with 6 ablated versions of our method. The number of user selections for the ablation study is $18\times6\times9\times3=2916$. The total number of user selections for all experiments is $8991$.

\subsubsection{Metrics and additional results.} 

Our primary evaluation metric is the percentage of \textit{images} that our method was preferred over the baseline, following~\cite{wang2022neural}. 
Specifically, for each sample, we collect the binary selection from 9 different users and do majority voting from the 9 users to determine which method is more preferred on this \textit{sample}. The majority voting can efficiently filter the effects of random users, and we report this as the primary metric in the main paper. 
The full experiments are repeated three times to calculate the mean and standard deviation. We additionally report the standard deviation in Table~\ref{table:userstudy_waymo_supp1}, \ref{table:userstudy_polyhaven_supp}, \ref{table:ablation_waymo_supp}. 
Note that the standard deviation reflects the consistency in user evaluations \textit{after} majority vote, where a high standard deviation suggests the compared methods performed on par on some of the examples. 

We also report the percentage of \textit{user selections} that our method is preferred over the baselines in Table~\ref{table:userstudy_waymo_supp1}, \ref{table:userstudy_polyhaven_supp}, \ref{table:ablation_waymo_supp}. Our method consistently outperforms baseline methods and ablated versions of our method.

\begin{table}[t!]
    \centering
    \caption{\textbf{User study: benchmark on Waymo outdoor street scenes. }
        We report the percentage of \textit{images} and \textit{user selections} that our method is preferred over baselines.
        A preferred percentage > 50\% indicates Ours outperforming baselines. 
    } 
    \vspace{-3mm}
    \label{table:userstudy_waymo_supp1} 
    \resizebox{0.99\linewidth}{!}{
        \addtolength{\tabcolsep}{6pt}
        \begin{tabular}{lccccc}
            \toprule
            \multirow{2}{*}{\% images Ours is preferred} &  \multicolumn{2}{c}{Daytime} & \multirow{2}{*}{Twilight} & \multirow{2}{*}{Night} & \multirow{2}{*}{All scenes} \\
             &  Sunny &  Cloudy &  &  &  \\
            \midrule
            DiffusionLight~\cite{Phongthawee2023DiffusionLight}  & $80.4 \pm 12.2$\%  & $68.9 \pm 20.4$\%  & $55.6 \pm 11.1$\%  & $71.4 \pm 14.3$\% & $70.8 \pm  3.6$\%   \\ 
            Hold-Geoffroy \etal~\cite{hold2019deep}              & $60.8 \pm  6.8$\%  & $66.7 \pm 13.3$\%  & $74.1 \pm 25.7$\%  & $85.7 \pm 24.7$\% & $68.8 \pm  2.1$\%   \\ 
            NLFE~\cite{wang2022neural}                           & $80.4 \pm  6.8$\%  & $73.3 \pm 11.5$\%  & $44.4 \pm 11.1$\%  & $52.4 \pm 21.8$\% & $67.4 \pm  3.2$\%   \\ 
            StyleLight~\cite{wang2022stylelight}                 & $76.5 \pm 15.6$\%  & $91.1 \pm 10.2$\%  & $66.7 \pm 22.2$\%  & $66.7 \pm  8.2$\% & $77.8 \pm 12.6$\%   \\ 
            \bottomrule
        \end{tabular}
    }
    \resizebox{0.99\linewidth}{!}{
        \addtolength{\tabcolsep}{6pt}
        \begin{tabular}{lccccc}
            \toprule
            \multirow{2}{*}{\% user selection Ours is preferred} &  \multicolumn{2}{c}{Daytime} & \multirow{2}{*}{Twilight} & \multirow{2}{*}{Night} & \multirow{2}{*}{All scenes} \\
             &  Sunny &  Cloudy &  &  &  \\
            \midrule
            DiffusionLight~\cite{Phongthawee2023DiffusionLight}  & $63.4 \pm 4.0$\%  & $61.2 \pm 7.6$\%  & $53.9 \pm 8.7$\%  & $59.3 \pm 8.1$\% & $60.3 \pm 1.9$\%   \\ 
            Hold-Geoffroy \etal~\cite{hold2019deep}              & $56.4 \pm 1.6$\%  & $54.1 \pm 7.5$\%  & $61.7 \pm 9.8$\%  & $68.8 \pm 12.1$\% & $58.5 \pm 1.9$\%   \\ 
            NLFE~\cite{wang2022neural}                           & $65.4 \pm 1.1$\%  & $58.3 \pm 4.1$\%  & $50.6 \pm 1.2$\%  & $56.6 \pm 2.4$\% & $59.1 \pm 1.2$\%   \\ 
            StyleLight~\cite{wang2022stylelight}                 & $60.6 \pm 8.2$\%  & $68.9 \pm 8.5$\%  & $61.7 \pm 6.5$\%  & $59.8 \pm 10.3$\% & $63.3 \pm 6.0$\%   \\ 
            \bottomrule
        \end{tabular}
    }
\end{table}

\begin{table}[t!]
    \centering
    \caption{\textbf{User study: benchmark on PolyHaven scenes. }
        We report the percentage of \textit{images} and \textit{user selections} that our method is preferred over baselines.
        A preferred percentage > 50\% indicates Ours outperforming baselines. 
    } 
    \vspace{-3mm}
    \label{table:userstudy_polyhaven_supp} 
    \resizebox{0.85\linewidth}{!}{
        \addtolength{\tabcolsep}{6pt}
        \begin{tabular}{lcc}
            \toprule
            Methods & \% images Ours is preferred & \% user selection Ours is preferred  \\
            \midrule
            DiffusionLight~\cite{Phongthawee2023DiffusionLight}  & $66.7 \pm 5.2$\%  & $57.2 \pm 0.6$\%   \\  
            Wang~\etal~\cite{wang2021learning}                    & $84.8 \pm 18.9$\%  & $66.3 \pm 1.5$\%   \\       
            StyleLight~\cite{wang2022stylelight}                 & $75.8 \pm 5.2$\%  & $60.6 \pm 5.2$\%   \\  
            \bottomrule
        \end{tabular}
    }
\end{table}

\begin{table}[t!]
    \centering
    \caption{\textbf{User study: ablation study on Waymo outdoor street scenes. } 
        We report the percentage of \textit{images} and \textit{user selections} that our method is preferred over baselines.
        A preferred percentage > 50\% indicates Ours outperforming the ablated versions. 
    } 
    \vspace{-3mm}
    \label{table:ablation_waymo_supp} 
    \resizebox{0.95\linewidth}{!}{
        \addtolength{\tabcolsep}{6pt}
        \begin{tabular}{lcc}
            \toprule
            Methods & \% images Ours is preferred & \% user selection Ours is preferred  \\
            \midrule
            Ours (dataset update)                             & $85.2 \pm 25.7$\%  & $67.9 \pm 10.2$\%    \\ 
            Ours (SDS~\cite{poole2022dreamfusion})            & $74.1 \pm 12.8$\%  & $64.0 \pm 5.0$\%    \\ 
            Ours (SDS~\cite{poole2022dreamfusion} w/o LoRA)   & $90.7 \pm 8.5$\%  & $71.6 \pm 9.6$\%    \\ 
            Ours (w/o concept preservation)                   & $64.8 \pm 14.0$\%  & $56.0 \pm 5.8$\%    \\ 
            Ours (w/o tone curve)                             & $68.5 \pm 12.8$\%  & $56.2 \pm 6.5$\%    \\ 
            Ours (w/o env.~map fusion)                        & $66.7 \pm 9.6$\%  & $57.6 \pm 6.6$\%    \\ 
            \bottomrule
        \end{tabular}
    }
\end{table} 

\begin{figure}[t!]
    \centering
    \begin{subfigure}{0.49\linewidth}
    \centering
    \includegraphics[width=0.98\linewidth,trim={0 0 0 0},clip]{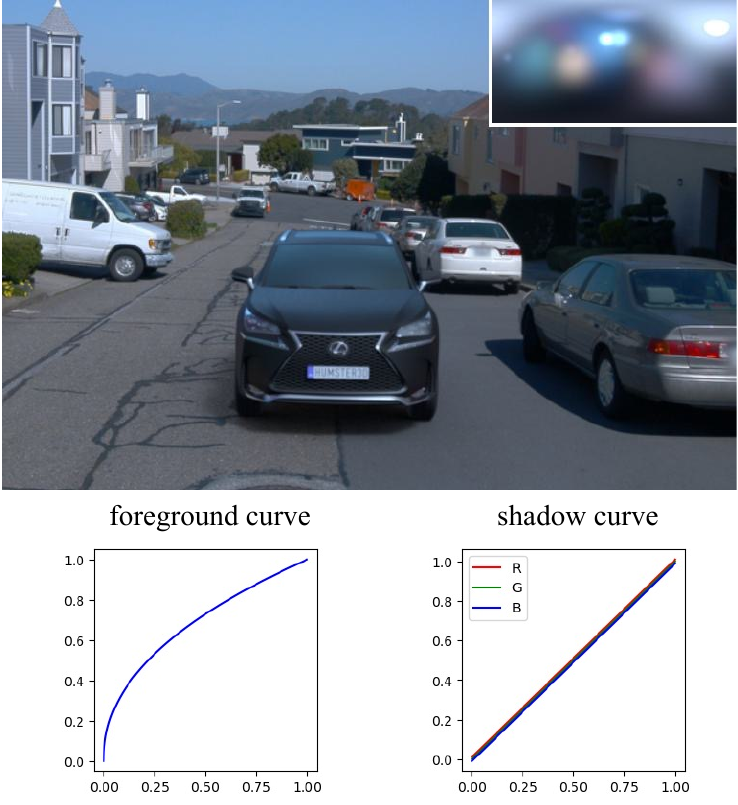}
    \caption{w/o optimizing tone curves}
    \label{fig:wo_tone}
    \end{subfigure}
    \begin{subfigure}{0.49\linewidth}
    \centering
    \includegraphics[width=0.98\linewidth,trim={0 0 0 0},clip]{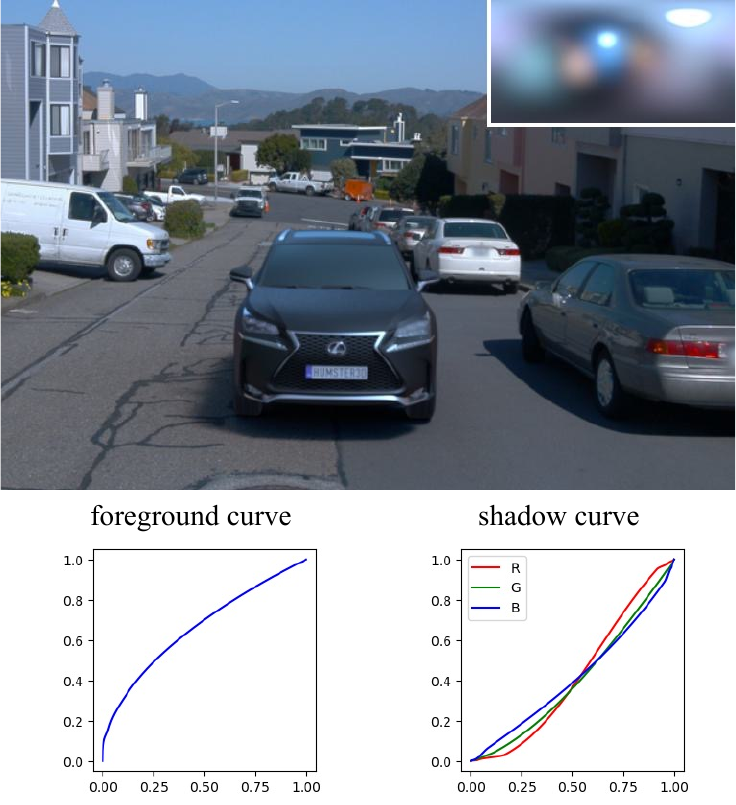}
    \caption{w/ optimizing tone curves}
    \label{fig:w_tone}
    \end{subfigure}
    \caption{Qualitative ablation on tone-mapping curve optimization. 
    The optimizable tone-mapping curve provides the capacity and flexibility to match the scale and color of the shadows. 
    (The visualized foreground curve considers gamma correction $\gamma=2.2$.)
    }
    \label{fig:ablation_tone}
\end{figure}

\subsection{Additional Qualitative Results}

Fig.~\ref{fig:qual_waymo_supp} and Fig.~\ref{fig:qual_polyhaven_supp} show the additional qualitative comparison against other baseline methods. Our method consistently performs well in various background images with challenging light conditions, while the baseline models often fail to capture the correct lighting direction or intensity scale. We also include more insertion examples in Fig.~\ref{fig:waymo_more} and Fig.~\ref{fig:hdri_more}. Video examples can be found on the project page.

\subsection{Tone-mapping Curve}
In Fig.~\ref{fig:ablation_tone}, we visualize the optimized tone-mapping curve and ablate the effect of the optimizable tone-mapping curves in our method. 
Comparing the results, the optimizable tone-mapping curve can effectively adjust the color and scale of the rendered shadow, and be blended more naturally in the background image. The foreground curve learns a residual from Reinhard tonemapping and is often close to identity mapping. 

\subsection{Failure Case Analysis}

In Fig.~\ref{fig:failure}, we show examples for the limitations mentioned in Sec. \ref{sec:limitations}.
\textit{a) Shiny reflection}. 
Our insertion of a shiny sphere correctly captures the general highlight direction, but cannot get all high-frequency details due to the limitations of SG lighting; 
\textit{b) Color drift}. The inserted dustbin is relatively brighter than the reference. This is because DM's data prior tends to assume a ``recycle dustbin" with a bright green color;
\textit{c) Double shadowing}. Our current method does not handle double shadowing with local occlusion, which can be improved when combined with 3D reconstruction methods. 

\begin{table}[ht]
    \centering
    \footnotesize
    \setlength\tabcolsep{0pt}
    \begin{tabularx}{0.92\linewidth}%
        {p{1em}*{3}{>{\centering\arraybackslash}X}} 
        {\rotatebox{90}{\,\,\, Reference}}&
        \includegraphics[width=0.98\linewidth, trim={0 0 0 0},clip]{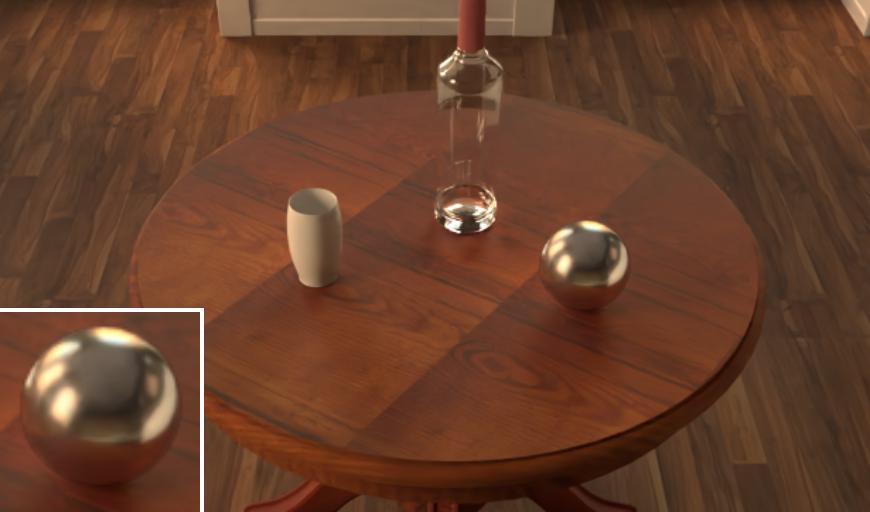} &
        \includegraphics[width=0.98\linewidth, trim={0 0 0 0},clip]{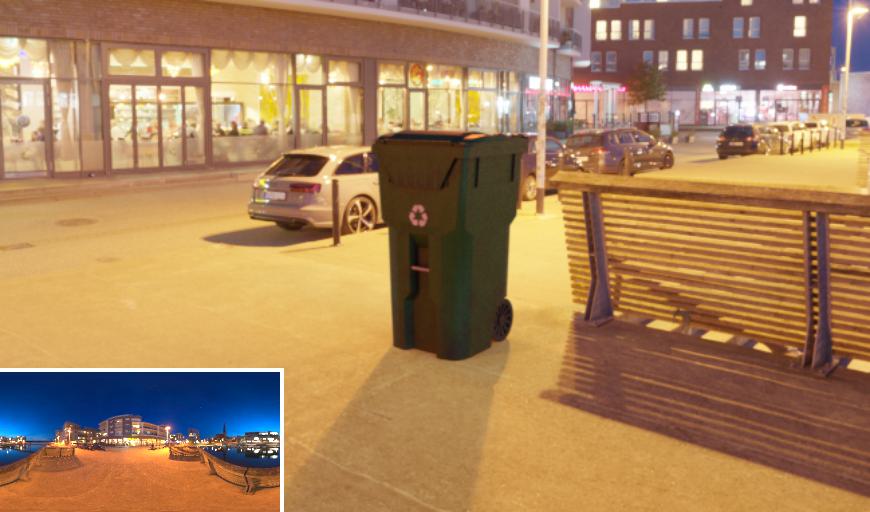} &
        \includegraphics[width=0.98\linewidth, trim={0 0 0 0},clip]{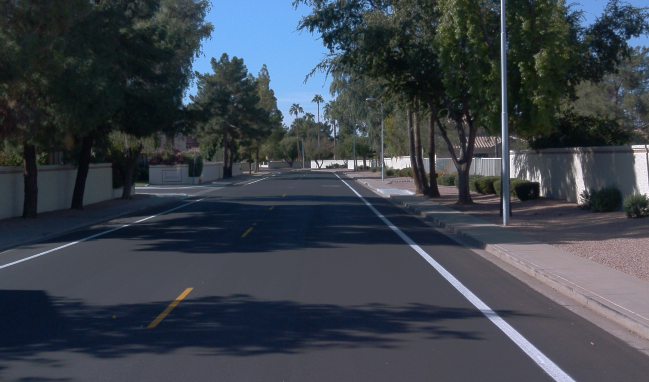}
         \\
        {\rotatebox{90}{\quad\,\,\, Ours}}&
        \includegraphics[width=0.98\linewidth, trim={0 0 0 0},clip]{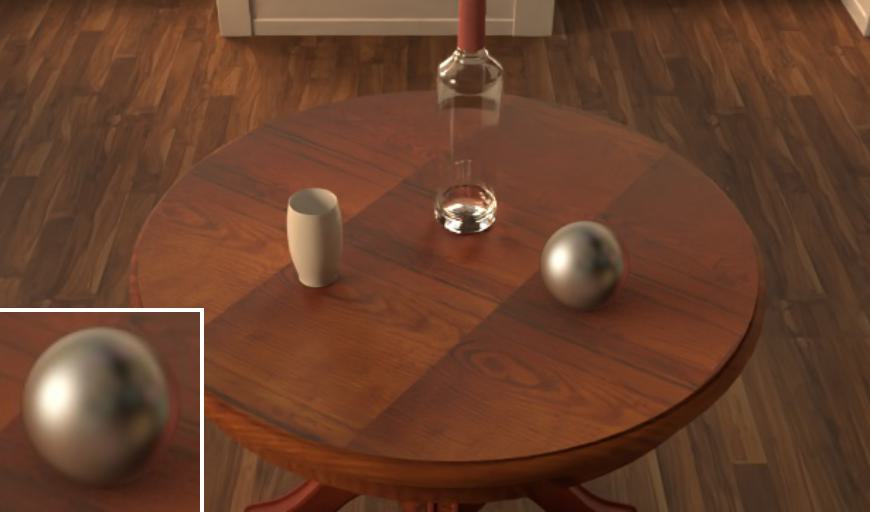} &
        \includegraphics[width=0.98\linewidth, trim={0 0 0 0},clip]{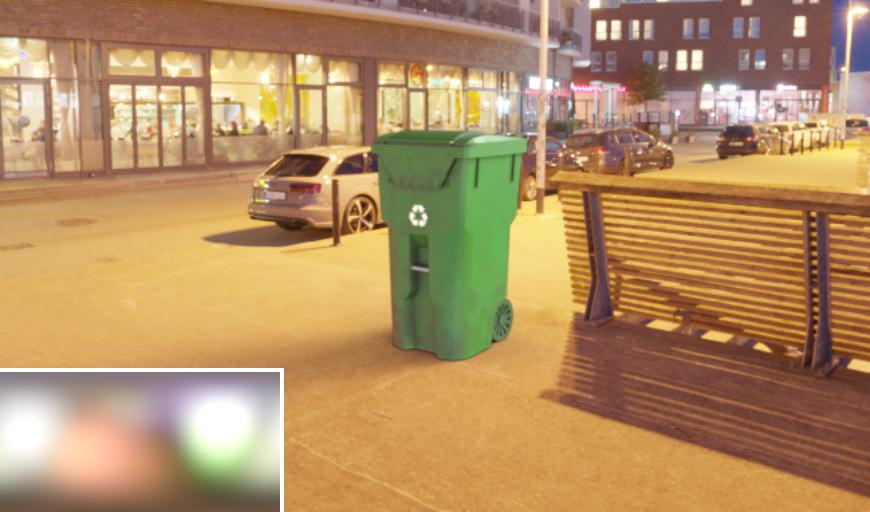} &
        \includegraphics[width=0.98\linewidth, trim={0 0 0 0},clip]{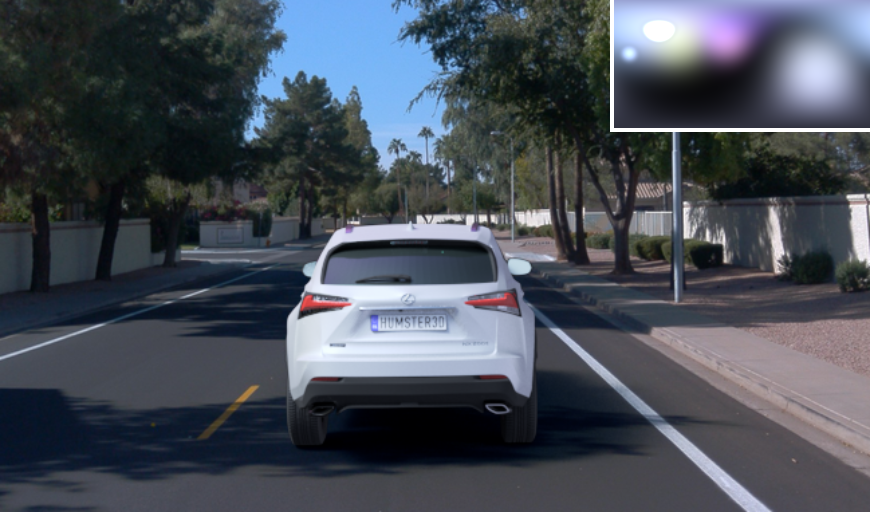}
         \\
        & Shiny reflection & Color drift & Double shadowing 
    \end{tabularx}%
    \makeatletter\def\@captype{figure}\makeatother
    \caption{
        Failure case examples.
    }
    \label{fig:failure}
    \vspace{-10pt}
\end{table}

\begin{table}[t]
    \centering
    \scriptsize
    \captionsetup{font=small}
        \setlength\tabcolsep{0pt}
        \begin{tabularx}{\linewidth}%
        {*{5}{>{\centering\arraybackslash}X}}
        Ours &
        DiffusionLight\cite{Phongthawee2023DiffusionLight} & StyleLight\cite{wang2022stylelight} & NLFE\cite{wang2022neural} & H-G \etal\cite{hold2019deep} \\
        \includegraphics[width=0.98\linewidth, trim={0 0 0 0},clip]{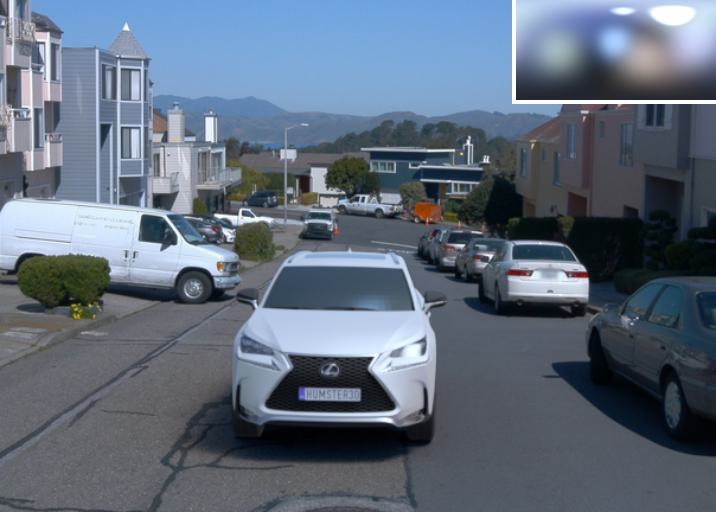} &
        \includegraphics[width=0.98\linewidth, trim={0 0 0 0},clip]{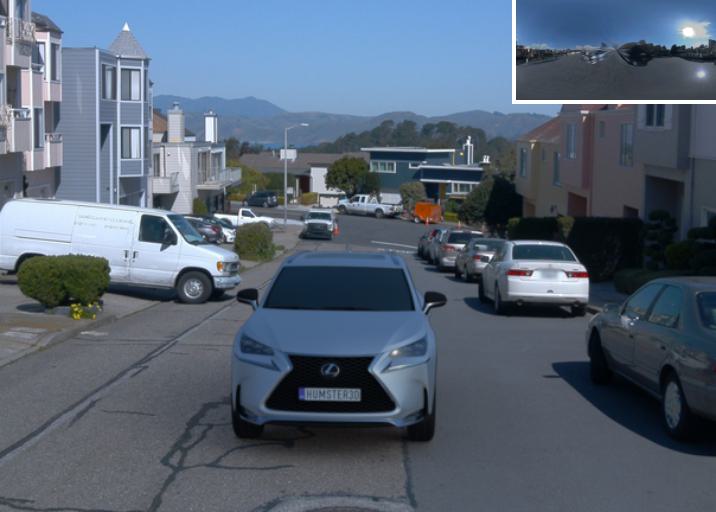} & 
        \includegraphics[width=0.98\linewidth, trim={0 0 0 0},clip]{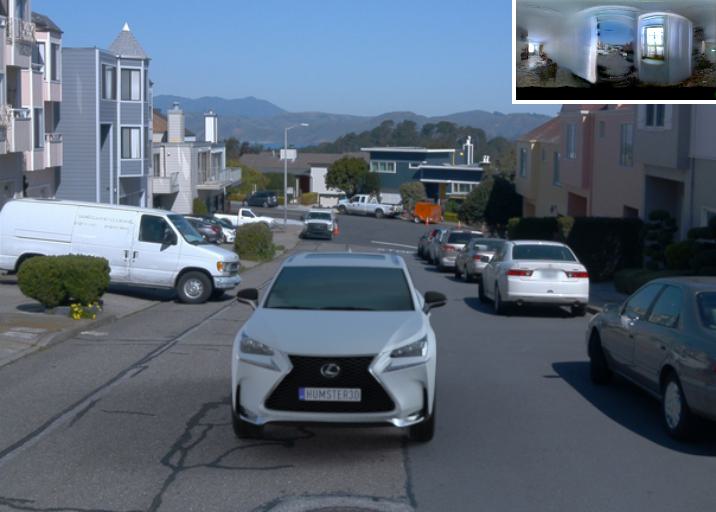} & 
        \includegraphics[width=0.98\linewidth, trim={0 0 0 0},clip]{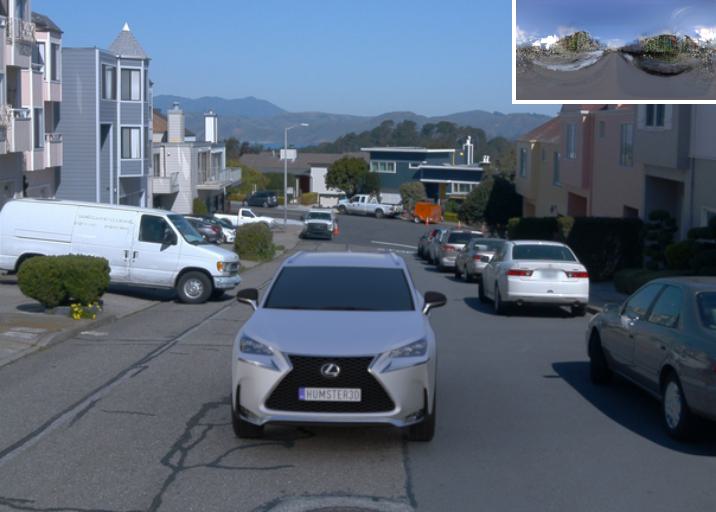} &
        \includegraphics[width=0.98\linewidth, trim={0 0 0 0},clip]{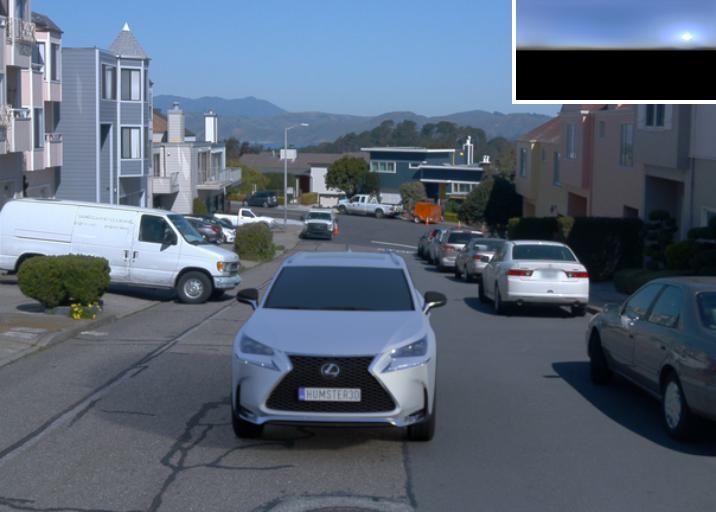}
        \\
        \includegraphics[width=0.98\linewidth, trim={0 0 0 0},clip]{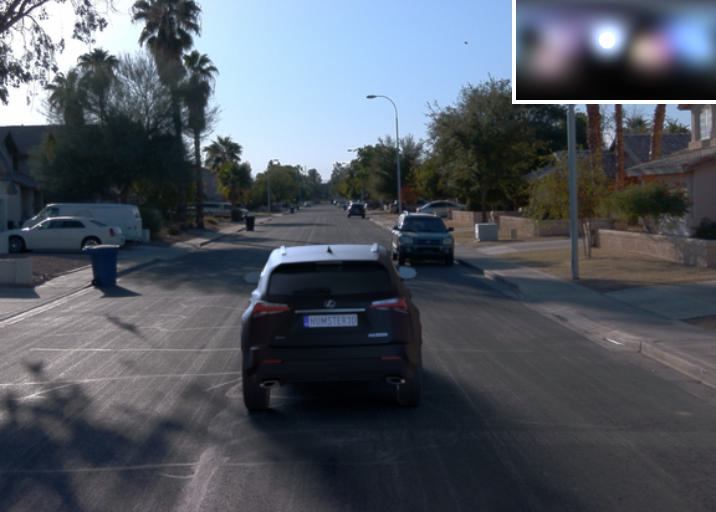} &
        \includegraphics[width=0.98\linewidth, trim={0 0 0 0},clip]{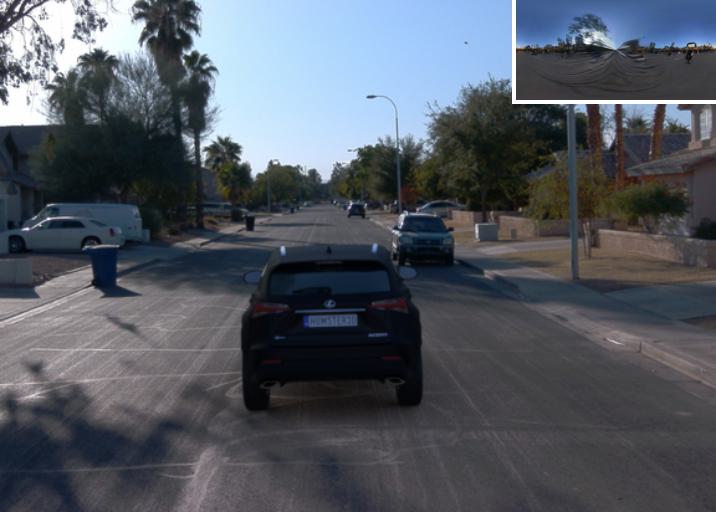} & 
        \includegraphics[width=0.98\linewidth, trim={0 0 0 0},clip]{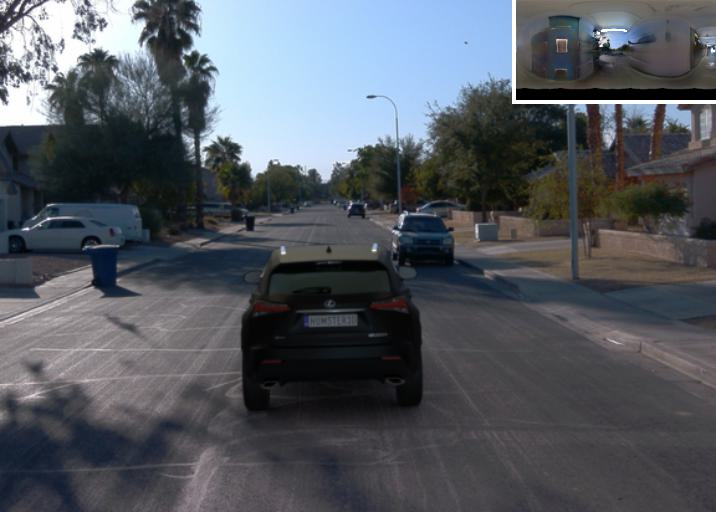} & 
        \includegraphics[width=0.98\linewidth, trim={0 0 0 0},clip]{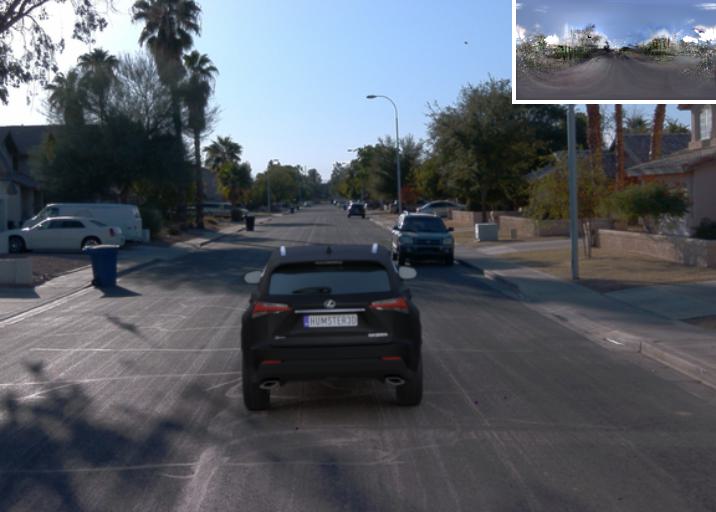} &
        \includegraphics[width=0.98\linewidth, trim={0 0 0 0},clip]{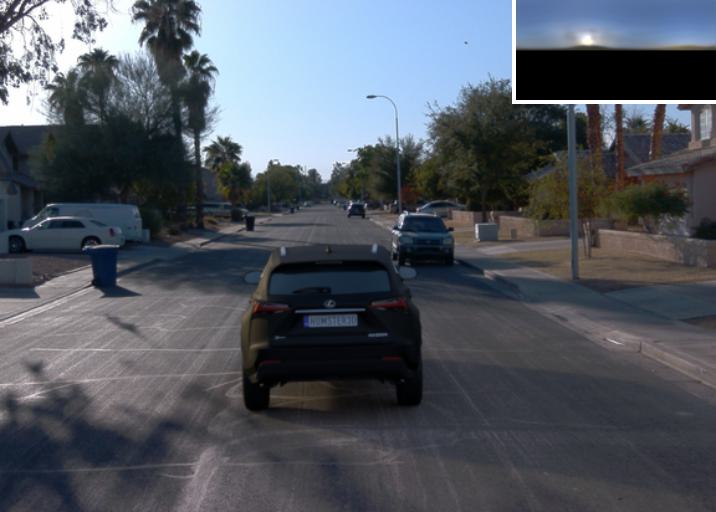}
        \\
        \includegraphics[width=0.98\linewidth, trim={0 0 0 0},clip]{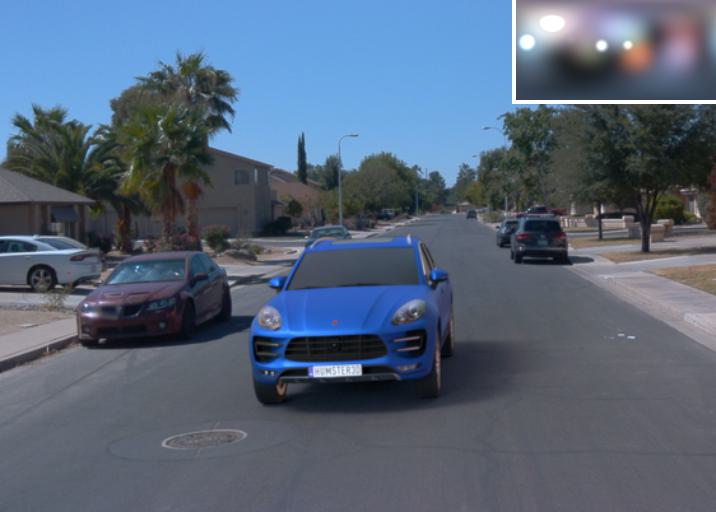} &
        \includegraphics[width=0.98\linewidth, trim={0 0 0 0},clip]{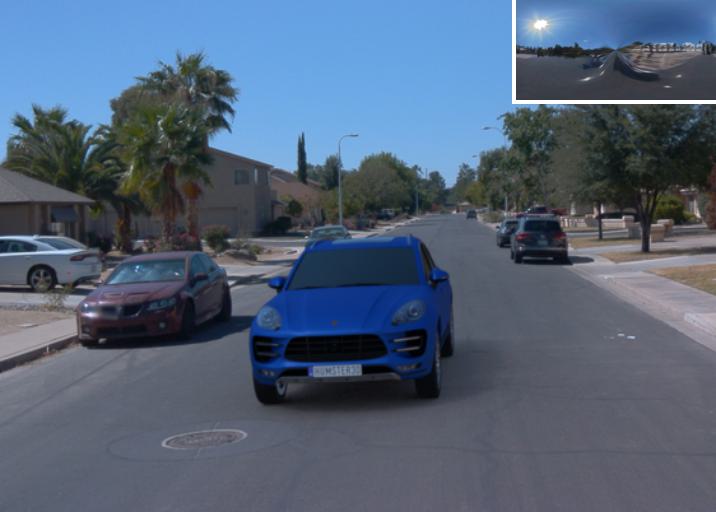} & 
        \includegraphics[width=0.98\linewidth, trim={0 0 0 0},clip]{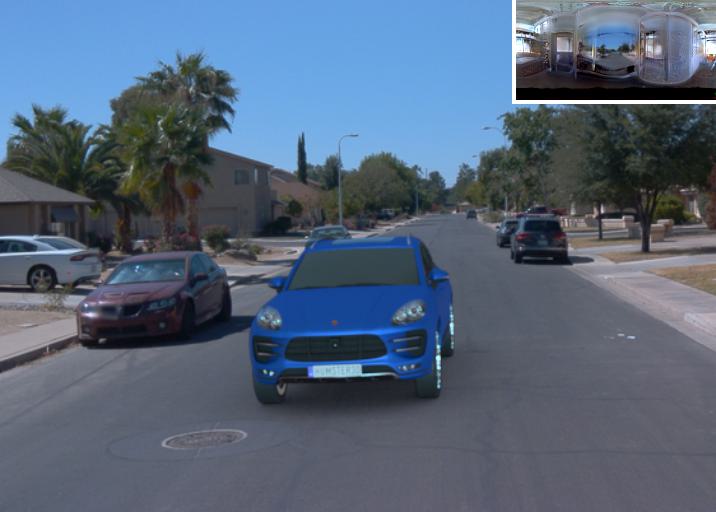} & 
        \includegraphics[width=0.98\linewidth, trim={0 0 0 0},clip]{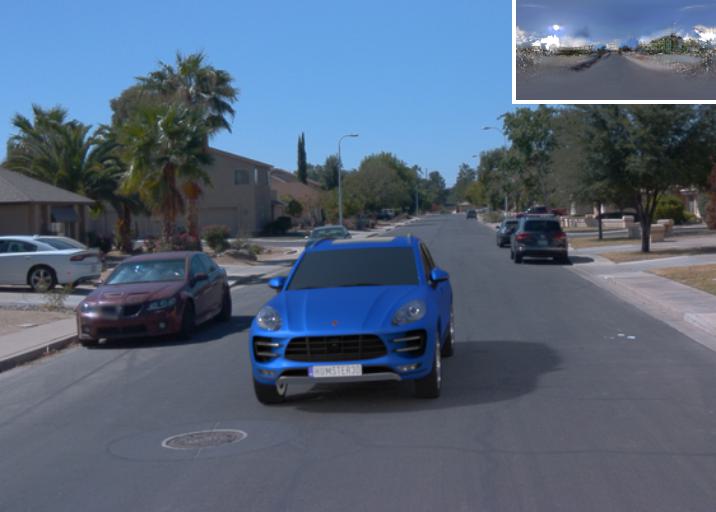} &
        \includegraphics[width=0.98\linewidth, trim={0 0 0 0},clip]{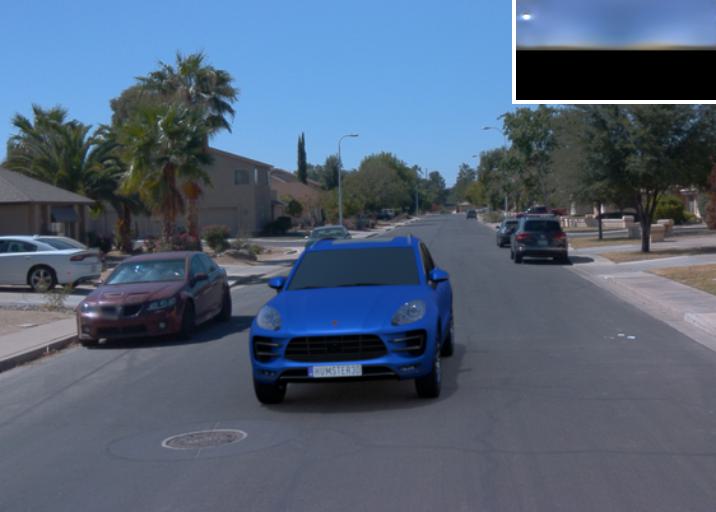}
        \\
        \includegraphics[width=0.98\linewidth, trim={0 0 0 0},clip]{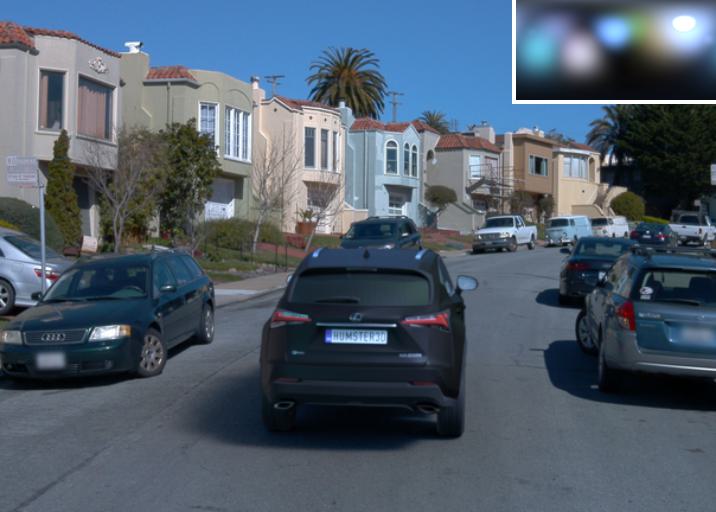} &
        \includegraphics[width=0.98\linewidth, trim={0 0 0 0},clip]{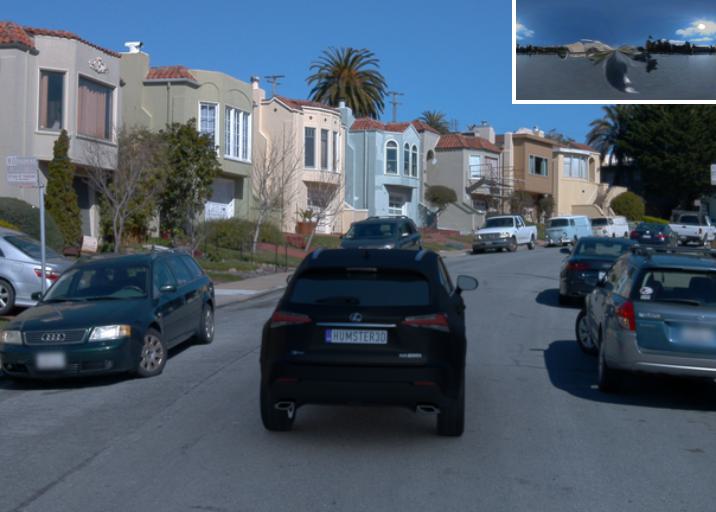} & 
        \includegraphics[width=0.98\linewidth, trim={0 0 0 0},clip]{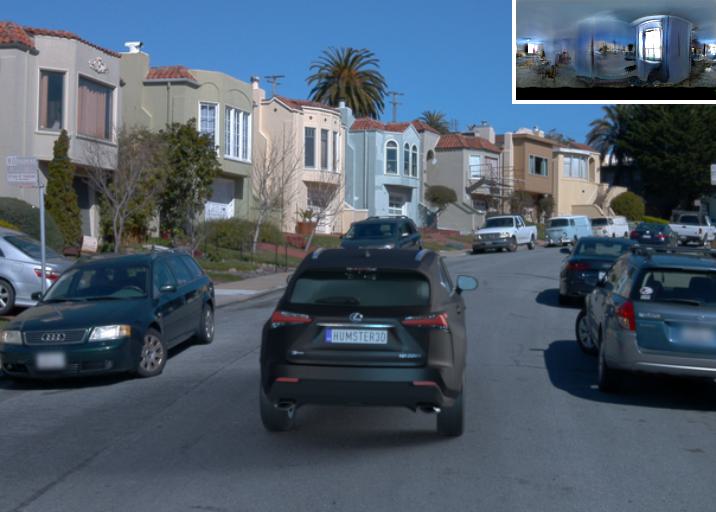} & 
        \includegraphics[width=0.98\linewidth, trim={0 0 0 0},clip]{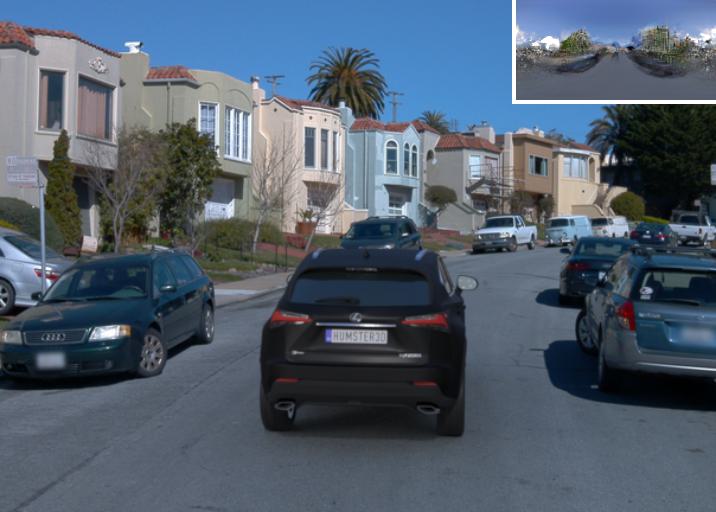} &
        \includegraphics[width=0.98\linewidth, trim={0 0 0 0},clip]{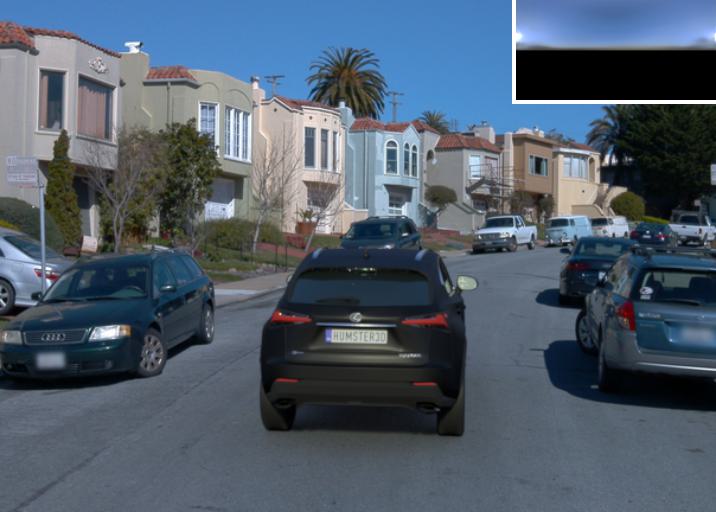}
        \\
        \includegraphics[width=0.98\linewidth, trim={0 0 0 0},clip]{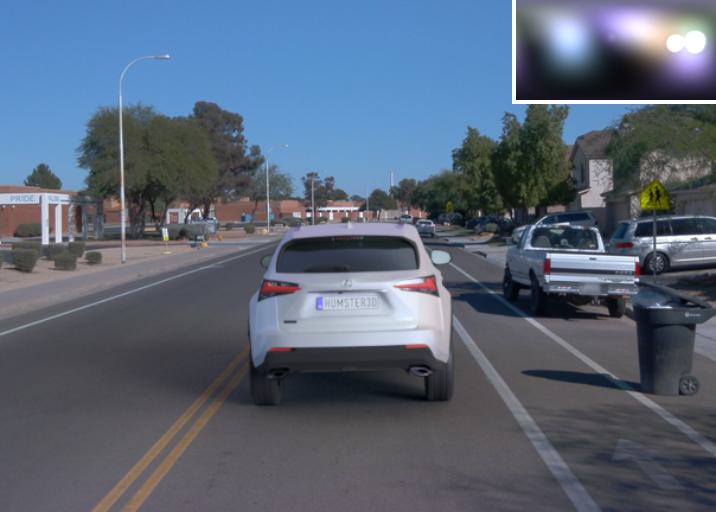} &
        \includegraphics[width=0.98\linewidth, trim={0 0 0 0},clip]{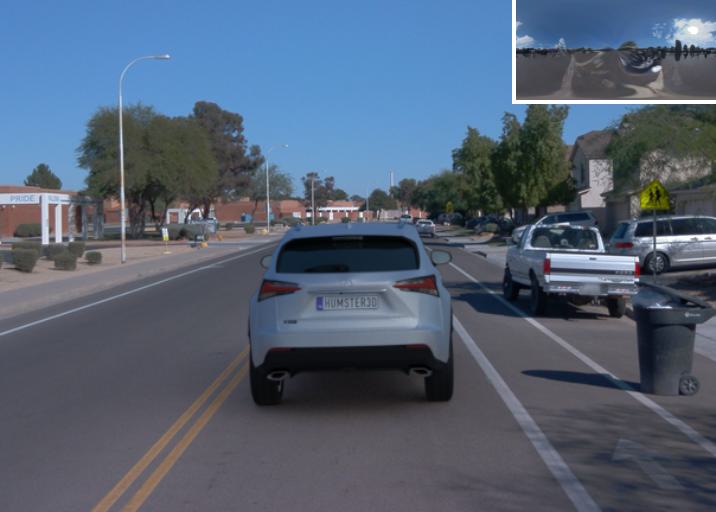} & 
        \includegraphics[width=0.98\linewidth, trim={0 0 0 0},clip]{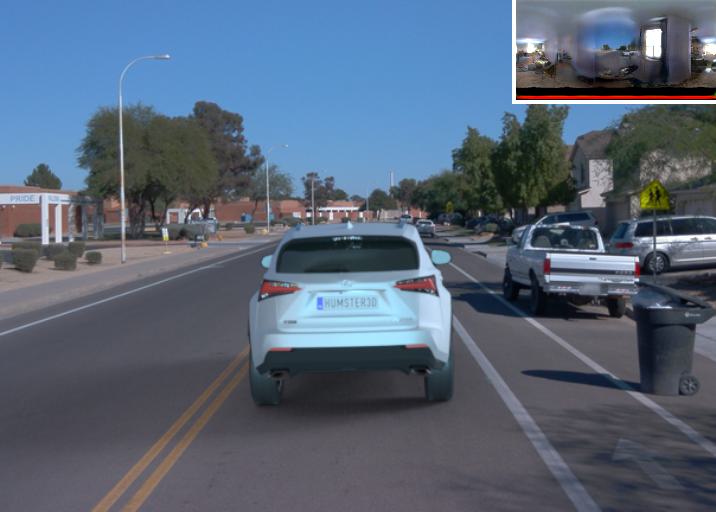} & 
        \includegraphics[width=0.98\linewidth, trim={0 0 0 0},clip]{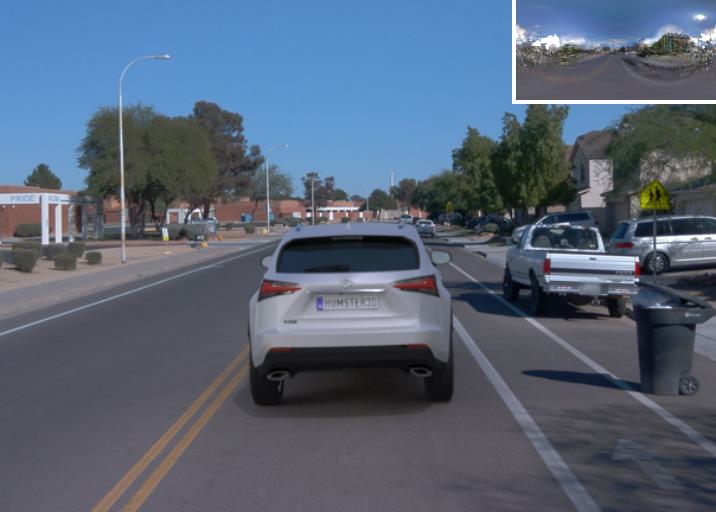} &
        \includegraphics[width=0.98\linewidth, trim={0 0 0 0},clip]{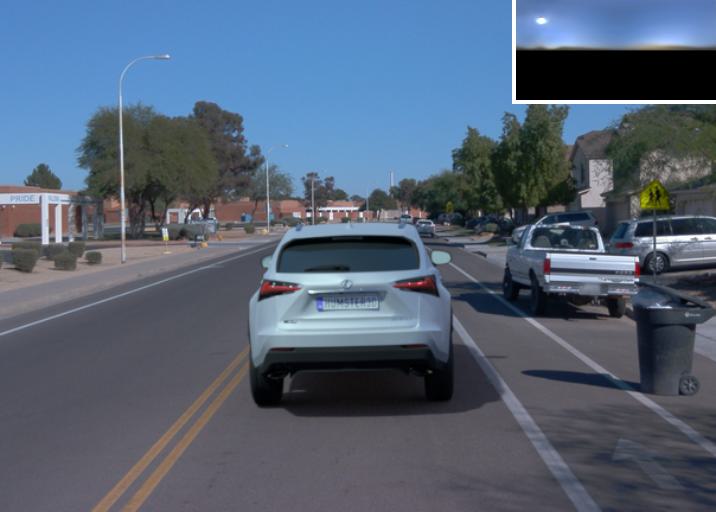}
        \\
        \includegraphics[width=0.98\linewidth, trim={0 0 0 0},clip]{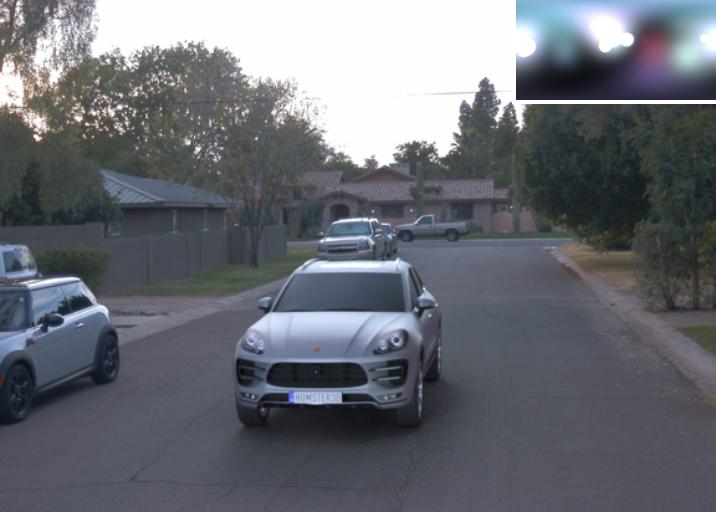} &
        \includegraphics[width=0.98\linewidth, trim={0 0 0 0},clip]{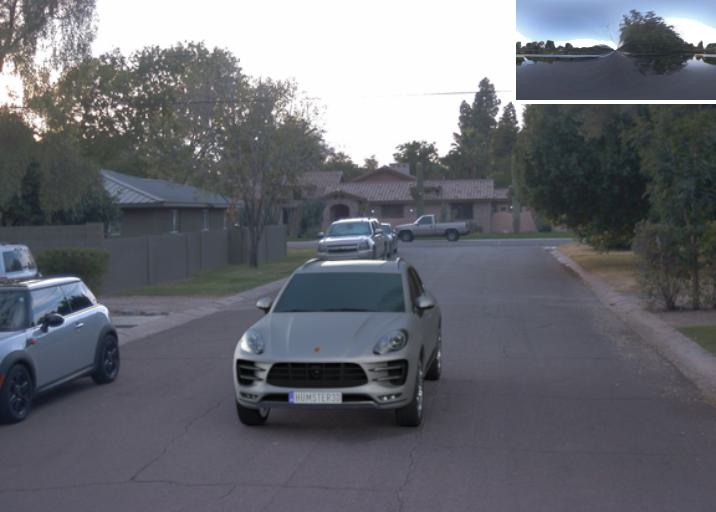} & 
        \includegraphics[width=0.98\linewidth, trim={0 0 0 0},clip]{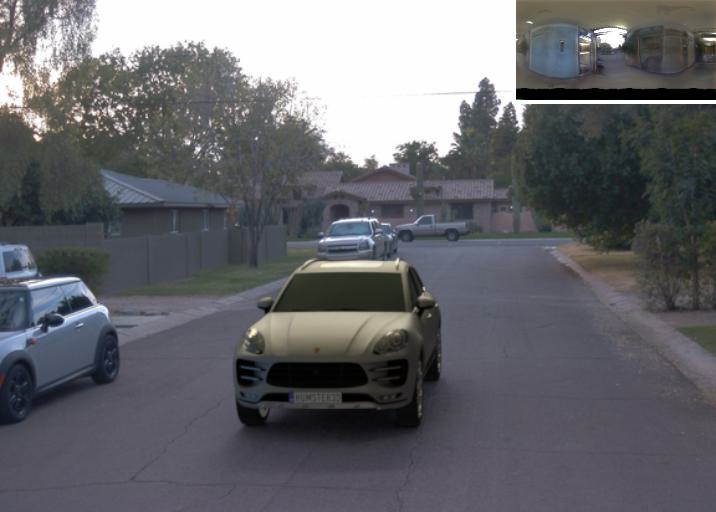} & 
        \includegraphics[width=0.98\linewidth, trim={0 0 0 0},clip]{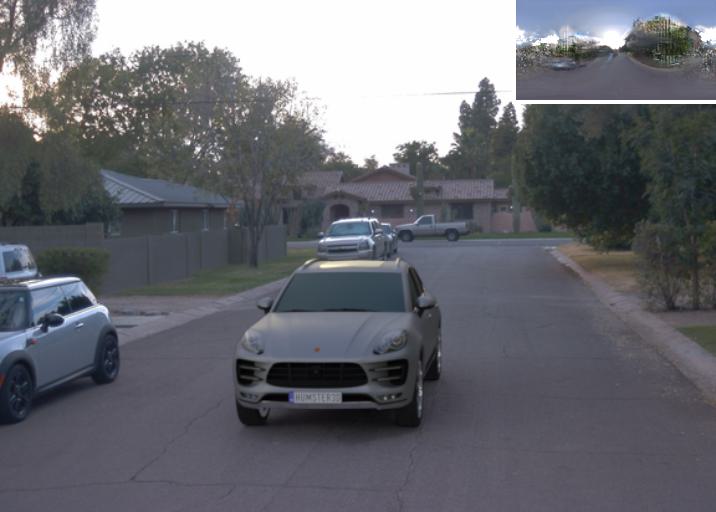} &
        \includegraphics[width=0.98\linewidth, trim={0 0 0 0},clip]{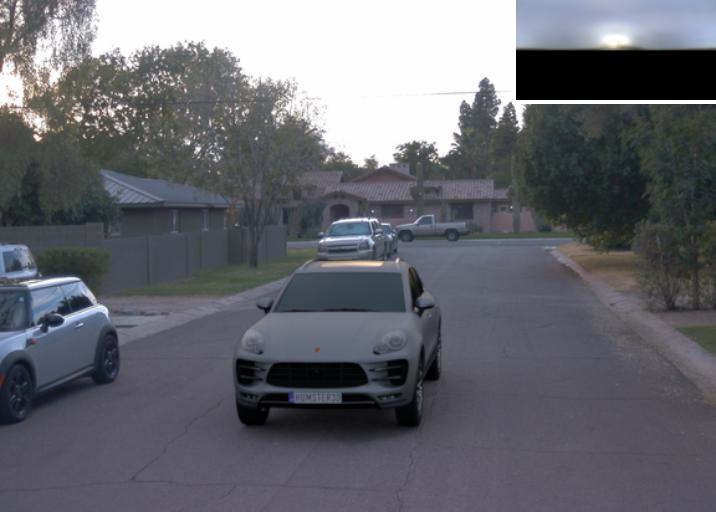}
        \\
        \includegraphics[width=0.98\linewidth, trim={0 0 0 0},clip]{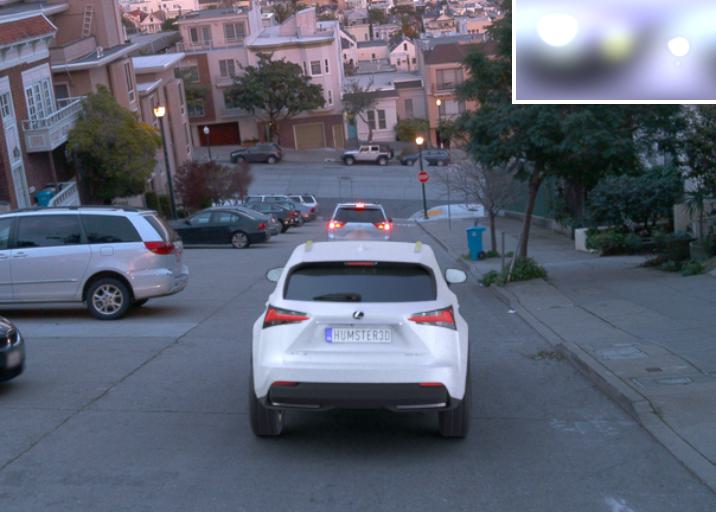} &
        \includegraphics[width=0.98\linewidth, trim={0 0 0 0},clip]{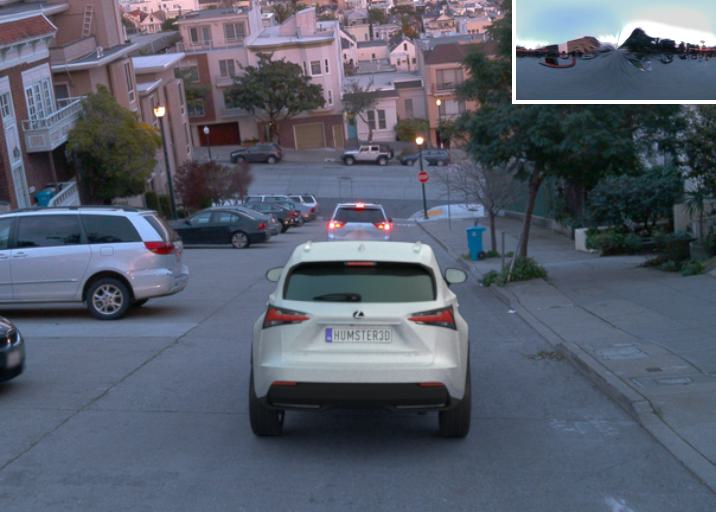} & 
        \includegraphics[width=0.98\linewidth, trim={0 0 0 0},clip]{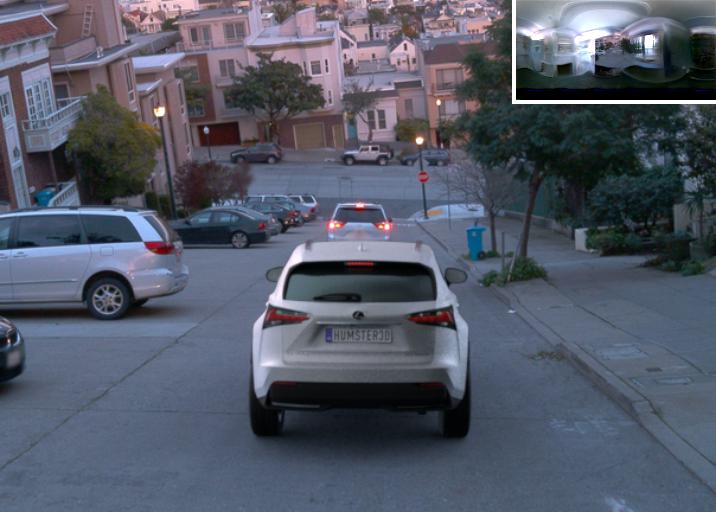} & 
        \includegraphics[width=0.98\linewidth, trim={0 0 0 0},clip]{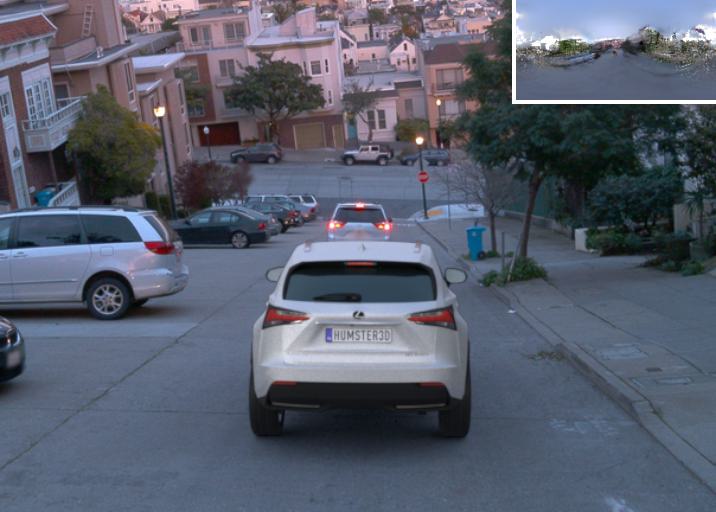} &
        \includegraphics[width=0.98\linewidth, trim={0 0 0 0},clip]{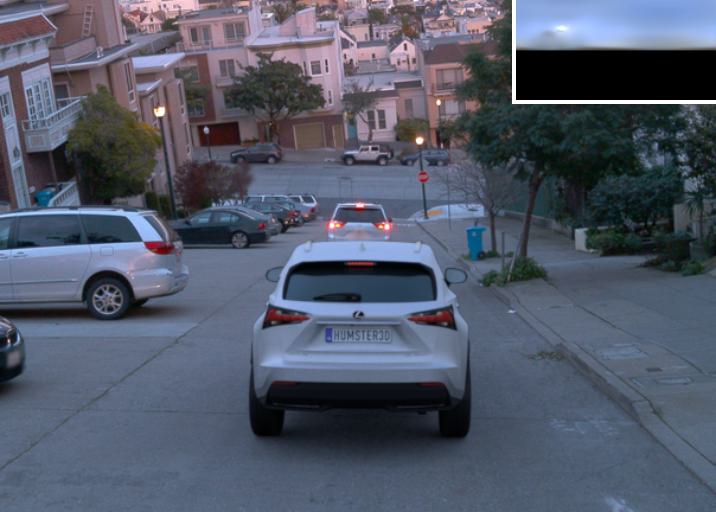}
        \\
        \includegraphics[width=0.98\linewidth, trim={0 0 0 0},clip]{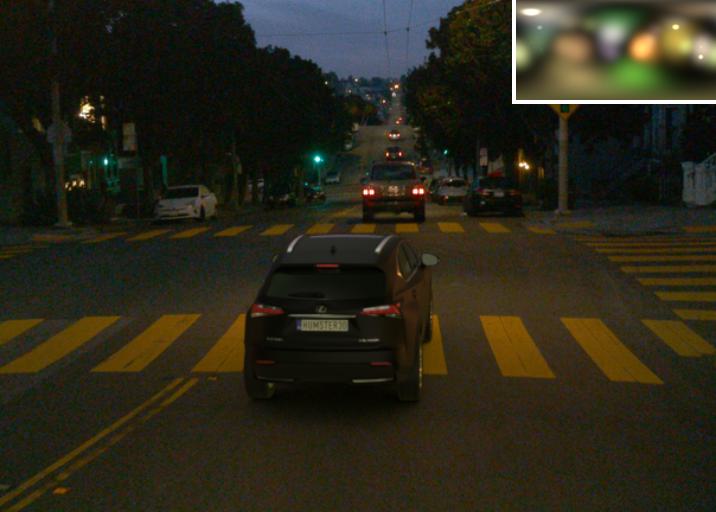} &
        \includegraphics[width=0.98\linewidth, trim={0 0 0 0},clip]{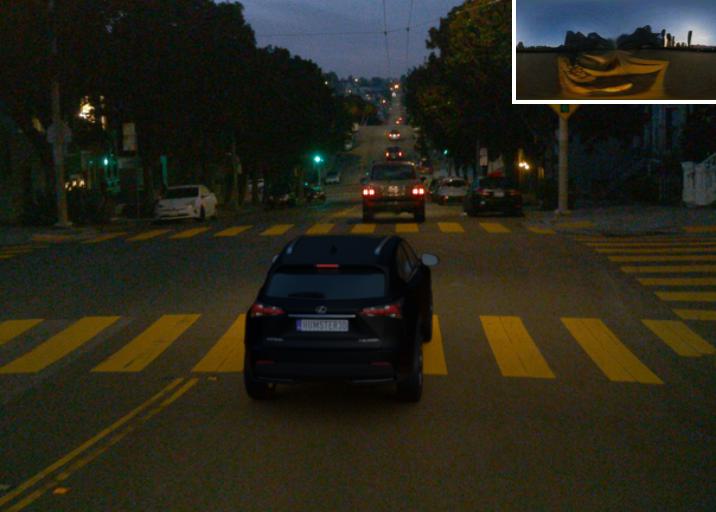} & 
        \includegraphics[width=0.98\linewidth, trim={0 0 0 0},clip]{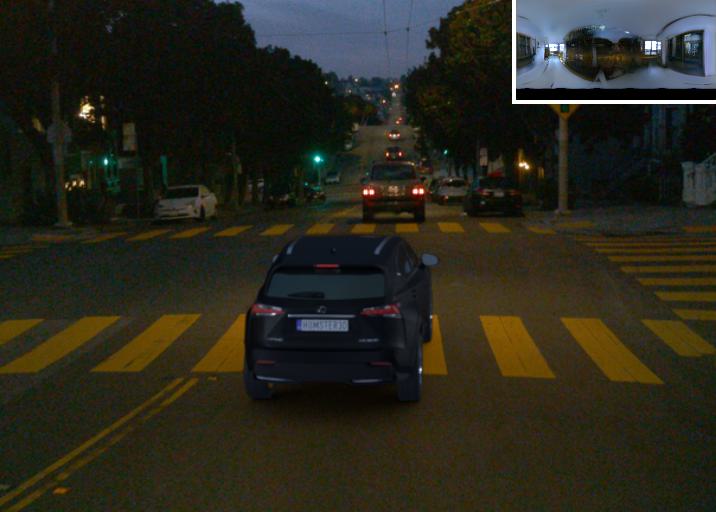} & 
        \includegraphics[width=0.98\linewidth, trim={0 0 0 0},clip]{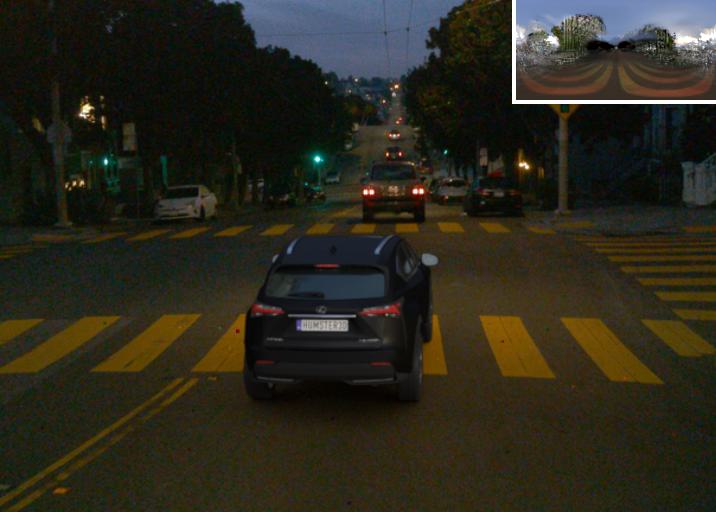} &
        \includegraphics[width=0.98\linewidth, trim={0 0 0 0},clip]{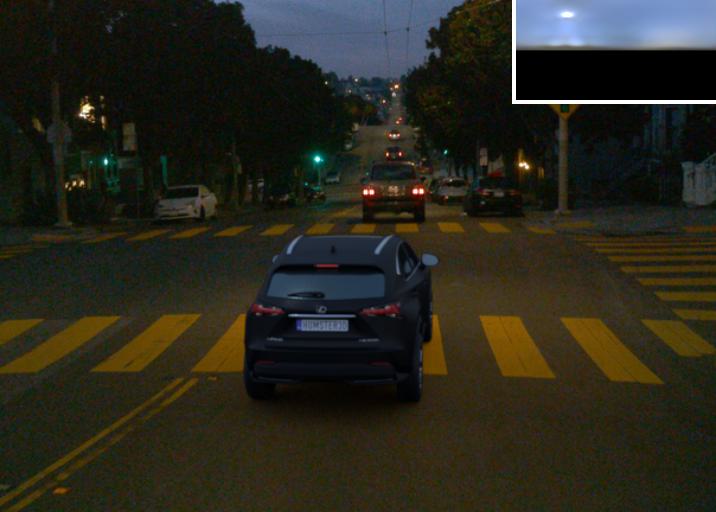}
        \\
        \includegraphics[width=0.98\linewidth, trim={0 0 0 0},clip]{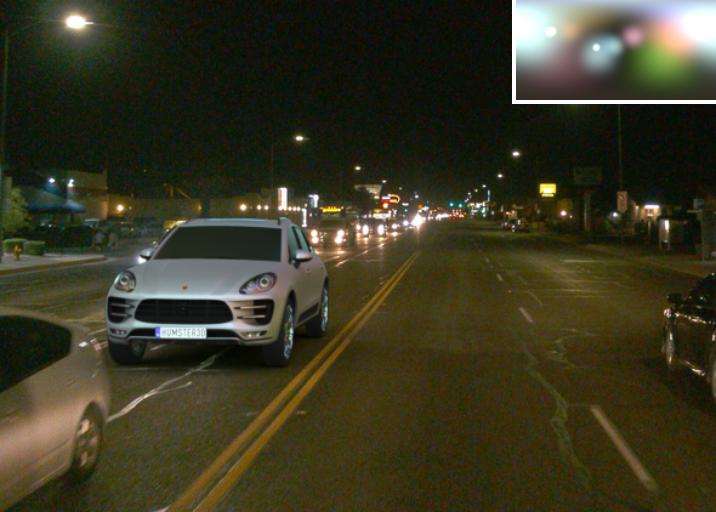} &
        \includegraphics[width=0.98\linewidth, trim={0 0 0 0},clip]{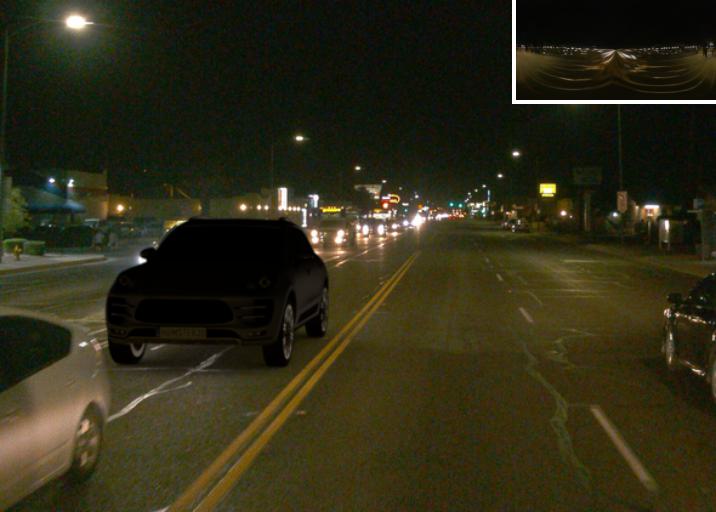} & 
        \includegraphics[width=0.98\linewidth, trim={0 0 0 0},clip]{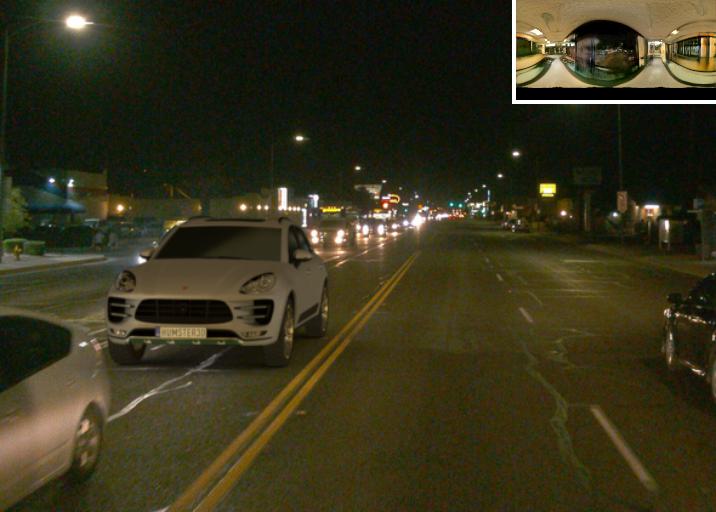} & 
        \includegraphics[width=0.98\linewidth, trim={0 0 0 0},clip]{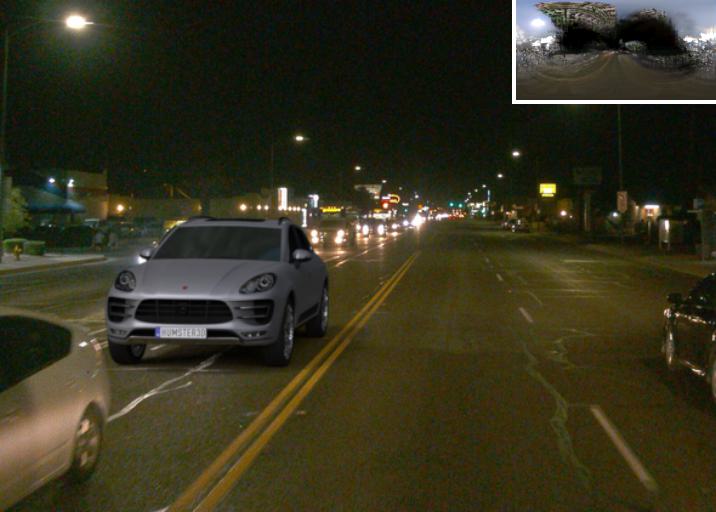} &
        \includegraphics[width=0.98\linewidth, trim={0 0 0 0},clip]{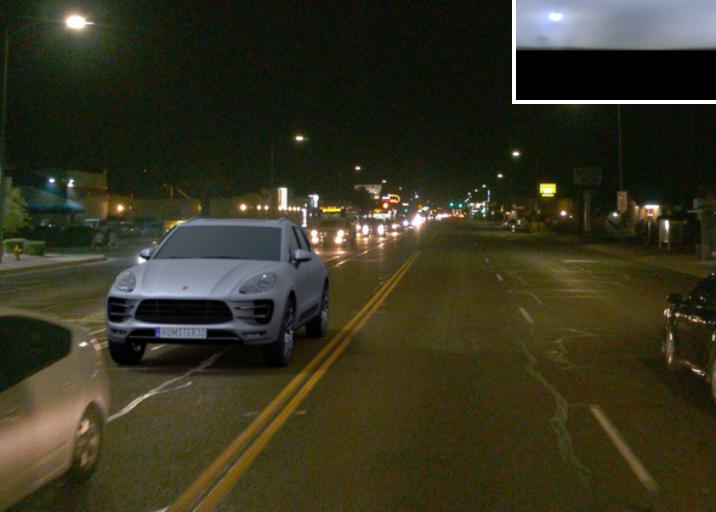}
    \end{tabularx}%
    \makeatletter\def\@captype{figure}\makeatother

    \caption{
        Additional visual comparisons on inserting virtual car assets into Waymo driving scenes.
    }
    \label{fig:qual_waymo_supp}
    \vspace{-10mm}
\end{table}

\begin{table}[t]
    \centering
    \scriptsize
    \captionsetup{font=small}
    \setlength\tabcolsep{0pt}
    \begin{tabularx}{\linewidth}%
        {*{5}{>{\centering\arraybackslash}X}}
        Reference   & Ours & DiffusionLight\cite{Phongthawee2023DiffusionLight} & StyleLight\cite{wang2022stylelight} & InvRend3D\cite{wang2021learning} \\
        \includegraphics[width=0.98\linewidth, trim={0 0 0 0},clip]{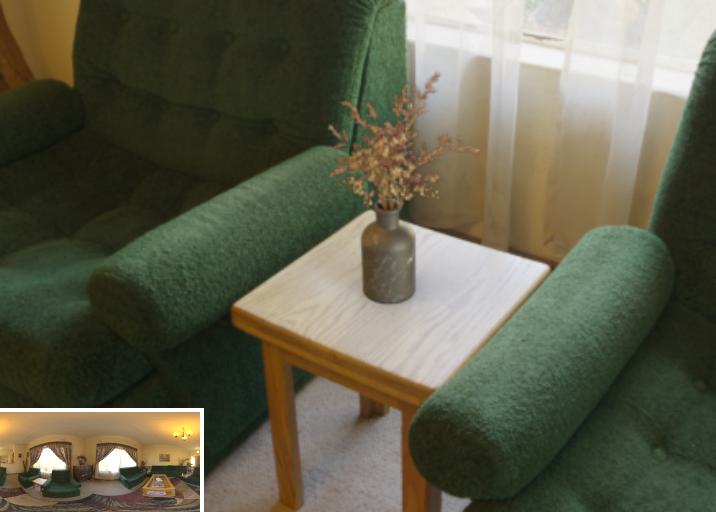} &
        \includegraphics[width=0.98\linewidth, trim={0 0 0 0},clip]{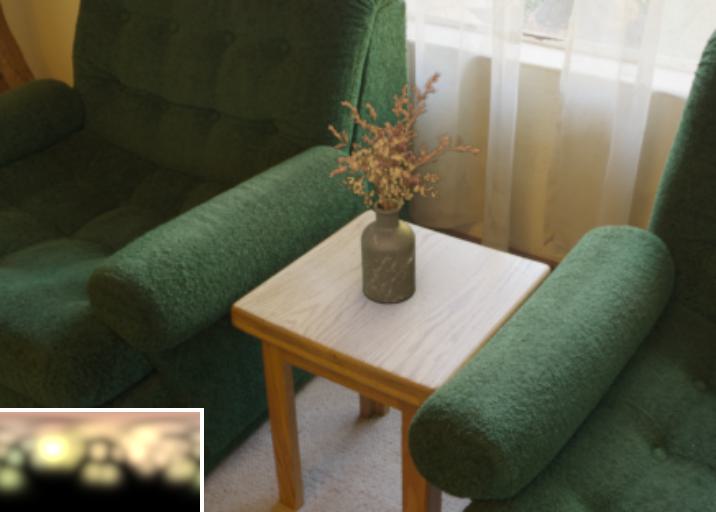} & 
        \includegraphics[width=0.98\linewidth, trim={0 0 0 0},clip]{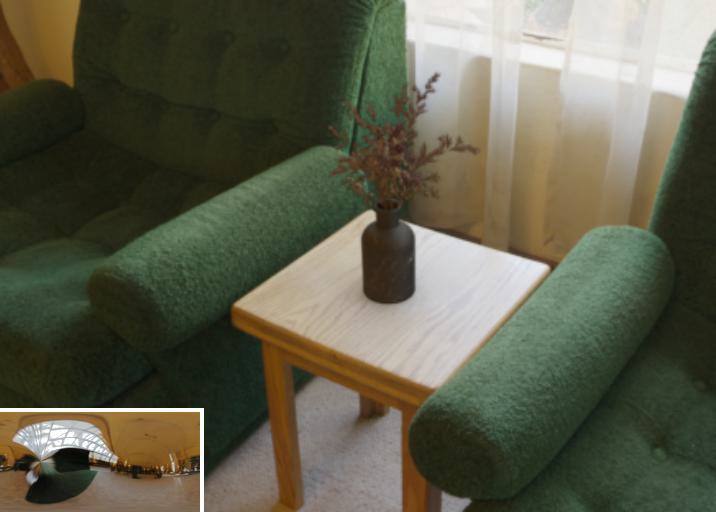} & 
        \includegraphics[width=0.98\linewidth, trim={0 0 0 0},clip]{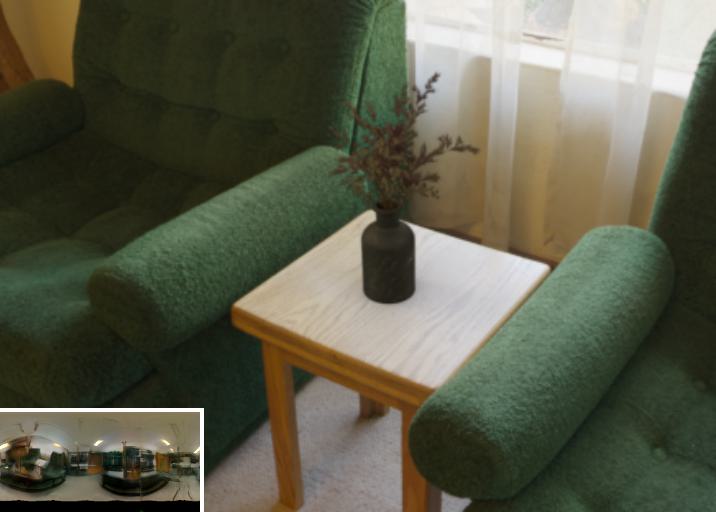} &
        \includegraphics[width=0.98\linewidth, trim={0 0 0 0},clip]{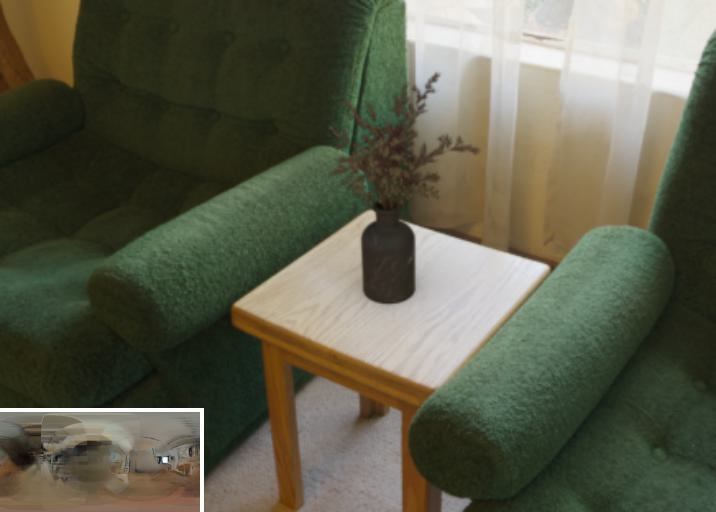}
        \\
        \includegraphics[width=0.98\linewidth, trim={0 0 0 0},clip]{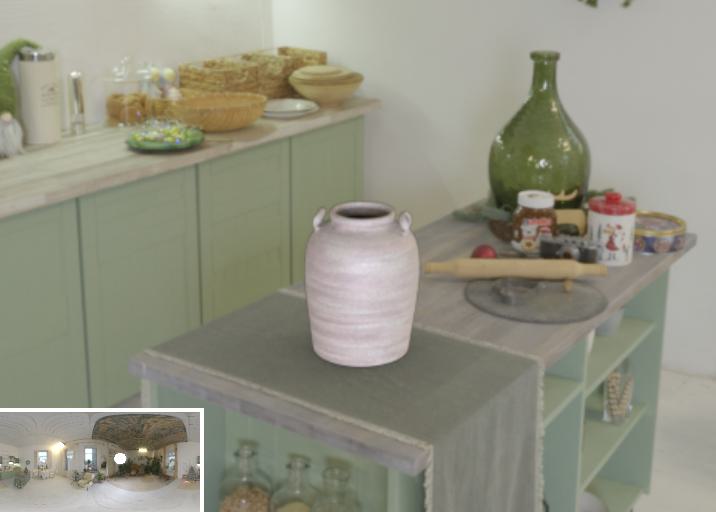} &
        \includegraphics[width=0.98\linewidth, trim={0 0 0 0},clip]{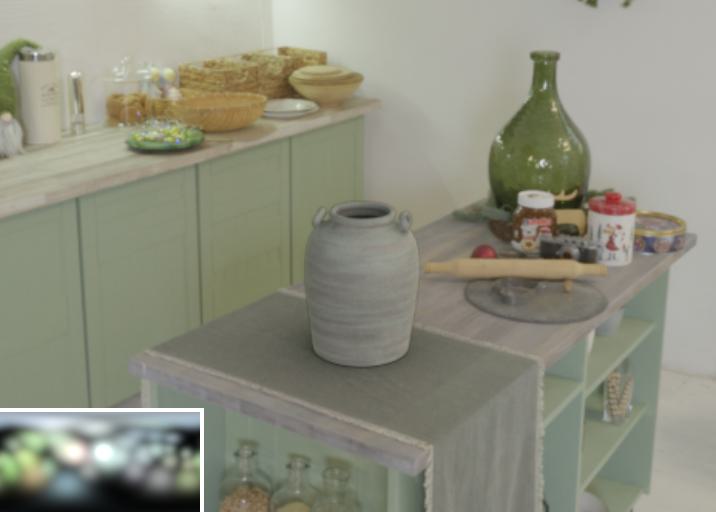} &
        \includegraphics[width=0.98\linewidth, trim={0 0 0 0},clip]{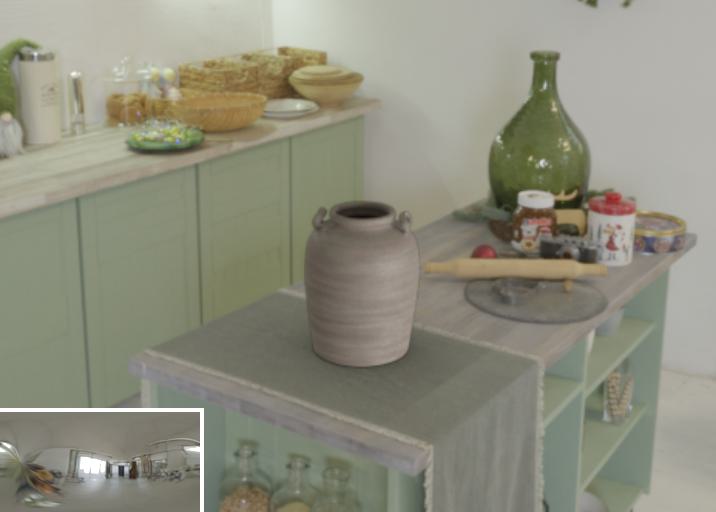} &
        \includegraphics[width=0.98\linewidth, trim={0 0 0 0},clip]{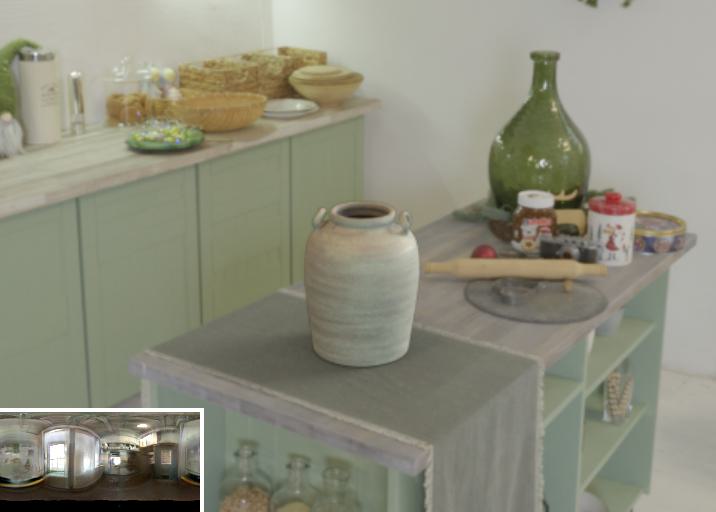} &
        \includegraphics[width=0.98\linewidth, trim={0 0 0 0},clip]{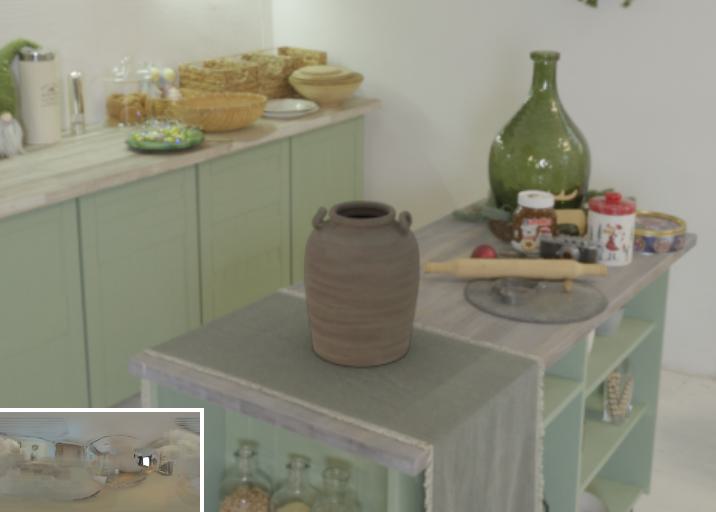}
        \\
        \includegraphics[width=0.98\linewidth, trim={0 0 0 0},clip]{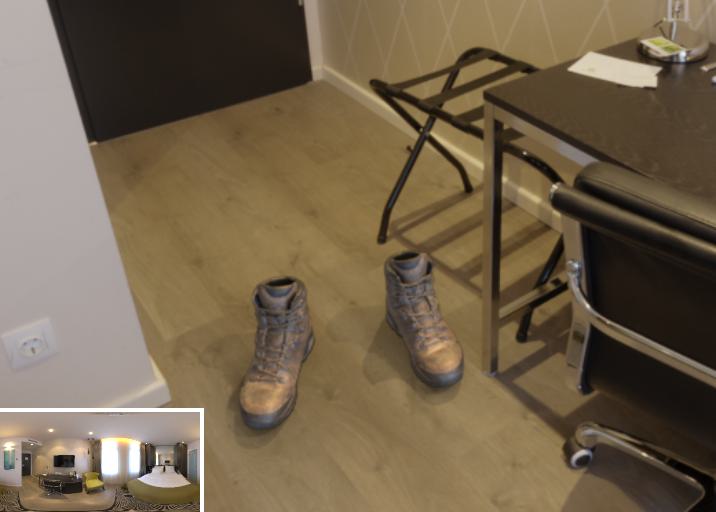} &
        \includegraphics[width=0.98\linewidth, trim={0 0 0 0},clip]{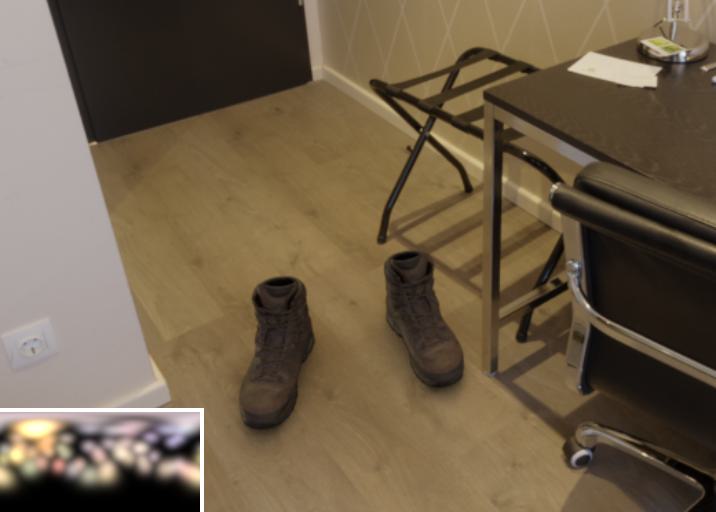} &
        \includegraphics[width=0.98\linewidth, trim={0 0 0 0},clip]{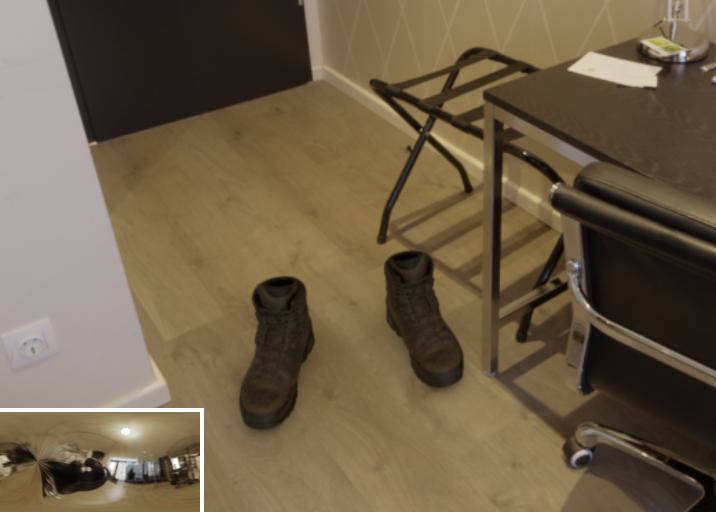} &
        \includegraphics[width=0.98\linewidth, trim={0 0 0 0},clip]{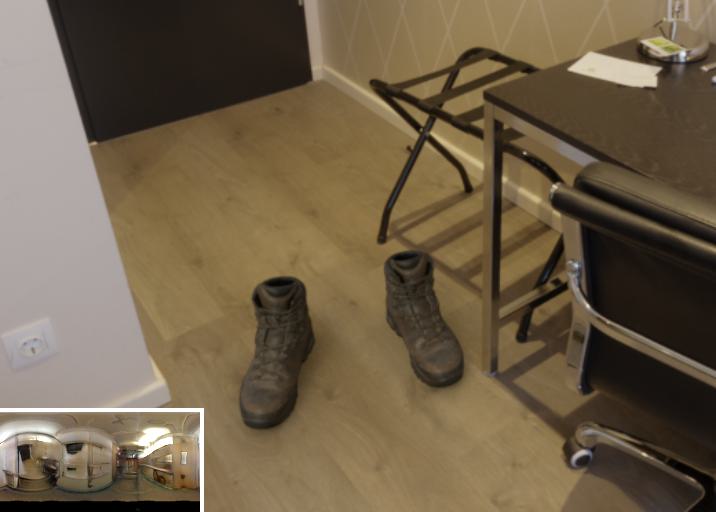} &
        \includegraphics[width=0.98\linewidth, trim={0 0 0 0},clip]{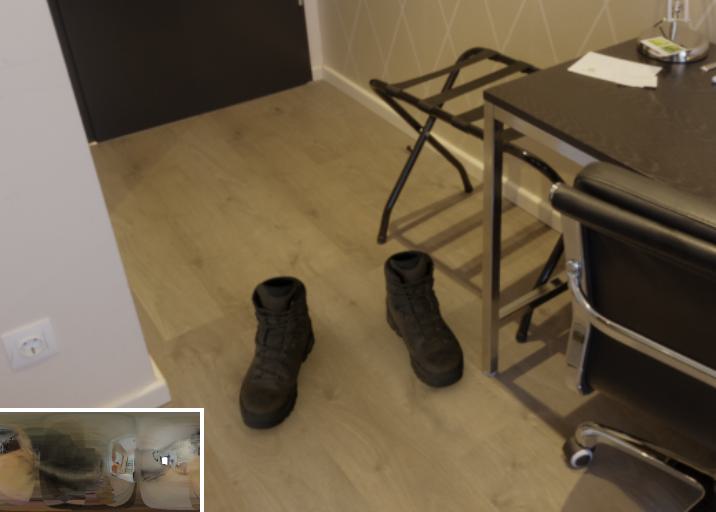}
        \\
        \includegraphics[width=0.98\linewidth, trim={0 0 0 0},clip]{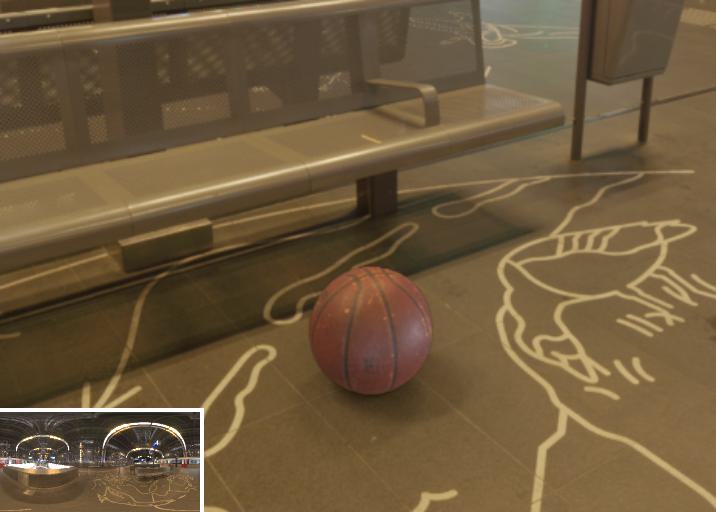} &
        \includegraphics[width=0.98\linewidth, trim={0 0 0 0},clip]{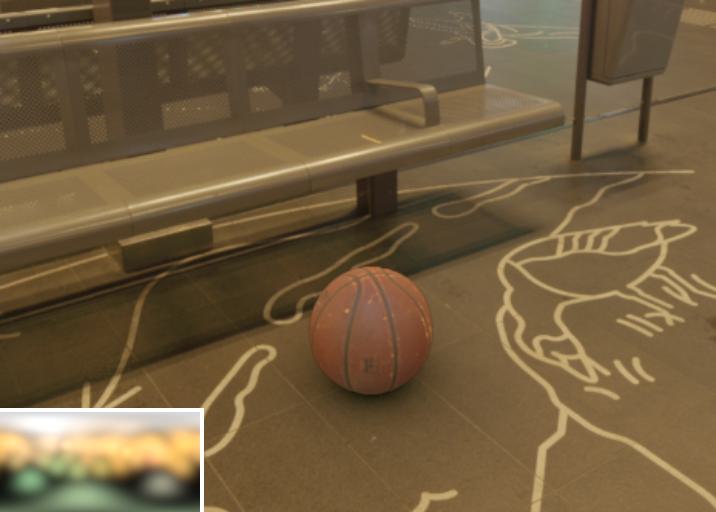} &
        \includegraphics[width=0.98\linewidth, trim={0 0 0 0},clip]{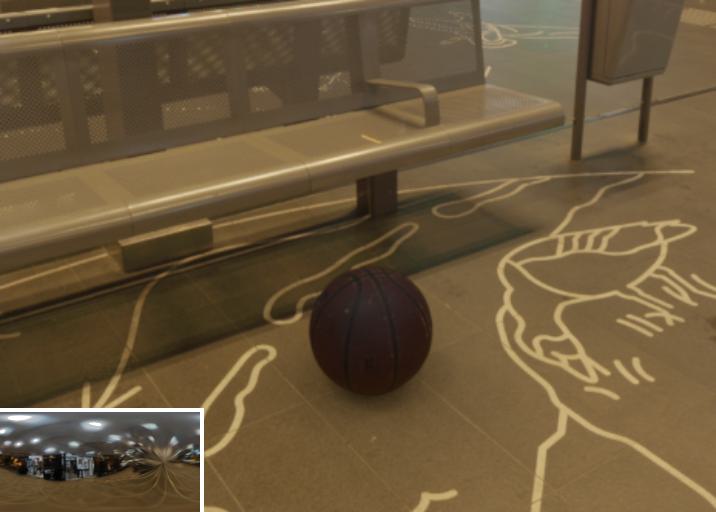} &
        \includegraphics[width=0.98\linewidth, trim={0 0 0 0},clip]{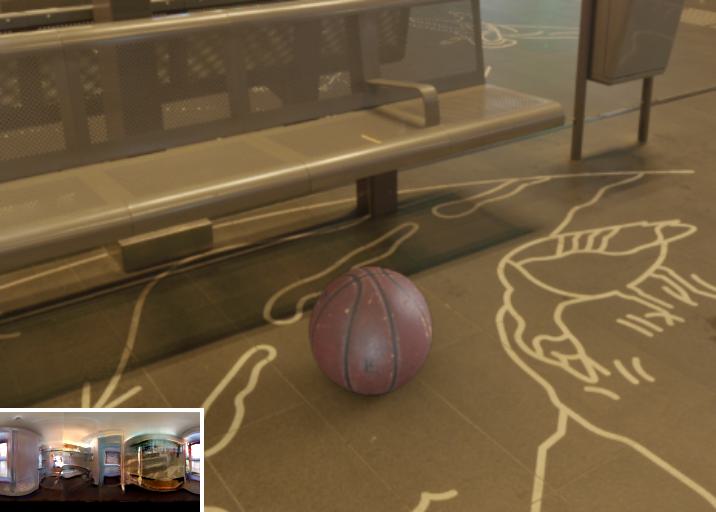} &
        \includegraphics[width=0.98\linewidth, trim={0 0 0 0},clip]{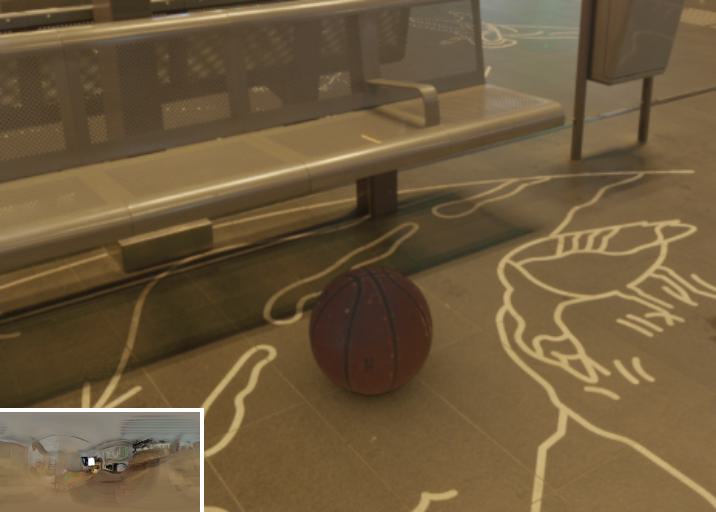}
        \\
        \includegraphics[width=0.98\linewidth, trim={0 0 0 0},clip]{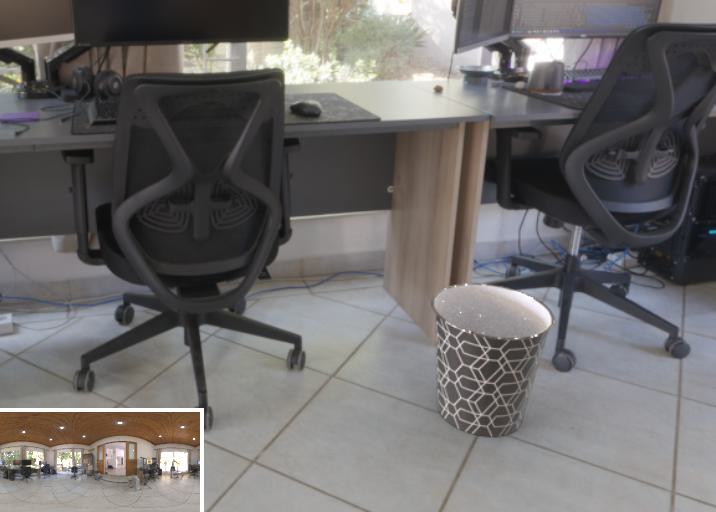} &
        \includegraphics[width=0.98\linewidth, trim={0 0 0 0},clip]{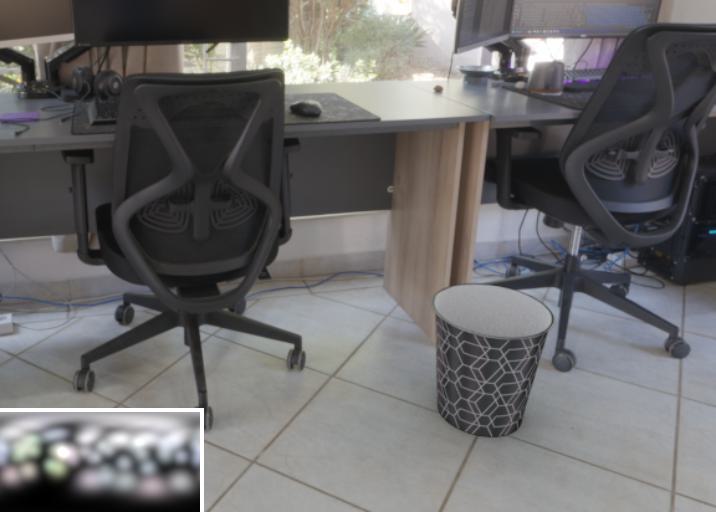} &
        \includegraphics[width=0.98\linewidth, trim={0 0 0 0},clip]{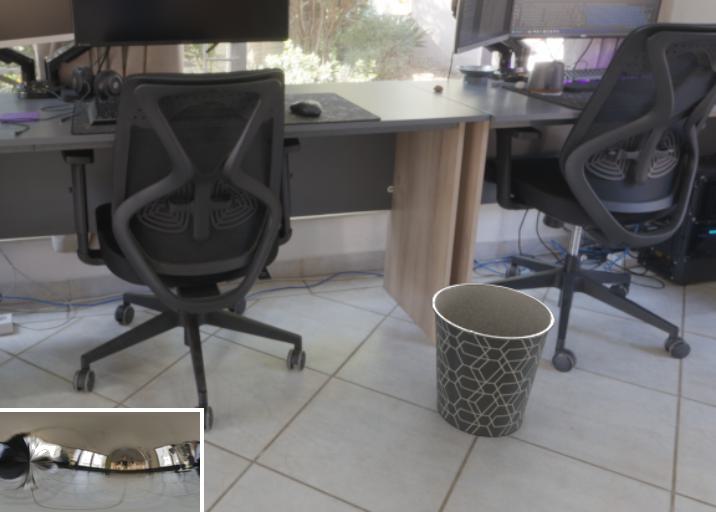} &
        \includegraphics[width=0.98\linewidth, trim={0 0 0 0},clip]{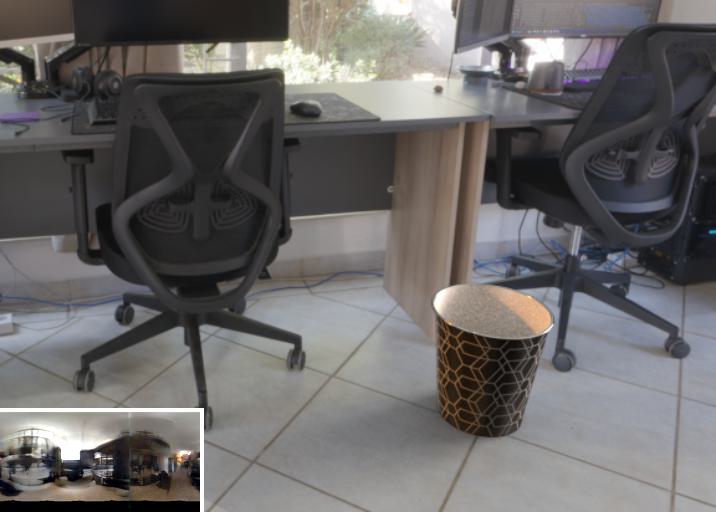} &
        \includegraphics[width=0.98\linewidth, trim={0 0 0 0},clip]{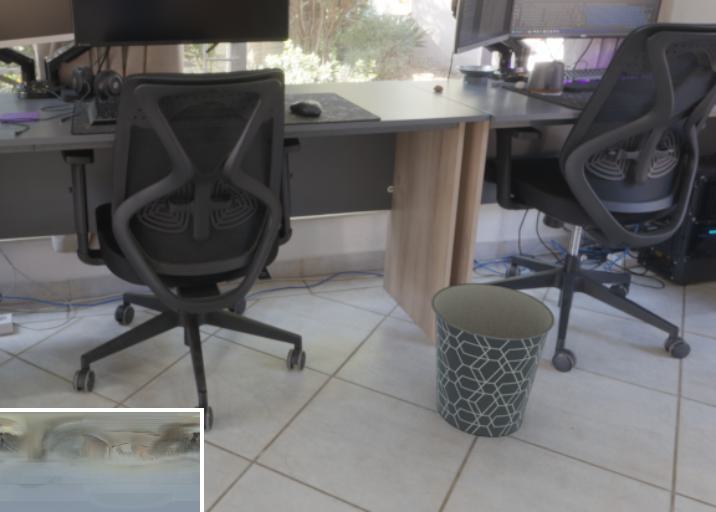}
        \\
        \includegraphics[width=0.98\linewidth, trim={0 0 0 0},clip]{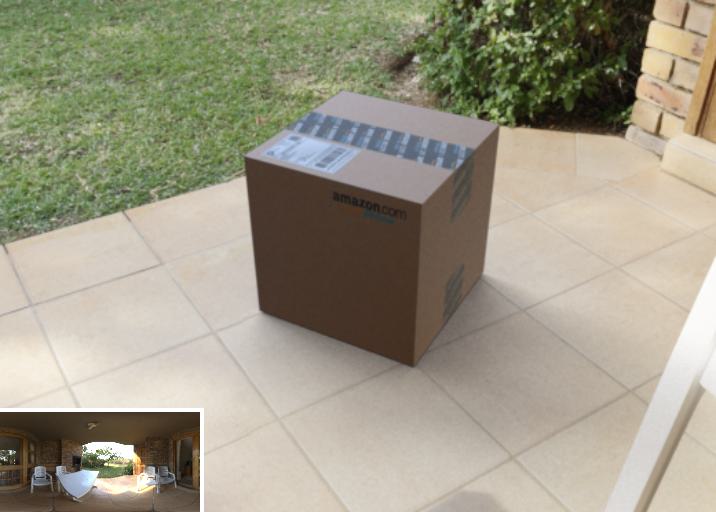} &
        \includegraphics[width=0.98\linewidth, trim={0 0 0 0},clip]{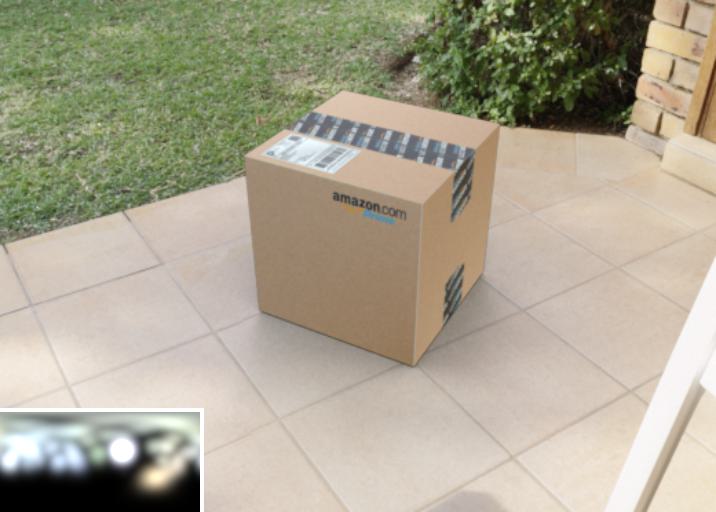} &
        \includegraphics[width=0.98\linewidth, trim={0 0 0 0},clip]{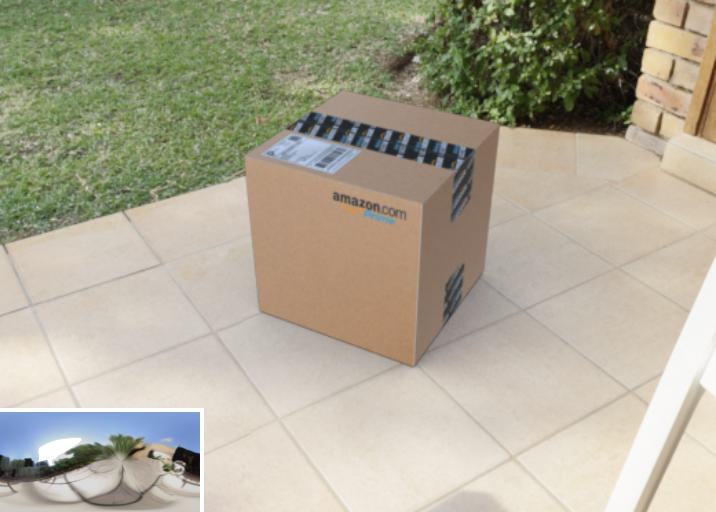} &
        \includegraphics[width=0.98\linewidth, trim={0 0 0 0},clip]{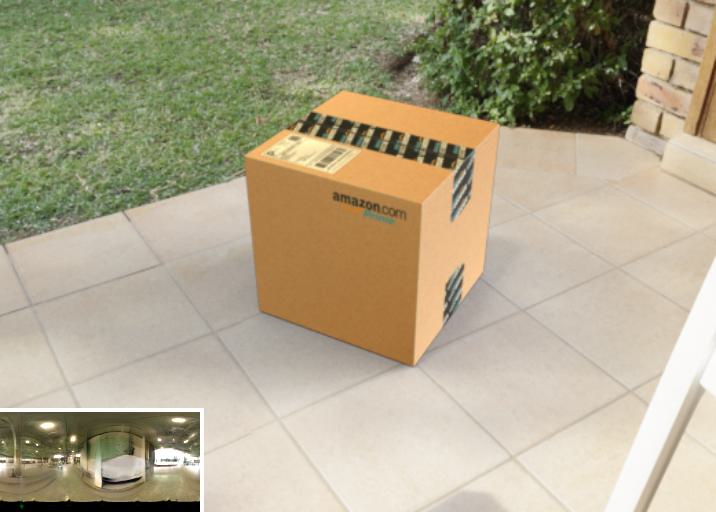} &
        \includegraphics[width=0.98\linewidth, trim={0 0 0 0},clip]{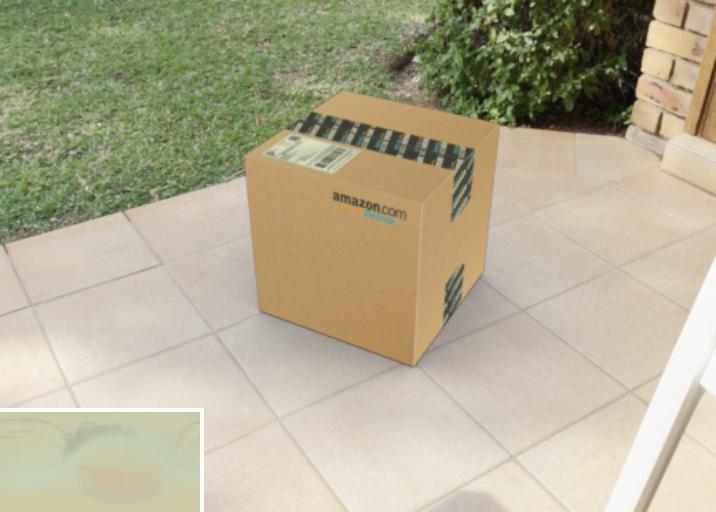}
        \\
        \includegraphics[width=0.98\linewidth, trim={0 0 0 0},clip]{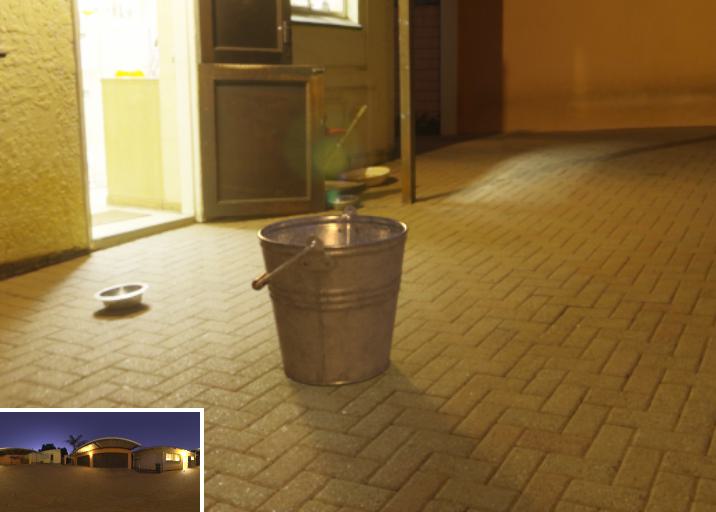} &
        \includegraphics[width=0.98\linewidth, trim={0 0 0 0},clip]{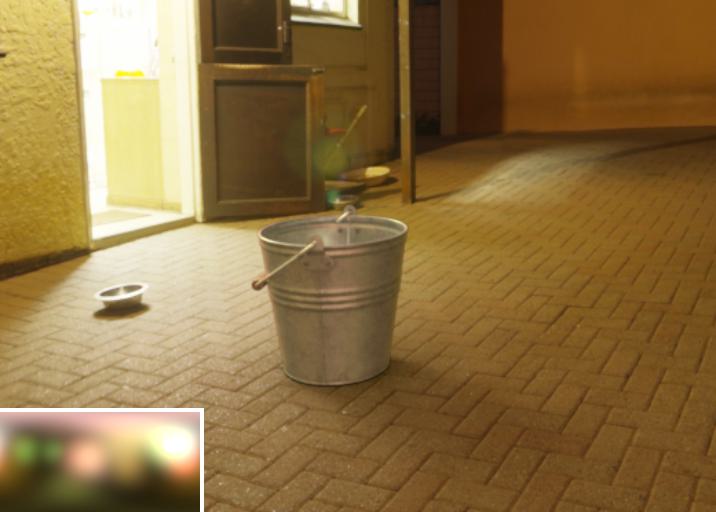} &
        \includegraphics[width=0.98\linewidth, trim={0 0 0 0},clip]{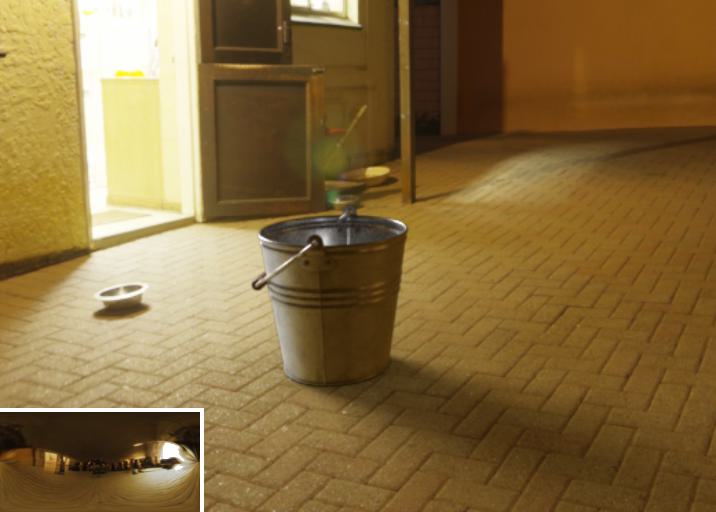} &
        \includegraphics[width=0.98\linewidth, trim={0 0 0 0},clip]{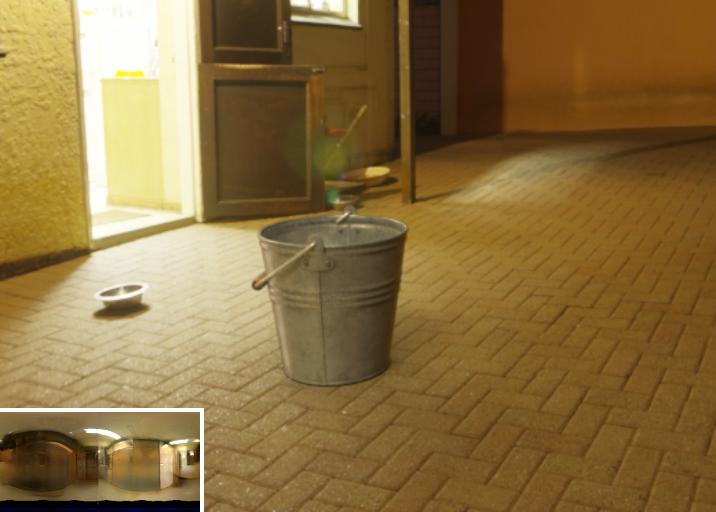} &
        \includegraphics[width=0.98\linewidth, trim={0 0 0 0},clip]{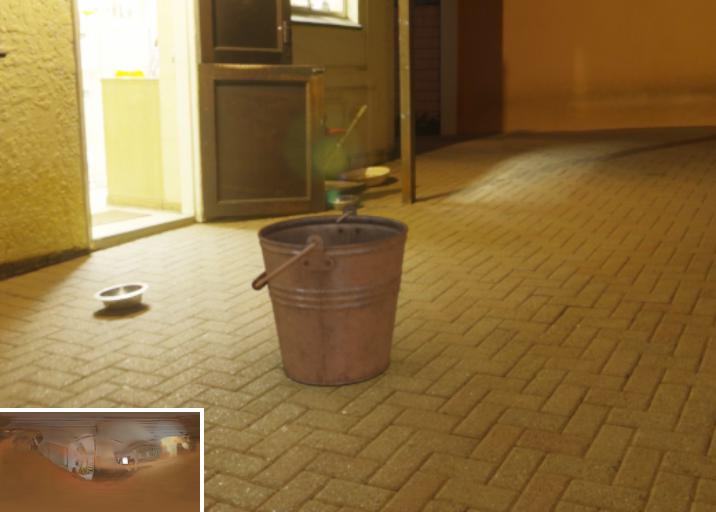}
        \\
        \includegraphics[width=0.98\linewidth, trim={0 0 0 0},clip]{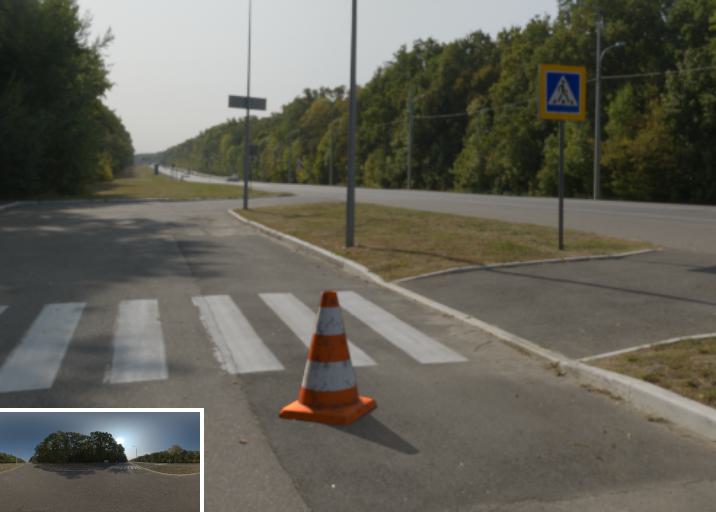} &
        \includegraphics[width=0.98\linewidth, trim={0 0 0 0},clip]{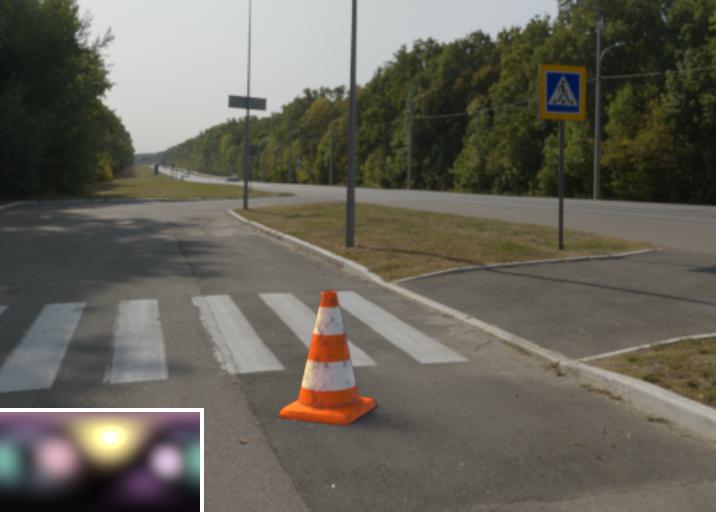} &
        \includegraphics[width=0.98\linewidth, trim={0 0 0 0},clip]{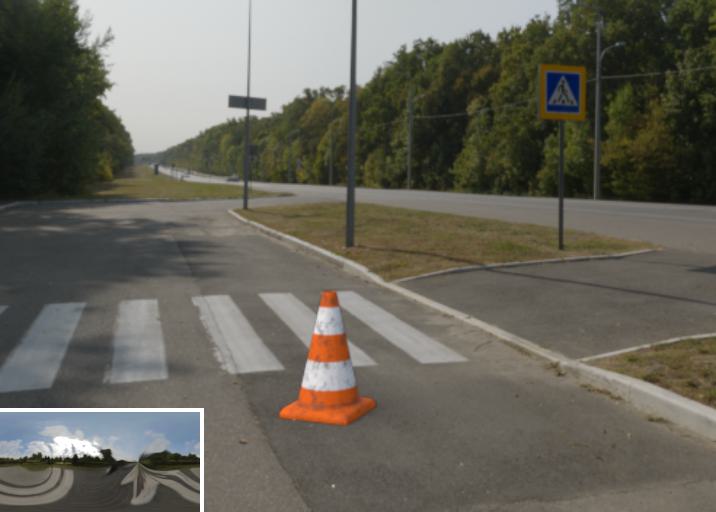} &
        \includegraphics[width=0.98\linewidth, trim={0 0 0 0},clip]{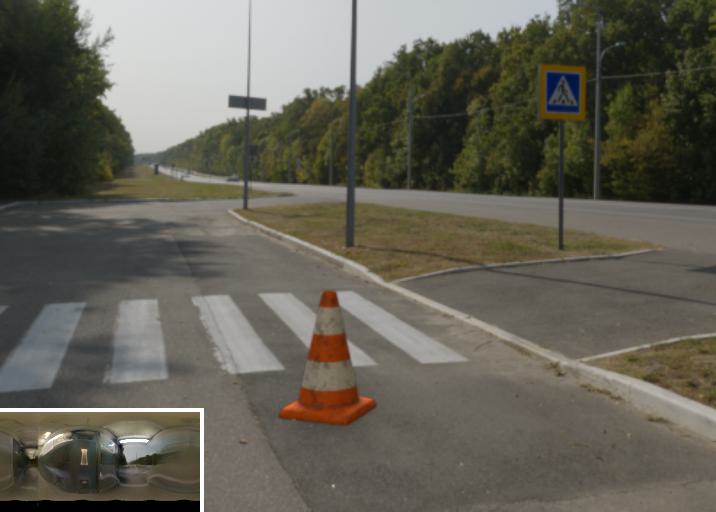} &
        \includegraphics[width=0.98\linewidth, trim={0 0 0 0},clip]{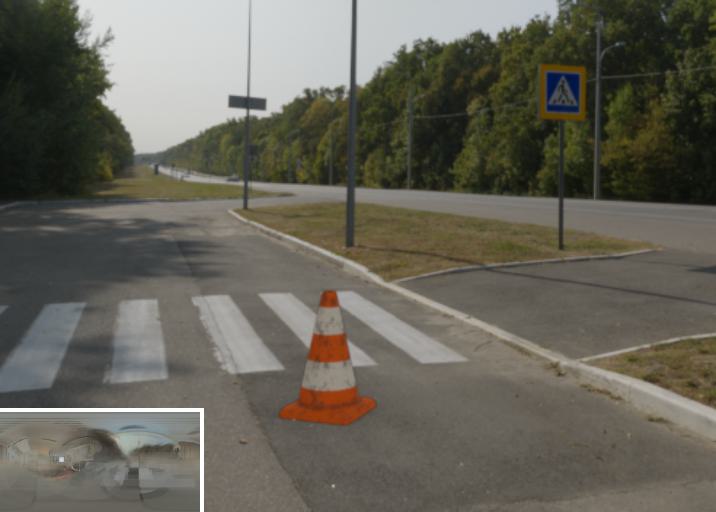}
        \\
        \includegraphics[width=0.98\linewidth, trim={0 0 0 0},clip]{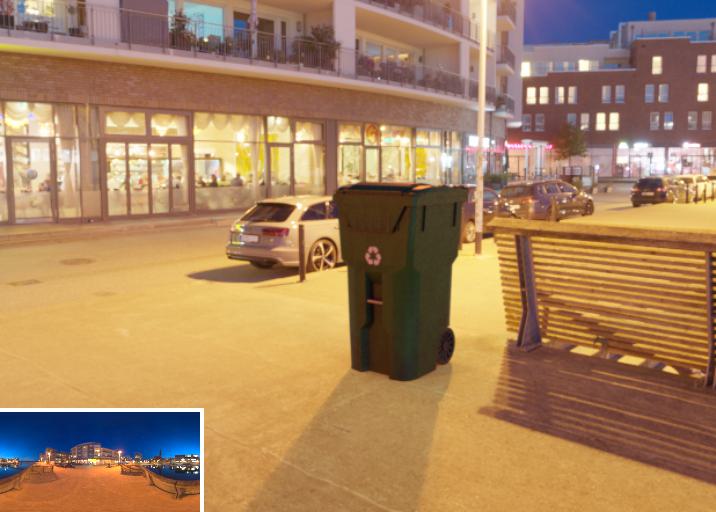} &
        \includegraphics[width=0.98\linewidth, trim={0 0 0 0},clip]{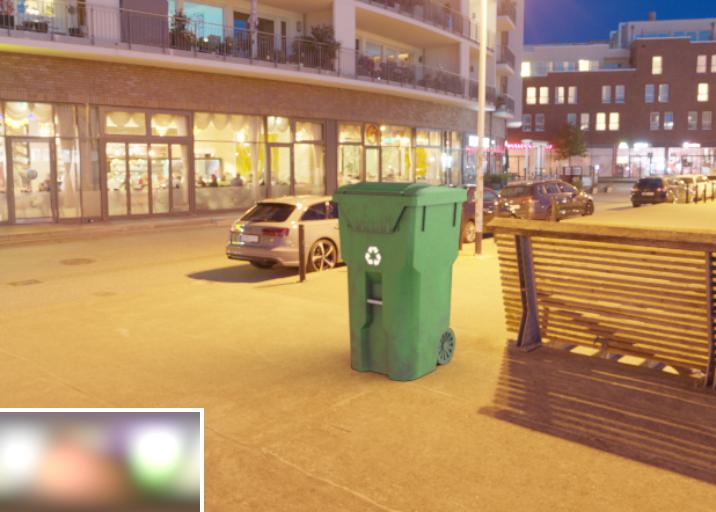} &
        \includegraphics[width=0.98\linewidth, trim={0 0 0 0},clip]{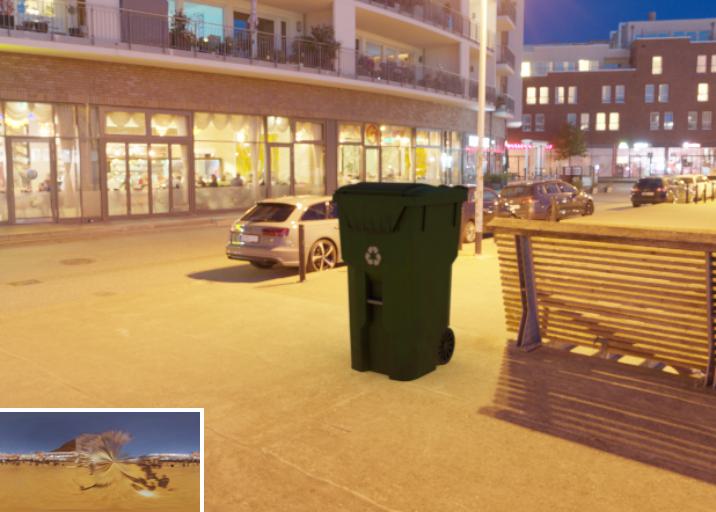} &
        \includegraphics[width=0.98\linewidth, trim={0 0 0 0},clip]{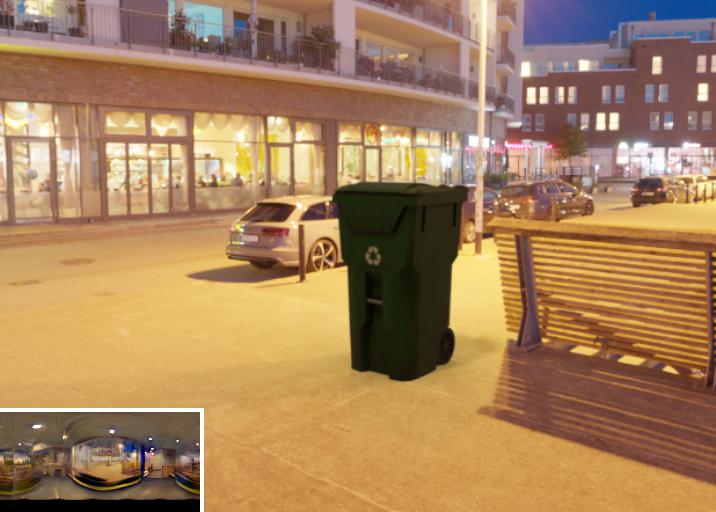} &
        \includegraphics[width=0.98\linewidth, trim={0 0 0 0},clip]{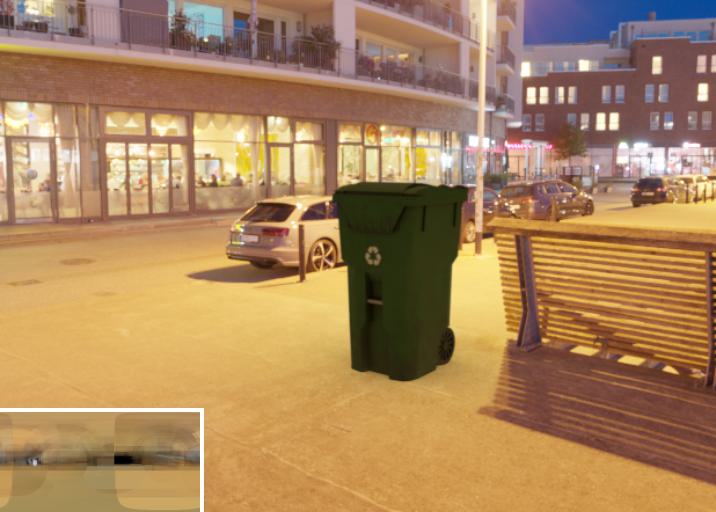}
    \end{tabularx}%
    \makeatletter\def\@captype{figure}\makeatother

    \caption{
        Additional visual comparisons on inserting objects into cropped HDRIs from PolyHaven.
    }
    \vspace{-7mm}
    \label{fig:qual_polyhaven_supp}
\end{table}

\begin{table}[t]
    \centering
    \scriptsize
    \captionsetup{font=small}
        \setlength\tabcolsep{0pt}
    \begin{tabularx}{\linewidth}%
        {*{4}{>{\centering\arraybackslash}X}}
        \includegraphics[width=0.98\linewidth, trim={0 0 0 0},clip]{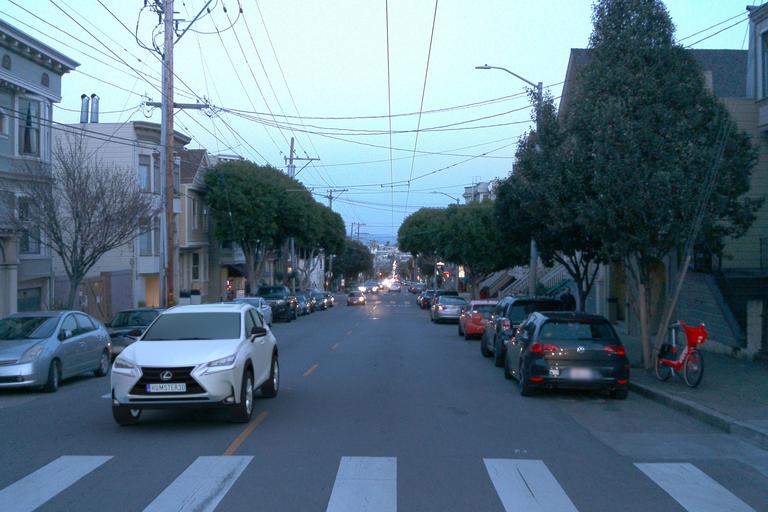} &
        \includegraphics[width=0.98\linewidth, trim={0 0 0 0},clip]{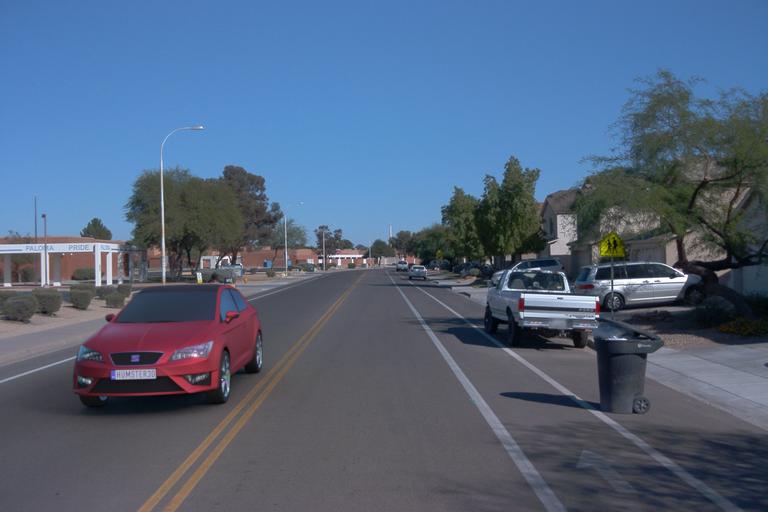} & 
        \includegraphics[width=0.98\linewidth, trim={0 0 0 0},clip]{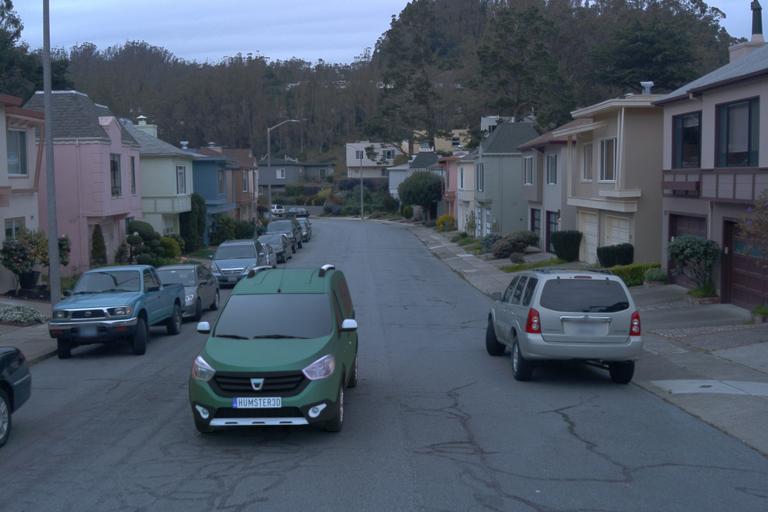} & 
        \includegraphics[width=0.98\linewidth, trim={0 0 0 0},clip]{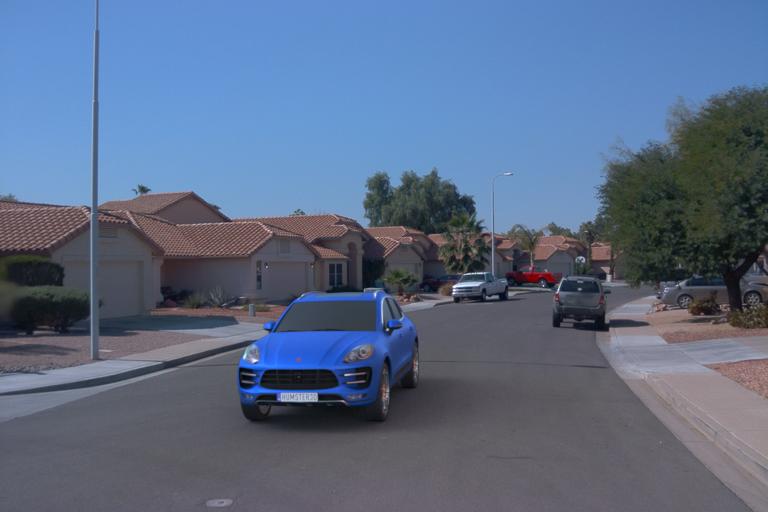} 
        \\
        \includegraphics[width=0.98\linewidth, trim={0 0 0 0},clip]{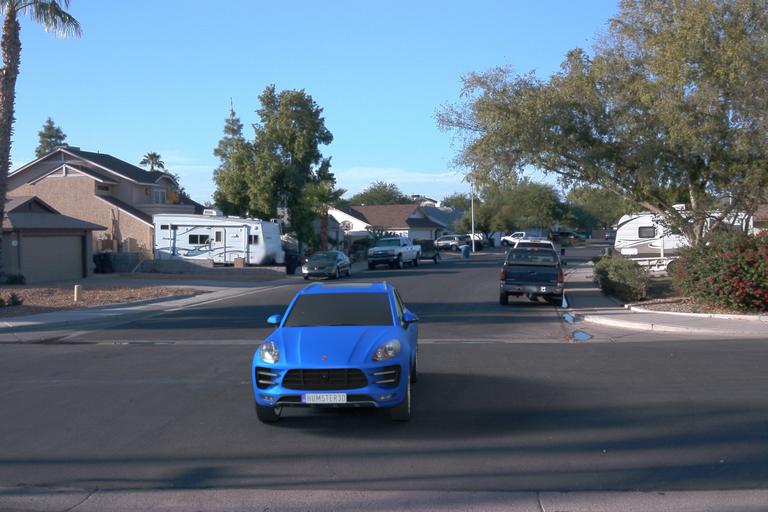} &
        \includegraphics[width=0.98\linewidth, trim={0 0 0 0},clip]{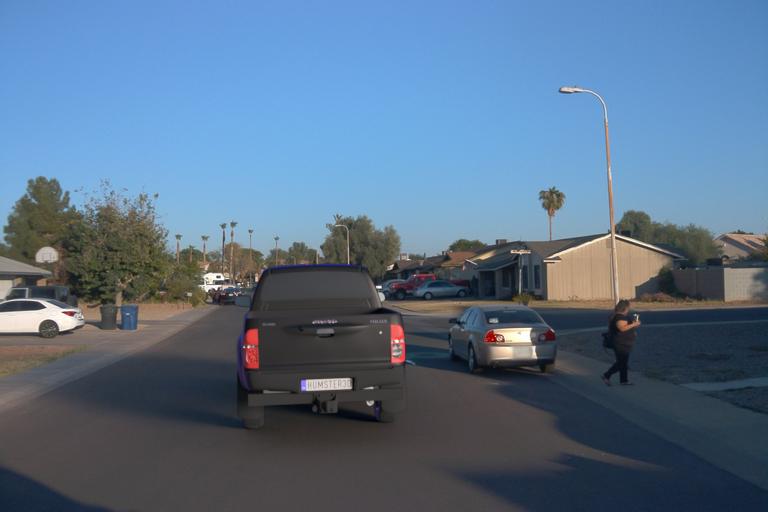} & 
        \includegraphics[width=0.98\linewidth, trim={0 0 0 0},clip]{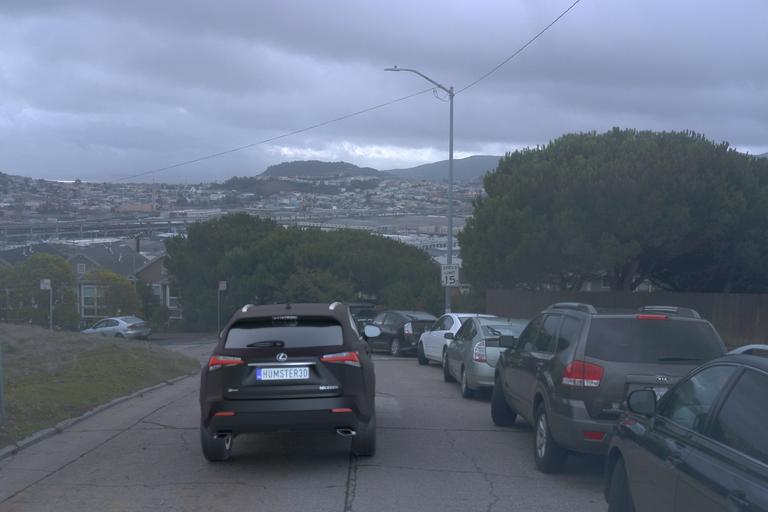} & 
        \includegraphics[width=0.98\linewidth, trim={0 0 0 0},clip]{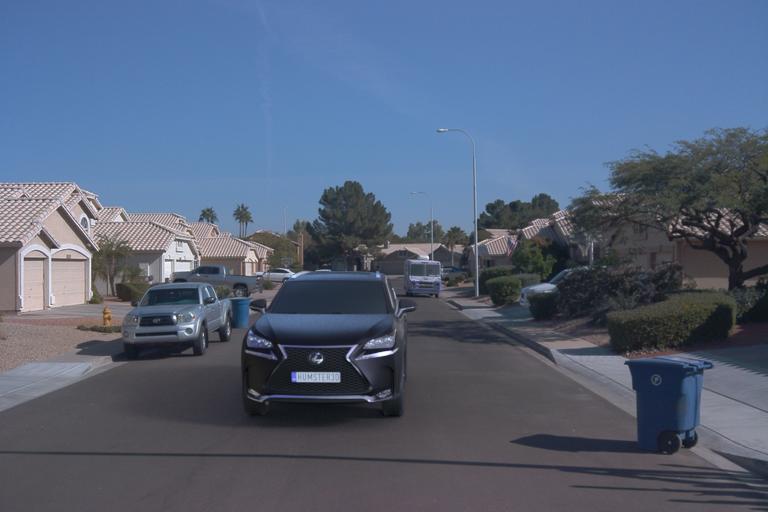} 
        \\
        \includegraphics[width=0.98\linewidth, trim={0 0 0 0},clip]{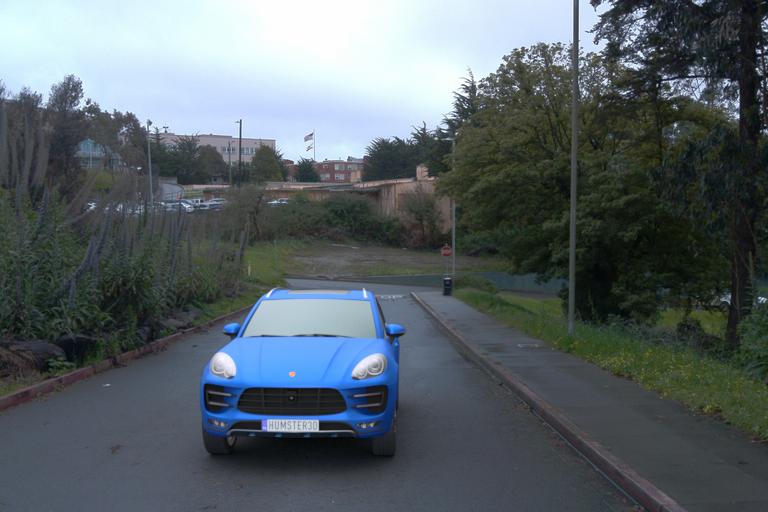} &
        \includegraphics[width=0.98\linewidth, trim={0 0 0 0},clip]{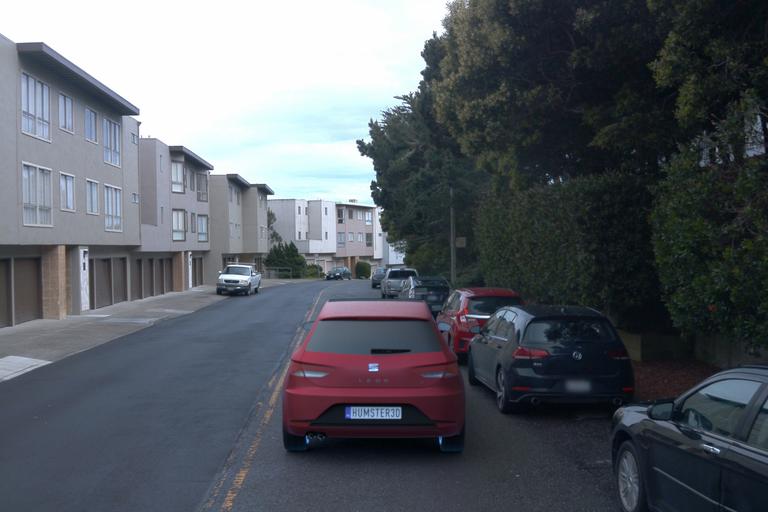} & 
        \includegraphics[width=0.98\linewidth, trim={0 0 0 0},clip]{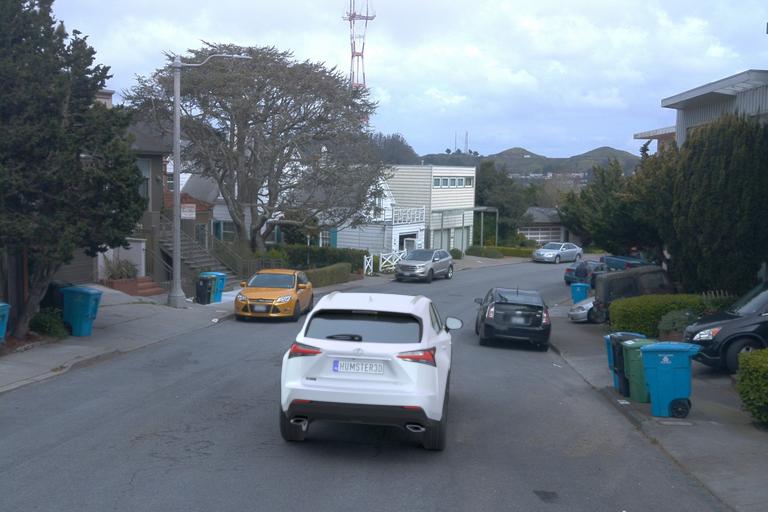} & 
        \includegraphics[width=0.98\linewidth, trim={0 0 0 0},clip]{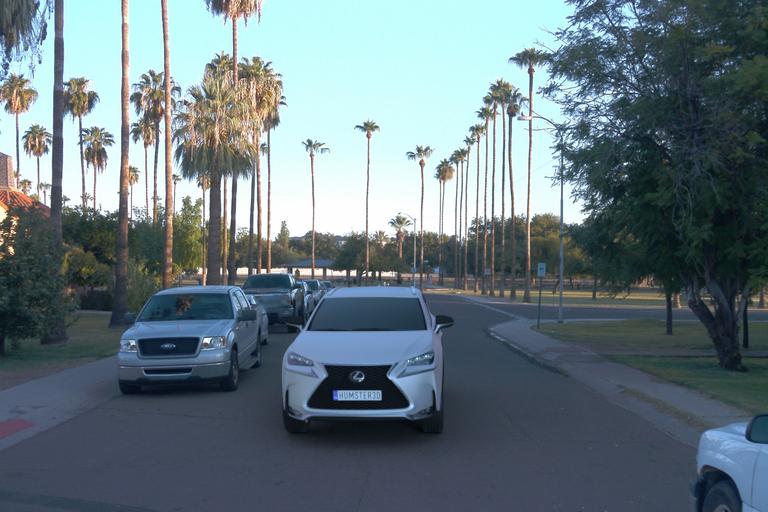} 
        \\
        \includegraphics[width=0.98\linewidth, trim={0 0 0 0},clip]{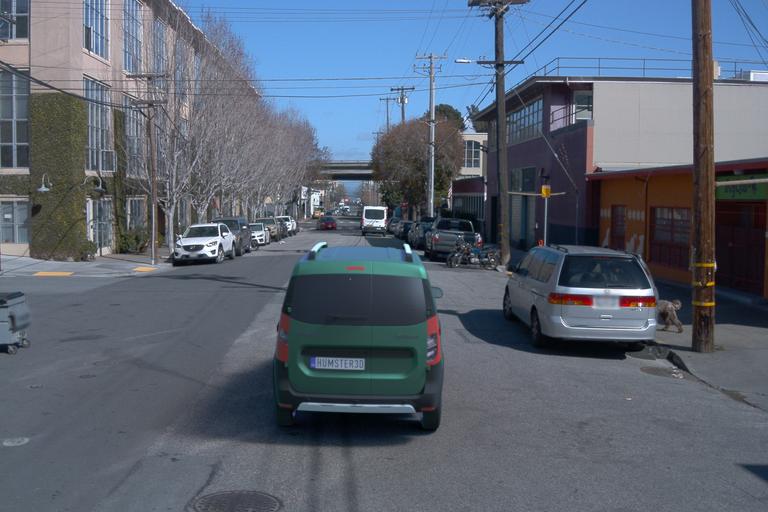} &
        \includegraphics[width=0.98\linewidth, trim={0 0 0 0},clip]{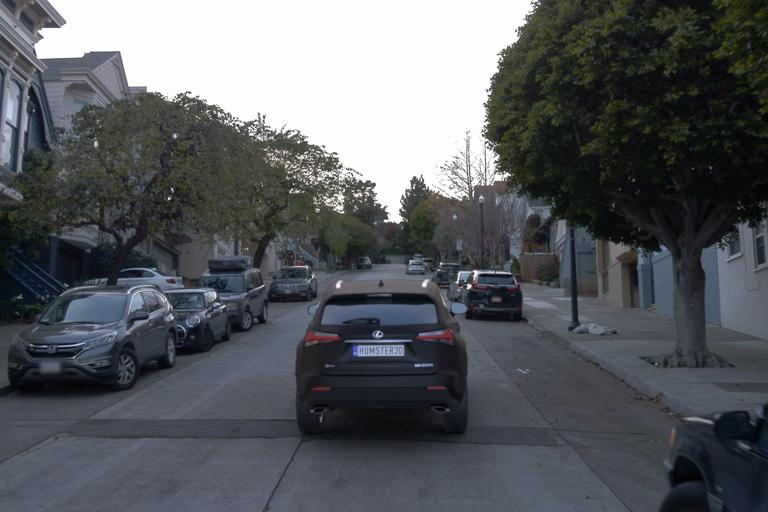} & 
        \includegraphics[width=0.98\linewidth, trim={0 0 0 0},clip]{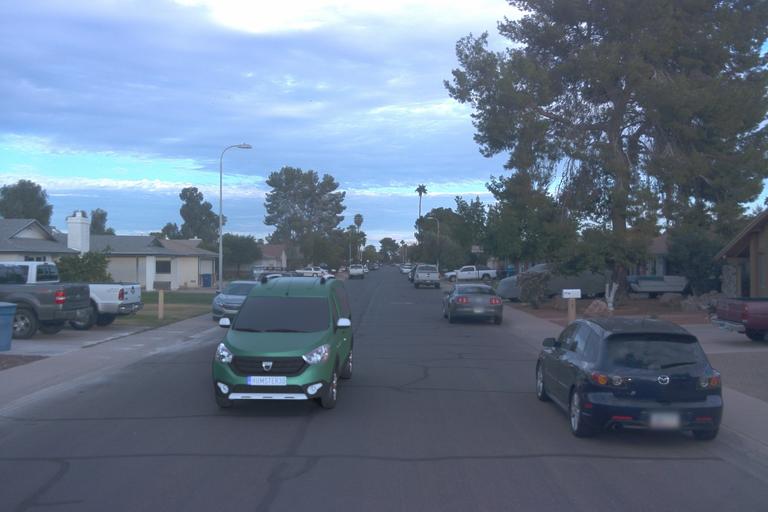} & 
        \includegraphics[width=0.98\linewidth, trim={0 0 0 0},clip]{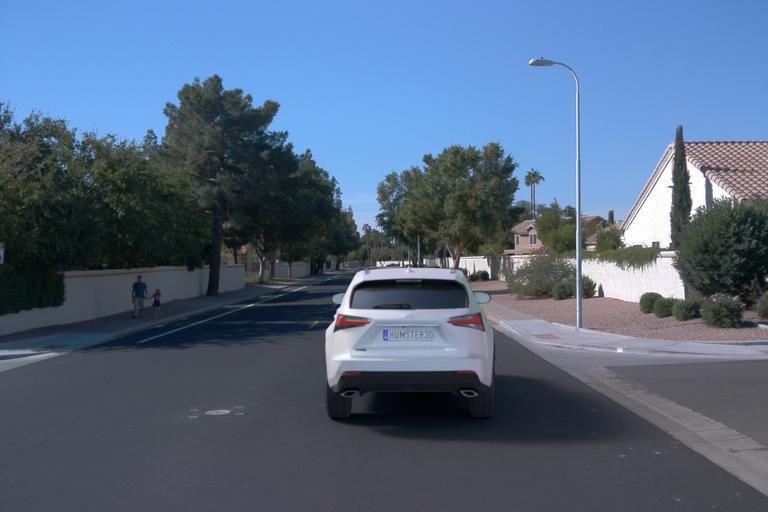} 
        \\
        \includegraphics[width=0.98\linewidth, trim={0 0 0 0},clip]{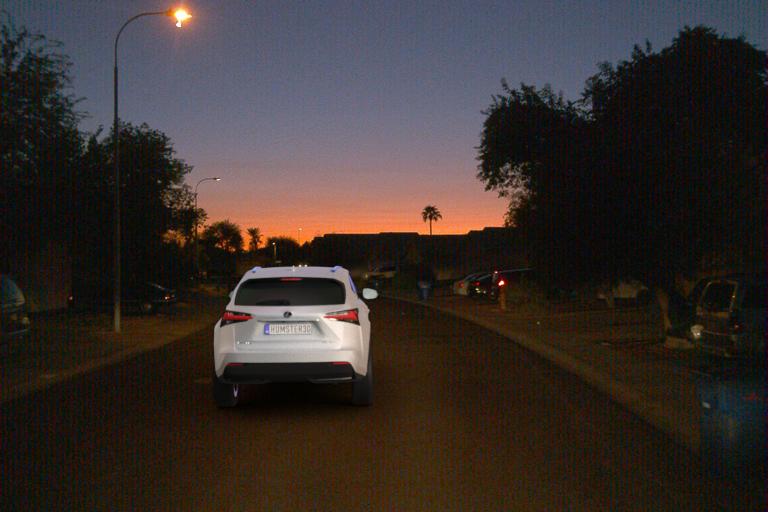} &
        \includegraphics[width=0.98\linewidth, trim={0 0 0 0},clip]{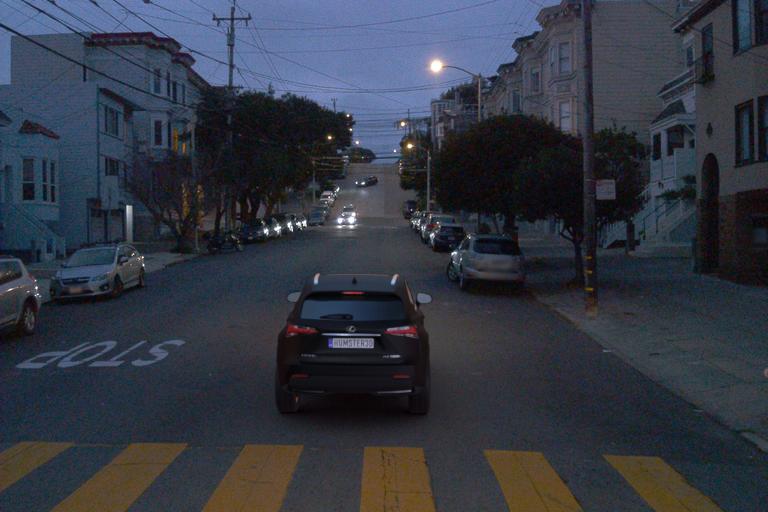} & 
        \includegraphics[width=0.98\linewidth, trim={0 0 0 0},clip]{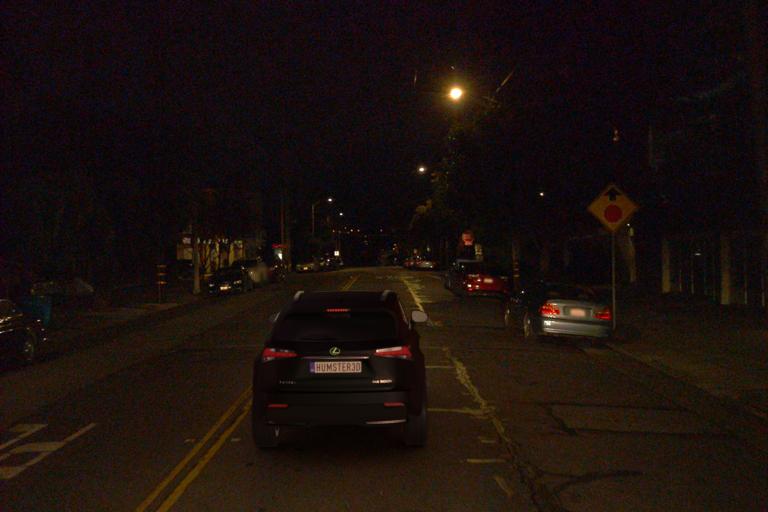} & 
        \includegraphics[width=0.98\linewidth, trim={0 0 0 0},clip]{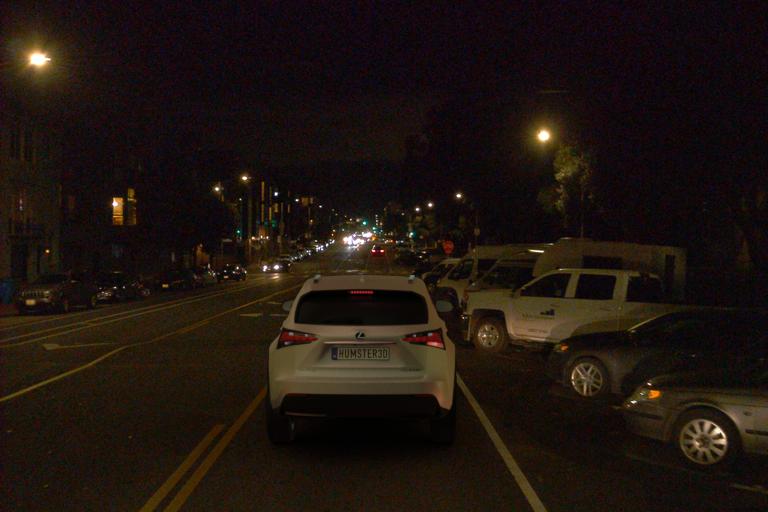} 
    \end{tabularx}%
    \makeatletter\def\@captype{figure}\makeatother

    \vspace{2mm}
    \caption{
        Additional car insertion examples on Waymo driving scenes.
    }
    \vspace{-7mm}
    \label{fig:waymo_more}
\end{table}

\begin{table}[ht]
    \centering
    \footnotesize
    \setlength\tabcolsep{0pt}
    \begin{tabularx}{\linewidth}%
        {p{1em}*{3}{>{\centering\arraybackslash}X}} 
        {\rotatebox{90}{\,\,\, Background}}&
        \includegraphics[width=0.98\linewidth, trim={0 0 0 0},clip]{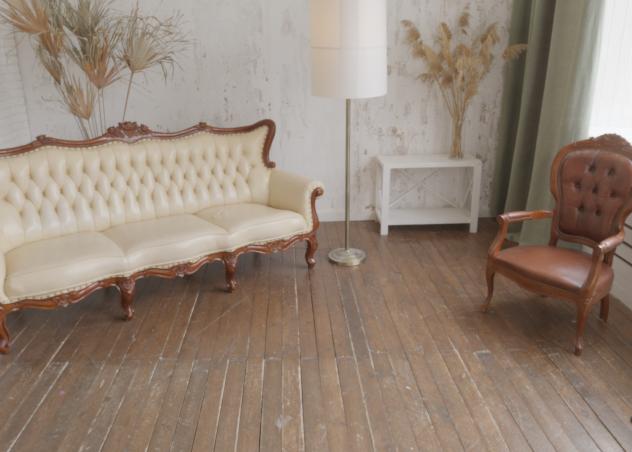} &
        \includegraphics[width=0.98\linewidth, trim={0 0 0 0},clip]{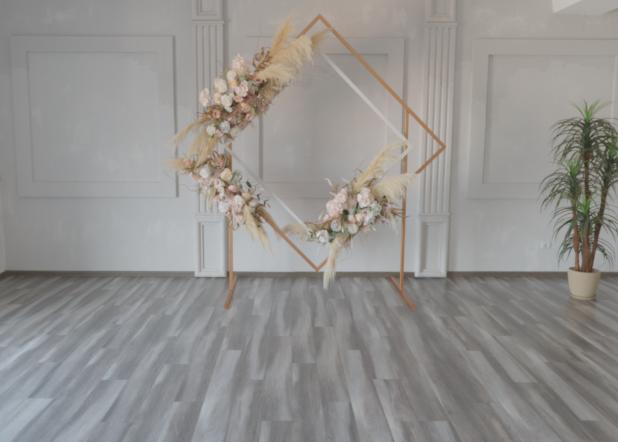} &
        \includegraphics[width=0.98\linewidth, trim={0 0 0 0},clip]{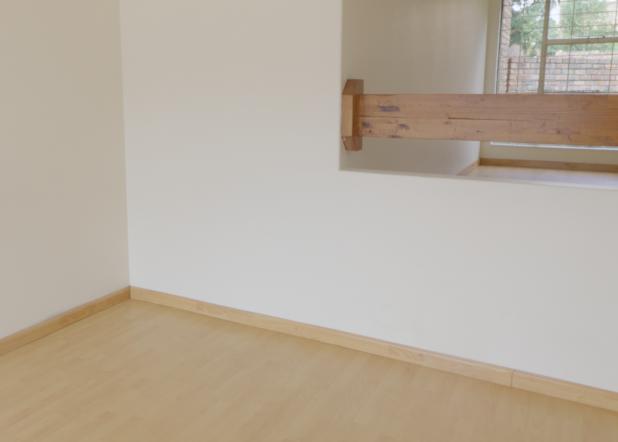}
         \\
        {\rotatebox{90}{\quad\, Our Insertion}}&
        \includegraphics[width=0.98\linewidth, trim={0 0 0 0},clip]{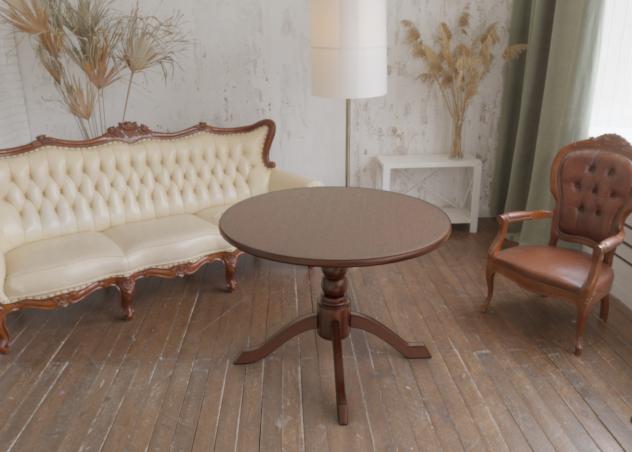} &
        \includegraphics[width=0.98\linewidth, trim={0 0 0 0},clip]{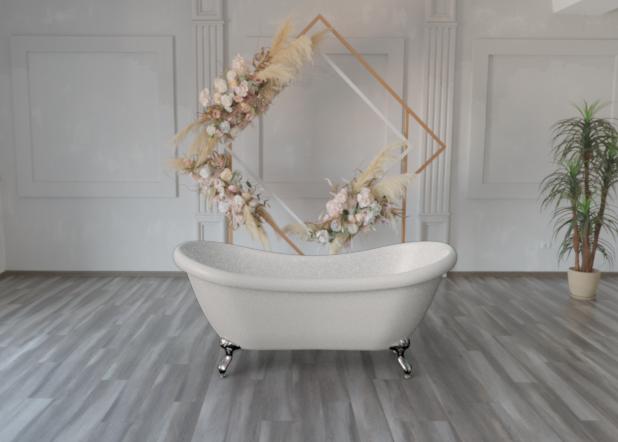} &
        \includegraphics[width=0.98\linewidth, trim={0 0 0 0},clip]{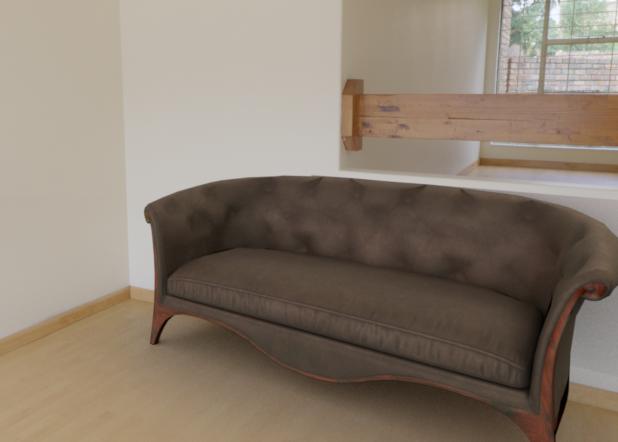}
         \\
        & \textit{Table} & \textit{Bathtub} & \textit{Sofa}
    \end{tabularx}%
    \makeatletter\def\@captype{figure}\makeatother
    \vspace{2mm}
    \caption{
        Additional object insertion examples on cropped PolyHaven HDRIs.
    }
    \label{fig:hdri_more}
    \vspace{-10pt}
\end{table}

\section{Discussion}
\label{sec:supp_discussion}

\subsubsection{Broader impact.} 

This paper introduces a novel approach to producing virtual object insertion in images leveraging the power of diffusion models and inverse rendering techniques. 
It could benefit digital content creation by providing filmmakers and game developers with a powerful tool to create novel scenarios and reducing costly manual editing. 
Its application in AR and VR can enhance user experiences, making digital interactions feel more natural and engaging. 
On a broader scale, this work contributes to the field of computer vision and graphics, and showcases the potential of combining powerful diffusion models with classic rendering techniques.

Similar to other photorealistic image editing technologies such as ``deep fakes'', there is also the potential for misuse, \eg it might potentially be used to create misleading content and propagate misinformation. Related technology on identifying and filtering out such content can mitigate these negative applications.

\end{document}